\documentclass[]{style/iromlab}

\DeclareDocumentEnvironment{example}{}{\noindent\textbf{Running example:}\itshape}{}

\usepackage{ifthen}
\newboolean{include-notes}
\newboolean{include-new}
\newboolean{include-remove}
\setboolean{include-notes}{true}
\setboolean{include-new}{false}
\setboolean{include-remove}{false}

\usepackage[dvipsnames]{xcolor}
\usepackage[normalem]{ulem}
\newcommand{\justin}[1]{\ifthenelse{\boolean{include-notes}}{\textcolor{orange}{\textbf{Jaime:} #1}}{}}

\newcommand{\princeton}[1]{\ifthenelse{\boolean{include-notes}}{\textcolor{orange}{#1}}{}}

\newcommand{\p}[1]{\smallskip \noindent \textbf{{#1}.}}

\usepackage{multicol}
\usepackage{amsmath, amsfonts, amssymb, amsthm}
\usepackage{enumerate}
\usepackage[inline]{enumitem}
\usepackage{mathtools}
\usepackage{graphicx}
\usepackage{longtable,tabularx}
\usepackage{placeins} 
\usepackage{float}
\usepackage{multirow}
\usepackage{bbm}
\usepackage{threeparttable}
\usepackage{balance}
\usepackage[ruled,algo2e]{algorithm2e}
\usepackage{algpseudocode}
\usepackage{adjustbox}
\usepackage{booktabs}
\usepackage{bm}
\usepackage{etoolbox}
\usepackage{microtype}
\usepackage[title]{appendix}
\usepackage{units}
\usepackage{cleveref}
\usepackage{xspace}
\usepackage{tcolorbox}
\usepackage{caption}

\definecolor{claude_color}{HTML}{F89E62}
\definecolor{deepseek_color}{HTML}{78B6E8}
\definecolor{o3_mini_color}{HTML}{6CD5A1}
\definecolor{red_color}{HTML}{E13B55}
\definecolor{prompt_color}{HTML}{71502B}

\tcbset{
  promptstyle/.style={
    colback=gray!10,
    colframe=gray!60,
    boxrule=0.5pt,
    arc=2pt,
    outer arc=2pt,
    left=4pt,
    right=4pt,
    top=4pt,
    bottom=4pt,
    fonttitle=\bfseries,
    sharp corners=south,
  }
}

\tcbset{
  promptstyle_prompt_iuq/.style={
    colback=prompt_color!10,
    colframe=prompt_color!60,
    coltitle=white,
    boxrule=1.0pt,
    arc=2pt,
    outer arc=2pt,
    left=4pt,
    right=4pt,
    top=4pt,
    bottom=4pt,
    fonttitle=\bfseries,
  }
}

\tcbset{
  promptstyle_claude/.style={
    colback=claude_color!10,
    colframe=claude_color!60,
    coltitle=black,
    boxrule=1.0pt,
    arc=2pt,
    outer arc=2pt,
    left=4pt,
    right=4pt,
    top=4pt,
    bottom=4pt,
    fonttitle=\bfseries,
  }
}

\tcbset{
  promptstyle_deepseek/.style={
    colback=deepseek_color!10,
    colframe=deepseek_color!60,
    coltitle=black,
    boxrule=1.0pt,
    arc=2pt,
    outer arc=2pt,
    left=4pt,
    right=4pt,
    top=4pt,
    bottom=4pt,
    fonttitle=\bfseries,
  }
}

\tcbset{
  promptstyle_o3_mini/.style={
    colback=o3_mini_color!10,
    colframe=o3_mini_color!60,
    coltitle=black,
    boxrule=1.0pt,
    arc=2pt,
    outer arc=2pt,
    left=4pt,
    right=4pt,
    top=4pt,
    bottom=4pt,
    fonttitle=\bfseries,
  }
}

\DeclareMathOperator*{\argmin}{arg\,min}

\newcommand*{\trace}[1]{\ensuremath{\mathrm{trace}\left({#1}\right)}}

\newcommand*{\gt}{\ensuremath{\mathrm{gt}}}

\newcommand{\mcal}[1]{\mathcal{#1}}
\newcommand{\mbb}[1]{\mathbb{#1}}
\newcommand{\T}{\mathsf{T}}

\newcommand{\mbf}[1]{\ensuremath{\mathbf{#1}}}

\newcommand{\SE}{\ensuremath{\mathrm{SE}}}
\newcommand{\SO}{\ensuremath{\mathrm{SO}}}
\newcommand{\so}{\ensuremath{\mathfrak{so}}}

\newcommand*{\meter}{\ensuremath{\mathrm{m}}}
\newcommand{\algname}{\mbox{SPINE}\xspace}

\newbool{extended}
\setbool{extended}{false}

\makeatletter
\newcommand{\longdash}[1][2em]{%
  \makebox[#1]{$\m@th\smash-\mkern-7mu\cleaders\hbox{$\mkern-2mu\smash-\mkern-2mu$}\hfill\mkern-7mu\smash-$}}
\makeatother
\newcommand{\omitskip}{\kern-\arraycolsep}

\author[1*]{Zhiting Mei}
\author[1*]{Ola Shorinwa}
\author[1]{Anirudha Majumdar}

\affiliation[1]{Princeton University}
\contribution[*]{Equal contribution.}

\begin{document}

\title{
\LARGE{Geometry Meets Vision:} \\
\Large{Revisiting Pretrained Semantics in Distilled Fields}
}

\abstract{

Semantic distillation in radiance fields has spurred significant advances in open-vocabulary robot policies, e.g., in manipulation and navigation,
founded on pretrained semantics from large vision models.
While prior work has demonstrated the effectiveness of 
visual-only semantic features (e.g., DINO and CLIP) in Gaussian Splatting and neural radiance fields, the potential benefit of geometry-grounding in distilled fields remains an open question.
In principle, visual-geometry features seem very promising for spatial tasks such as pose estimation,
prompting the question: \textit{Do~geometry-grounded semantic features offer an edge in distilled fields?}
Specifically, we ask three critical questions: 
\textit{First, does spatial-grounding produce higher-fidelity geometry-aware semantic features?} We find that image features from geometry-grounded backbones contain finer structural details compared to their counterparts. Secondly, \textit{does geometry-grounding improve semantic object localization?} We observe no significant difference in this task. 
Thirdly, \textit{does geometry-grounding enable higher-accuracy radiance field inversion?}
Given the limitations of prior work and their lack of semantics integration, we propose a novel framework \textbf{\algname} for inverting radiance fields without an initial guess,
consisting of two core components:
(i) coarse inversion using distilled semantics, and
(ii) fine inversion using photometric-based optimization.
Surprisingly, we find that the pose estimation accuracy \emph{decreases} with geometry-grounded features.
Our results suggest that visual-only features offer greater versatility for a broader range of downstream tasks, although geometry-grounded features contain more geometric detail. Notably, our findings underscore the necessity of future research on effective strategies for geometry-grounding that augment the versatility and performance of pretrained semantic features.

}


\website{
https://spine-geo.github.io %
}
{
spine-geo.github.io   %
}

\code{
https://github.com/irom-princeton/spine %
}
{
github.com/irom-princeton/spine  %
}

\maketitle

\begin{figure}[th]
    \centering
    \includegraphics[width=\linewidth]{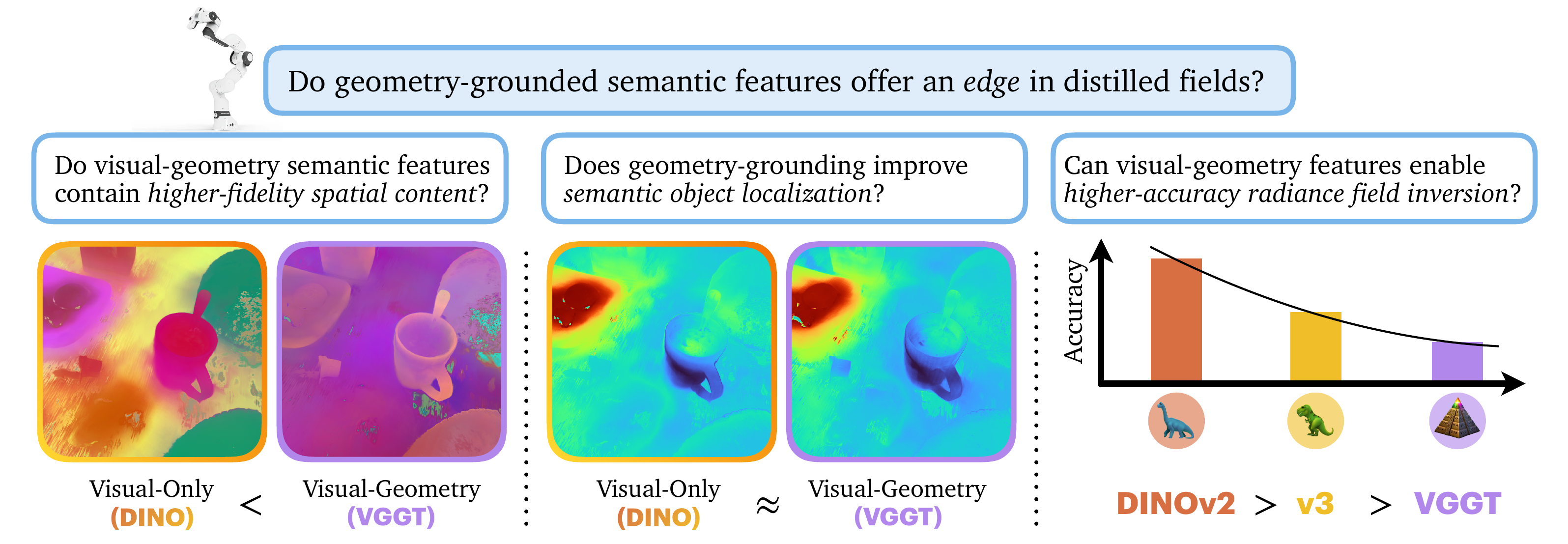}
    \caption{We revisit pretrained semantics in distilled radiance fields, asking three critical questions to compare \emph{visual-geometry} semantic features against \emph{visual-only} features. We find that while visual-geometry features retain richer spatial fidelity, they do not improve performance in downstream tasks such as semantic localization or radiance field inversion, suggesting the greater versatility of visual-only semantic features.
    }
    \label{fig:anchor}
\end{figure}

\section{Introduction}
\label{sec:introduction}
Large foundation models have driven rapid advances in open-vocabulary robot policies, enabling robots to perform complex, multi-stage tasks, entirely from natural-language instructions; \mbox{see~\citep{firoozi2025foundation}} for a detailed review. Through semantic distillation, prior work blends photorealistic novel-view synthesis from radiance fields, such as Gaussian Splatting \citep{kerbl20233d} and neural radiance fields (NeRFs) \citep{mildenhall2021nerf}, with generalizable pretrained semantics from foundation models to endow robots with semantic task understanding capabilities grounded in the real-world. This synergy underlies the success of many language-conditioned robot policies in manipulation \citep{ze2023gnfactor, shen2023distilled}
navigation \citep{chen2025splat}.

In general, large vision backbones (CLIP~\citep{radford2021learning}, DINOv2 \citep{oquab2023dinov2}, DINOv3 \citep{simeoni2025dinov3}) have been limited to visual feature learning without spatial grounding, potentially impeding the emergence of $3$D-aware features.
However, recent work VGGT \citep{wang2025vggt}
demonstrates that large vision backbones can be trained to produce visual-geometry features by grounding these models with a $3$D reconstruction task objective, achieving superior performance in many downstream tasks, such as keypoint tracking.
Although visual-geometry features seem promising for spatial tasks such as pose estimation, the actual performance of \emph{visual-geometry} semantic features relative to \emph{visual-only} semantic features remains unknown, particularly in distilled radiance fields.
To address this gap, we revisit pretrained semantics in distilled fields, asking: \emph{Do visual-geometry semantic features offer an edge in distilled fields?} To rigorously explore the relative performance between visual-only features (DINOv2
and DINOv3)
and visual-geometry features (VGGT),
we ask three critical questions, each focused on important downstream applications of distilled radiance fields in robotics, illustrated in \Cref{fig:anchor}.

First, \emph{do visual-geometry semantic features contain higher-fidelity spatial content?} We examine the information content of visual-only and visual-geometry features, mapping these features from their respective semantic spaces to a lower-dimensional visual space. We find that \textit{visual-geometry semantics provide finer spatial content, with more prominent structural detail, e.g., sharper edges, more accurate subpart decomposition, etc}. We note that high-fidelity geometric detail could be essential in some robotics applications, e.g., fine and dexterous manipulation.

Second, \emph{does geometry-grounding improve semantic object localization?} Many state-of-the-art robot policies rely on semantic object localization in distilled radiance fields for successful task execution, underscoring its immense relevance. We find that \textit{visual-geometry semantic features perform similarly to visual-only features}, which suggests that spatial grounding does not boost object-class semantic feature content. 
However, effective co-supervision of semantic features with geometry and vision can lead to improved performance, presenting an important area for future research.

Third, \emph{can visual-geometry features enable higher-accuracy radiance field inversion?} Unlike rendering/rasterization, radiance field inversion remains a challenging problem, with existing methods limited by need for good initial estimates. To address this problem and further facilitate semantics evaluation, we introduce a novel framework \emph{\algname} for inverting radiance fields without any camera pose. \algname directly leverages embedded pretrained semantics to compute: (i) coarse pose estimates using a co-trained semantic field which maps semantic features to a distribution over camera poses and (ii) fine pose estimates by refining the coarse solution through novel-view synthesis in radiance fields and robust perspective-$n$-point optimization.  
Surprisingly, we observe that visual-geometry semantic features underperform visual-only features, 
despite their greater spatial content.

Our findings highlight that \textit{visual-only semantic features are more versatile than visual-geometry features}, even in spatial tasks like pose estimation.
Importantly, our results underscore the need for
additional research on effective strategies for geometry-grounding to augment the versatility and performance of pretrained semantic features. 

\section{Related Work}
\label{sec:related_work}
\smallskip
\textbf{Radiance Fields }
marked a notable breakthrough in $3$D scene reconstruction, achieving photorealistic image rendering and novel-view synthesis entirely from RGB images. NeRFs learn separate density and color fields parameterized by multi-layer perceptrons, mapping a $3$D point at a specified camera direction to its volume density (associated with its opacity) and radiance (color). Given its generality to different sensing modalities, NeRFs have been widely applied in robotics, e.g., robot planning \citep{adamkiewicz2022vision, chen2024catnips}, localization \citep{yen2021inerf, maggio2022loc}, and manipulation \citep{kerr2022evo, weng2022neural}.
In contrast to implicit representation of NeRFs, Gaussian Splatting (GS) \citep{kerbl20233d}
utilizes explicit ellipsoidal primitives, each parameterized by a mean, covariance, opacity, and color, to represent non-empty space. This design choice enables the use of fast tile-based rasterization for faster training and real-time rendering speeds. Like NeRFs, GS has found widespread applications in robotics, e.g., robot planning and localization \citep{chen2025splat, michaux2025let} and manipulation \citep{lu2024manigaussian}.

\p{Distilled Semantics in Radiance Fields}
Foundation models, e.g., CLIP 
and DINO 
learn robust visual features from images/language that encapsulate open-world semantics through pretraining on internet-scale data. Increasingly, 
robot policies have embedded semantics from CLIP and DINO into 
radiance fields to enable language-conditioned robot manipulation \citep{rashid2023language, shen2023distilled, shorinwa2024splat}, mapping 
\citep{shorinwa2025siren},
and object localization \citep{yin2025womap}. 
In general, these methods combine visual-only image features, e.g., DINO, with vision-language semantics from CLIP to support language conditioning, particularly for semantic localization, which plays an essential role in downstream robotics tasks. 
Recent work 
trains foundation models for 3D reconstruction to learn visual-geometry features with potentially useful applications to spatial tasks. However, no existing work has examined the integration of these features in distilled radiance fields. 
In this work, we explore visual-geometry semantic features in distilled radiance fields, quantifying the performance of these features relative to visual-only features in important robotics applications.

\section{Preliminaries}
\label{sec:preliminaries}
We review  important technical concepts, necessary for understanding the discussion in this paper. We introduce NeRFs, GS, and discuss semantic distillation in radiance fields in \Cref{sec:app_preliminaries}.

\p{NeRFs}
NeRFs 
learn implicit volumetric color and density fields, encoding the occupancy and radiance of each point in the scene, given a set of images $\mcal{I}$ and corresponding camera poses, which is typically computed via structure-from-motion \citep{schonberger2016structure}. Specifically, the color field ${\mbf{c}: \mbb{R}^{3} \times \mbb{S}^{2} \mapsto \mbb{R}^{3}}$ maps a $3$D point ${\mbf{x} \in \mbb{R}^{3}}$ and a camera viewing direction ${d \in \mbb{S}^{2}}$ to an RGB color ${\mbf{c}(\mbf{x}, d)}$. Likewise, the density field ${\rho: \mbb{R}^{3} \mapsto \mbb{R}_{+}}$ maps $\mbf{x}$ to a non-negative volume density ${\rho(\mbf{x})}$, representing the differential probability of a light ray terminating at a particle located at $\mbf{x}$. Using ray-marching to render images from $\mbf{c}$ and $\rho$, NeRFs utilize stochastic gradient descent to optimize the parameters of $\mbf{c}$ and $\rho$ (represented as MLPs), minimizing the photometric error between the rendered and ground-truth images.

\p{Gaussian Splatting}
Gaussian Splatting (GS) utilizes $2$D \citep{huang20242d} or $3$D \citep{kerbl20233d} Gaussian primitives to represent non-empty space, each defined by a mean ${\mu \in \mbb{R}^{3}}$, covariance ${\Sigma \in \mbb{S}_{++}^{3}}$, opacity ${\alpha \in \mbb{R}_{+}}$, and spherical harmonics (for view-dependent color) parameters. Like NeRFs, GS optimizes these parameters by minimizing the photometric error between rendered and ground-truth images, initialized from a sparse point cloud generally computed using structure-from-motion. Notably, GS employs a tile-based rasterization procedure to efficiently project the Gaussian primitives to the image plane, circumventing the expensive volumetric ray-marching procedure used by NeRFs, for faster training and rendering. Moreover, the explicit scene representation enables relatively easier theoretical analysis compared to NeRFs, in addition to providing more accurate depth estimation and mesh extraction.

\section{Visual-Geometry Semantics in Distilled Radiance Fields}
\label{sec:method}
We present our approach to distilling spatially-grounded semantics from vision foundation models into radiance fields.
We utilize the state-of-the-art Visual Geometric Grounded Transformer (VGGT) 
as the vision backbone, given its effectiveness across a broad range of geometric scene-understanding tasks, e.g., camera intrinsics/extrinsics estimation, multi-view depth estimation, dense point cloud reconstruction, and multi-frame point tracking.

\begin{figure}
    \centering
    \includegraphics[width=\linewidth]{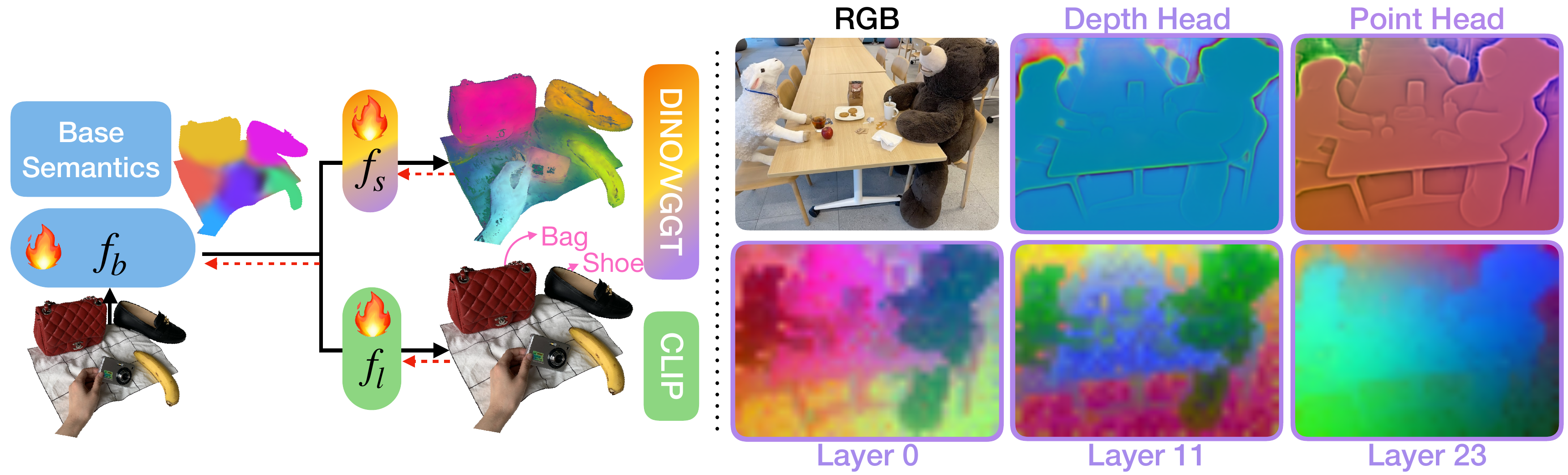}
    \caption{(left) \textbf{Semantics distillation architecture}, showing co-supervision of CLIP with DINO/VGGT via base semantics module. (right) \textbf{VGGT's semantic embeddings from different heads}, showing high-fidelity geometric content of the point head.}
    \label{fig:architecture}
\end{figure}

\p{Extracting Pretrained Visual-Geometry Semantic Features}
We extract ground-truth pretrained
semantic embeddings for each image from the depth and point heads of VGGT and its intermediate layers, which were trained for depth estimation and dense point cloud reconstruction, respectively. We visualize these semantic embeddings on the right side of \Cref{fig:architecture} using the first three principal components. We find that the point head produces features with the highest-fidelity spatial detail, compared to the depth head and other intermediate layers. 
Given a query image ${\mcal{I} \in \mbb{R}^{H \times W \times C}}$, the VGGT encoder outputs a semantic embedding ${f \in \mbb{R}^{H \times W \times d_{s}}}$, where ${d_{s} = 128}$. For computational efficiency, we preprocess the entire dataset prior to training.

\p{Distilling Semantics into Radiance Fields}
We learn a semantic field ${f_{s}: \mbb{R}^{3} \mapsto \mbb{R}^{d_{s}}}$, which maps a $3$D point $\mbf{x}$ to visual-geometry features $f_{s}(\mbf{x})$ alongside a semantic field ${f_{l}: \mbb{R}^{3} \mapsto \mbb{R}^{d_{l}}}$ that maps $3$D points to the shared image-language embedding space of CLIP. Note that $d_{s}$ and $d_{l}$ denote the dimension of the embedding space of VGGT and CLIP, respectively, which varies depending on the specific VGGT head and CLIP model, e.g.,
$128$ for VGGT's depth/point head and $768$ and $1024$ for CLIP's ViT and ResNet models.
We train a semantic field for CLIP features to enable downstream open-vocabulary tasks. 
For effective co-supervision of both semantic fields, the VGGT and CLIP semantic fields share the same hashgrid encodings (i.e., base semantics), associating their semantic embeddings with the same visual and geometric features, illustrated in the left side of \Cref{fig:architecture}.

Given a dataset $\mcal{D}$ of images and associated camera poses, we render images in the semantic space by back-projecting RGB images using the estimated depth from the radiance field to reconstruct a point cloud in the local camera frame. We subsequently compute the VGGT and CLIP semantic features for the constituent points with $f_{s}$ and $f_{l}$, respectively.

During training, we optimize the parameters of $f_{s}$ and $f_{l}$ simultaneously with the visual attributes of the radiance field using the loss function:
${\mcal{L} = \mcal{L}_{r} + \sum_{\mcal{I} \in \mcal{D}, c \in \{s, l\}} \Vert I_{f, c} - \hat{I}_{f, c} \Vert_{F}^{2} -  \sum_{\mcal{I} \in \mcal{D}, c \in \{s, l\}} \mathrm{csim}(I_{f, c} - \hat{I}_{f, c})}$
where $\mcal{L}_{r}$ denotes the RGB loss components of the base radiance field, $I_{f, c}$ and $\hat{I}_{f, c}$ denote the ground-truth and rendered semantic features, respectively, with $s$  and $l$ denoting the spatial and language components, and $\mathrm{csim}$ represents the cosine-similarity function. Although the \mbox{Frobenius-norm} term is not strictly required in 
the loss function,
we retain it to improve the numerical stability of the cosine-similarity term, which is undefined for vectors of zero norm. In our experiments, we apply the same distillation procedure to train DINO-embedded distilled radiance fields.

\p{Visualizing the Distilled Semantic Features}
We examine the content of the distilled semantic features extracted by the DINO and VGGT backbones via principal component analysis (PCA),%
resulting in three-dimensional features which we visualize as images.
Concretely, given a matrix ${\hat{I}_{f} \in \mbb{R}^{m \times d}}$ of \mbox{$d$-dimensional} semantic features, we compute the singular value decomposition (SVD) of $A$ to obtain the tuple ${(U, S, V)}$, such that: ${\hat{I}_{f} = U \Sigma V^{T}}$, where ${U \in \mbb{R}^{m \times p}}$, ${V \in \mbb{R}^{n \times p}}$, and ${\Sigma \in \mbb{R}^{p \times p}}$ denotes the left %
, right %
, and singular values, respectively. In practice, we compute the low-rank SVD for computational efficiency.
Subsequently, we map the semantic features in $I_{v}$ to the RGB image space using the first-three principal components via the transformation: ${I_{v} = \hat{I}_{f} V_{[:, :3}]}$, with ${I_{v} \in \mbb{R}^{m \times 3}}$, which is reshaped into a $3$-channel $2$D image for visualization (i.e., ${I_{v} \in \mbb{R}^{W \times H \times 3}}$). 

To quantify the geometric content of the distilled semantic features, we introduce the \emph{geometric fidelity factor} (GFF), which captures the edge information present in the semantic features relative to the physical scene, as determined by the RGB image. To do so, we apply the Sobel–Feldman operator \citep{duda1973pattern} to the RGB image, which approximates the gradient of the image intensity through simple convolutions with the image using $3$x$3$ kernels. The Sobel–Feldman operator suffices for this use-case, although more sophisticated gradient estimators could be applied in more complex problems. We apply hard-thresholding to the norm of the gradients to produce edges at varying resolutions.
We post-process the semantic image $I_{v}$ and RGB image to obtain the binary edge masks ${I_{e, \mathrm{sem}} \in \mbb{R}^{W \times H \times 3}}$ and ${I_{e, \mathrm{rgb}} \in \mbb{R}^{W \times H \times 3}}$, respectively.
After extracting edges from the semantic and RGB images, we compute the GFF using:
\begin{equation}
    \label{eq:GFF}
    \mathrm{GFF} \coloneqq {\sum_{(i, j, k)} I_{e, \mathrm{sem}}[i, j, k]}/{\sum_{(i, j, k)} I_{e, \mathrm{rgb}}[i, j, k]},
\end{equation}
representing the fraction of edges retained by the distilled features.
We examine the relative geometric content of the semantic features at different gradient thresholds in \Cref{subsec:eval-pca}.

\section{Semantic Localization}
\label{sec:method_segmentation}
The distilled semantic features enable open-vocabulary object localization within the radiance field given a query: e.g., ``find me a mug.'' For semantic localization, we compute the semantic embeddings of the language query $\phi_{\mathrm{query}}$ using CLIP and subsequently compute the cosine similarity between the query and all points in the radiance field to identify candidate matches. 
For increased robustness, we utilize the semantic relevancy score \citep{kerr2023lerf} given by:
${\nu(\phi_{\mathrm{query}}) = \min_{i} \frac{\exp(\phi_{\mathrm{query}} \cdot \phi_{f})}{\exp(\phi_{f} \cdot \phi_{\mathrm{canon}}^{i}) + \exp(\phi_{\mathrm{query}} \cdot \phi_{f})},}$
where $\phi_{f}$ represents the rendered semantic embeddings from the radiance field and $\phi_{\mathrm{canon}}$ represents the semantic embedding of \emph{canonical} prompts, i.e., generic or negative prompts to better distinguish between confident localization matches from non-confident ones. We use canonical prompts such as: ``object,'' ``stuff,'' and ``things.'' 
The semantic relevancy score can be viewed as the pairwise softmax over a set of positive and negative queries with respect to the rendered semantic embeddings.  
For conservative results, we take the minimum over all pairwise softmax distributions to define the semantic relevancy score.

To evaluate semantic localization accuracy, we map the ground-truth segmentation mask to the Euclidean space, assigning the value zero to pixels outside the map and one otherwise. Likewise, we normalize the semantic relevancy score generated by the radiance field at each view to values in ${[0, 1]}$ to obtain a relevancy mask. Afterwards, we map the ground-truth mask and the relevancy mask to the RGB space using a  colormap. Thereafter, we compute the objective localization accuracy using widely-used perceptual metrics, such as the structural similarity index measure (SSIM), learned perceptual image patch similarity (LPIPS), and peak-signal-to-noise ratio (PSNR).
We do not use metrics that require binary masks, e.g., mean intersection-over-union (mIoU), as these metrics would require the selection of a similarity threshold for each rendered semantic relevancy image. Identifying an optimal similarity threshold for DINO and VGGT features individually presents computational challenges, especially since the optimal values are generally view-dependent. Moreover, approximating the optimal thresholds could lead to confounding results.

\section{Inverting Radiance Fields}
\label{sec:inverting_radiance_fields}
We explore the application of visual-geometry semantics to the inversion of radiance fields. The inverse problem is particularly challenging compared to the well-posed forward problem of image rendering in radiance fields, especially without any simplifying assumptions. In fact, existing methods \citep{yen2021inerf, chen2025splat} struggle with camera pose estimation in radiance fields without a good initial guess. 
As a result, we introduce \algname, a novel algorithm for inverting radiance fields using distilled semantics for camera pose recovery without an initial guess. 
Moreover, \algname enables us to comprehensively evaluate the relative performance of spatially-grounded features in radiance field inversion problems. 
Using a semantics-conditioned inverse model, \algname directly maps semantic embeddings to a distribution over camera poses, which is subsequently refined to compute high-accuracy pose estimates using robust optimization, described in the following discussion.

\p{Learning an Inverse Model}
\algname learns a neural field ${p_{\psi}: \mbb{R}^{d} \mapsto \mcal{P}}$ which maps semantic (image) embeddings ${f(\mcal{I}) \in \mbb{R}^{d}}$ to a distribution over candidate poses, where $\mcal{P}$ denotes the space of valid distributions. For VGGT, we use the camera embeddings as input to $p_{\psi}$; whereas for DINO, we use the class token as input. We decompose the camera pose ${P \in \SE(3)}$ into its translation ${\mbf{t} \in \mbb{R}^{3}}$ and orientation ${\mbf{R} \in \SO(3)}$. Note that optimizing over the space of orientations is non-trivial, given that the orthogonality constraint in ${\SO(3)}$.
To circumvent this challenge, \algname parameterizes the camera orientation using the corresponding Lie algebra $\so(3)$, the vector space of three-dimensional skew-symmetric matrices. Leveraging the isomorphism between $\so(3)$ and 
$\mbb{R}^{3}$, we represent the camera rotation by ${\mbf{r} \in \mbb{R}^{3}}$. Note that we can construct a skew-symmetric matrix from $\mbf{r}$ and subsequently map elements of $\so(3)$ to $\SO(3)$ using the exponential map, i.e., ${\exp: \so(3) \mapsto \SO(3)}$.

We jointly optimize the parameters of the GMM and the visual and semantic attributes of the radiance field, detaching the gradients between both fields to simultaneously learn the forward and inverse maps of the radiance field without compromising visual fidelity.
Moreover, we make no additional assumptions beyond those made by the underlying radiance field and train \algname entirely on the same inputs as the radiance field using the mean-squared-error (MSE).

\p{Camera pose estimation}
Given a query image, we compute the semantic embedding of the image, which is mapped to $\mcal{P}$ using $p_{\psi}$. In general, the estimated camera pose is not always of sufficiently high accuracy. Consequently, we render images from the radiance field at the estimated coarse camera pose, using novel-view synthesis to generate an RGB-D image. Subsequently, we match image features from the query image to the rendered image, associated to a point-cloud in the radiance field through the rendered depth. Given the set of corresponding matches $\mcal{C}$, we solve the perspective-$n$-point (PnP) problem to refine the coarse pose estimates:
${(\hat{\mbf{t}}, \hat{\mbf{R}}) = \argmin_{\mbf{t} \in \mbb{R}, \mbf{R} \in \SO(3)} \sum_{(\mbf{p}, \mbf{q}) \in \mcal{C}} \Vert{\mbf{p} - \mbf{K} (\mbf{R} \mbf{q} + \mbf{t}) \Vert}_{2}^{2},}$
where ${(\mbf{p}, \mbf{q})}$ denote a corresponding pair with  homogeneous image coordinates $p$ (in the query image) and $3$D point $q$ (in the radiance field), $\mbf{K}$ denotes the camera intrinsic matrix, and $\hat{\mbf{t}}$ and $\hat{\mbf{R}}$ represent the estimated camera translation and rotation, respectively. For robustness to spurious correspondences, we solve 
the PnP problem
using RANSAC, mitigating the effects of outliers on the estimated pose. 

We evaluate the accuracy of the pose estimates using the $\SE(3)$ error, composed of the translation and rotation error between the ground-truth and estimated camera poses. We compute the translation error directly as the $\ell_{2}$ norm of the difference between the ground-truth and estimated translation and leverage the trace property of rotation matrices: ${\trace{\mbf{R}} = 1 + 2 \cos(\theta)}$ to compute the smallest rotation angle required to align the ground-truth and estimated camera orientations, where $\theta$ denotes the angle associated with rotation matrix $\mbf{R_{\delta}}$. Specifically, the absolute rotation error $\theta$ is given by: ${\theta = \arccos \left(\frac{\trace{\mbf{R_{\delta}}} - 1}{2} \right)}$, where $\mbf{R_{\delta} = \hat{\mbf{R}}^{\T}\mbf{R}_{\gt}}$ denotes the relative rotation matrix between the the ground-truth and estimated rotation matrices, $\hat{\mbf{R}}$ and $\mbf{R}_{\gt}$, respectively.

\section{Experiments}
\label{sec:evaluation}
We examine the performance of visual-geometry semantic features compared to visual-only features in distilled radiance fields.
Via extensive experiments, we explore the following questions, spanning the core applications of distilled radiance fields in robotics: 
(i) \emph{Do visual-geometry semantic features contain higher-fidelity spatial content?}
(ii) \emph{Does geometry-grounding improve semantic object localization?}
(iii) \emph{Can visual-geometry features enable higher-accuracy radiance field inversion?}
The full results for all datasets are provided in the Appendix.  

\subsection{Evaluation Setup}
\label{subsec:eval_setup}
We discuss the evaluation setup briefly, and provide additional details in \Cref{subsec:app_eval_setup}.
We evaluate the visual-only semantics from DINOv2
and DINOv3
, and visual-geometry features from VGGT
in nine scenes from three benchmark datasets, namely: \emph{Ramen}, \emph{Teatime}, and \emph{Waldo\_kitchen} in the LERF dataset \citep{kerr2023lerf}; \emph{Bed}, \emph{Covered Desk}, and \emph{Table} in the 3D-OVS dataset \citep{liu2023weakly}; and \emph{Office}, \emph{Kitchen}, and \emph{Drone} in the robotics dataset \citep{shorinwa2025siren}. For each scene, we train a semantic GS and NeRF representation and compute the following metrics across $100$ camera poses: geometric fidelity factor (GFF) metric for semantic content analysis (\Cref{sec:method}); SSIM, PSNR, and LPIPS for semantic object localization and radiance field inversion (\Cref{sec:method_segmentation}); and rotation and translation error in degrees and meters, respectively (\Cref{sec:inverting_radiance_fields}).

\subsection{Semantic Content of Distilled Features}
\label{subsec:eval-pca}
\begin{figure}[th]
    \centering
    \includegraphics[width=\linewidth]{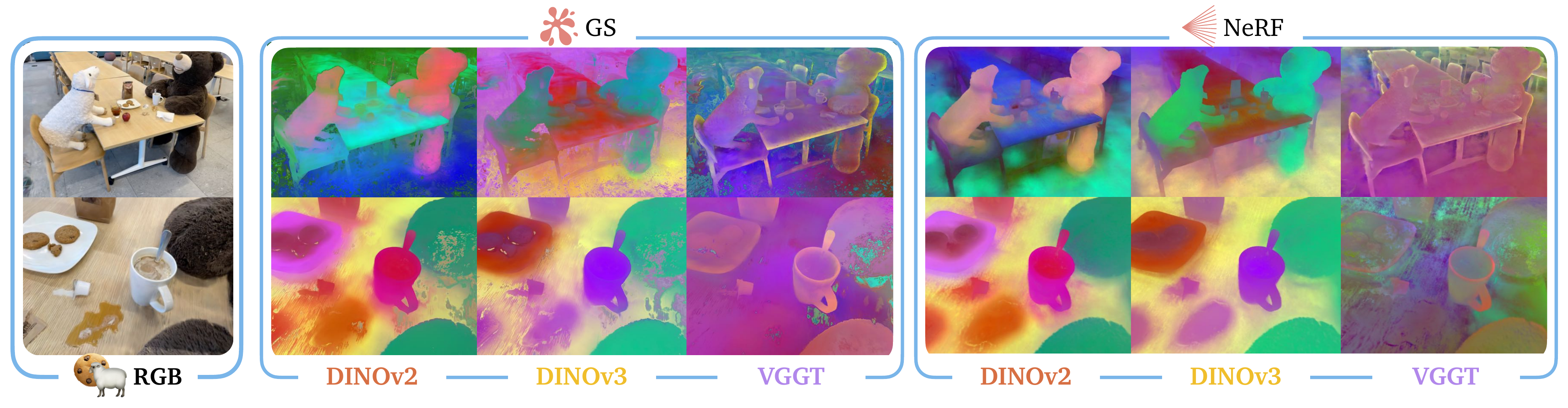}
    \caption{\textbf{Semantic content of distilled features.} Whereas visual-only features
    provide object-level  information, visual-geometry features 
    provide more 
    structural details, such as an object's contour.
    }
    \label{fig:pca_teatime}
\end{figure}

As described in \Cref{sec:method}, we project the distilled semantic features into a three-dimensional subspace using the first-three principal components to aid visualization.
\Cref{fig:pca_teatime} shows the PCA visualization of the semantic features in the \emph{Teatime} scene, highlighting the object-level composition of the scene. In the top row in \Cref{fig:pca_teatime}, we observe that the DINOv2 and DINOv3 features for the bear and the sheep are strongly distinct from the table and chairs, underscoring their focus on object-level decomposition. In contrast, VGGT features emphasize the geometric details of the scene, evidenced by the prominent edges of the bear, sheep, table, and chair, although some object-level features are visible. 
Likewise, in the bottom row, we see that VGGT highlights the structure (outline) of the mug and plate, unlike DINOv2 and DINOv3, although the DINO-based features provide more detailed object-level information. However, the wood grain on the table surface are more pronounced in the DINO-based features compared to those of VGGT.
These findings suggest that visual-geometry semantic features provide detailed geometric information compared to visual-only features; however, visual-only features may provide more consistent object-level semantics (e.g., object class).
We observe similar findings in the semantic content of other scenes, discussed in \Cref{sec:app_semantic_content}.

\begin{figure}[th]
    \centering
    \includegraphics[width=\linewidth]{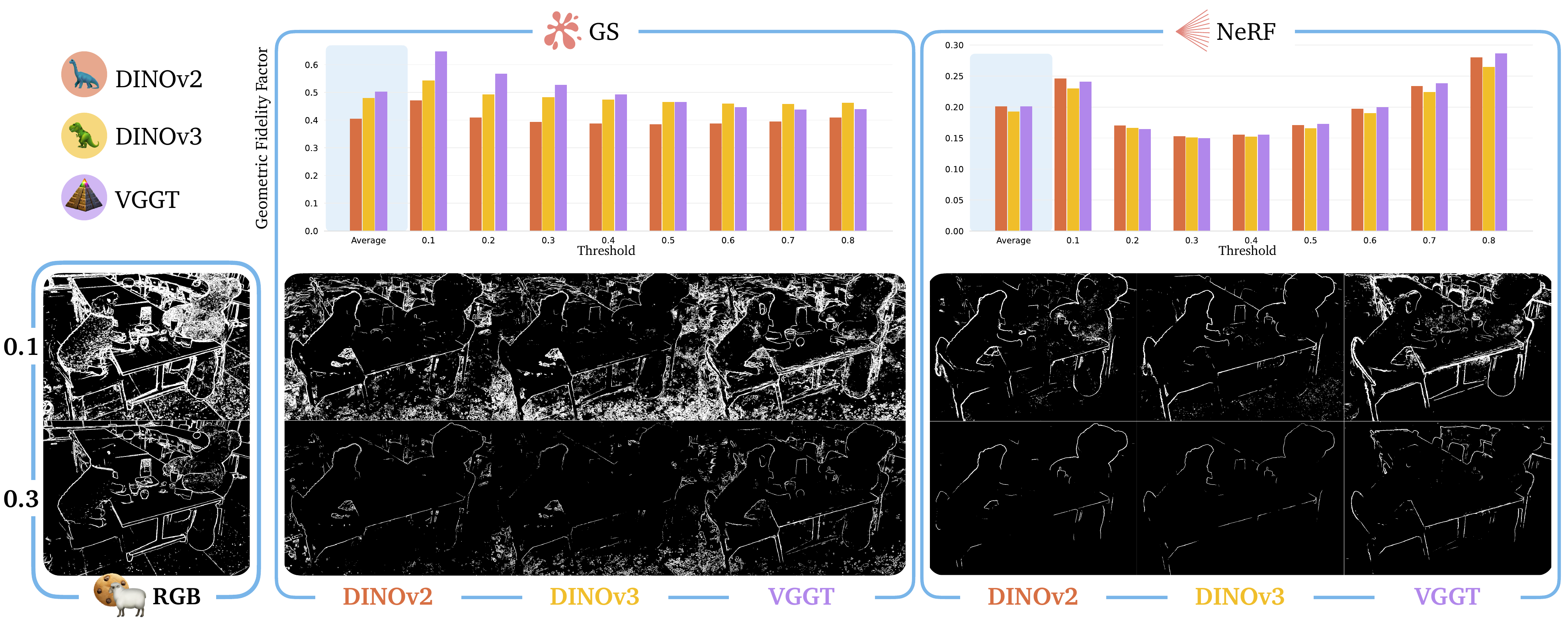}
    \caption{\textbf{Geometric fidelity factor (GFF) of visual-geometry and visual-only features.}
    VGGT's features contain prominent object edges, unlike visual-only semantic features. 
    }
    \label{fig:pca_edge_teatime}
\end{figure}

Next, we use the GFF (see \Cref{eq:GFF}) to quantitatively assess the geometric content of distilled features. We apply the Sobel-Feldman filter to the semantic images and extract the edges contained in these images at different resolutions, by varying the threshold of the edge gradient. In \Cref{fig:pca_edge_teatime}, we visualize the results for the \emph{Teatime} scene with thresholds of $0.1$ and $0.3$. Even at the lowest threshold of $0.1$, we observe more prominent geometry in the VGGT features.
Increasing the gradient threshold leads to an overall decrease in the number of edges contained in the spatially-grounded and visual-only features. However, VGGT still provides the most structural content. Surprisingly, we see that DINOv2 provides stronger geometric information compared to DINOv3 in this scene, although not in all scenes. 

We aggregate the quantitative results for all scenes and plot the GFF against gradient thresholds in \Cref{fig:pca_edge_teatime}. For GS, we see that VGGT's features have the most edges at lower gradient thresholds, with DINOv2's features having the least, consistent with our qualitative observations. Moreover, we observe that the GFF of DINOv2 and DINOv3 remains almost constant across different thresholds, suggesting a lack of diversity in their geometric content, unlike VGGT.
For NeRFs, the GFF remains similar across all semantic features, with DINOv3 having fewer edges at higher thresholds.

\subsection{Semantic Object Localization}
\label{subsec:eval-segmentation}
We examine the performance of spatially-grounded vs. visual-only features in semantic object localization using the procedure described in \Cref{sec:method_segmentation}. In each scene, we use CLIP to encode the natural-language queries and subsequently generate the continuous relevancy mask.
We use GroundingDINO \citep{liu2024grounding} and SAM-2 \citep{ravi2024sam} to annotate the ground-truth segmentation mask, used in computing the segmentation accuracy metrics: SSIM, PSNR, and LPIPS.

\begin{figure}[th]
    \centering
    \includegraphics[width=\linewidth]{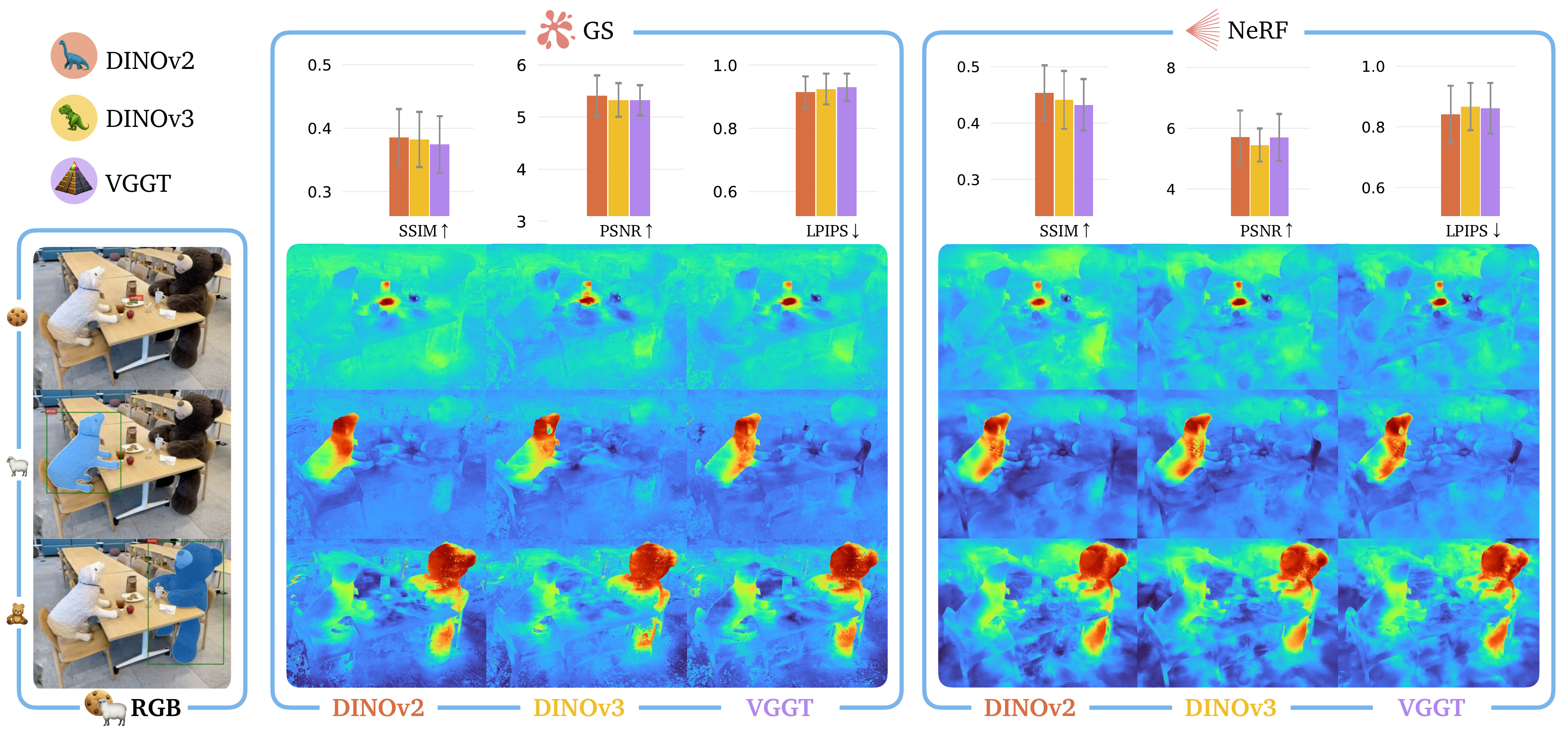}
    \caption{\textbf{Semantic object localization.} Both visual-only features (DINOv2/DINOv3) and visual-geometry features (VGGT) achieve similar localization accuracies (\emph{Teatime} scene visuals). 
    }
    \label{fig:semantic_localization}
\end{figure}

\Cref{fig:semantic_localization} summarizes our results.
We find no significant difference in the localization accuracy of visual-only vs. visual-geometry features across GS and NeRF, suggesting that both semantic features are effective in co-supervising CLIP for open-vocabulary localization. However, we observe marginal degradation in performance with geometry-grounded features (VGGT). %
In addition, we visualize the ground-truth RGB and segmentation mask and the relevancy masks in the \emph{Teatime} scene, highlighting the effectiveness of both kinds of semantic features in localizing the cookies, sheep, and bear. We provide additional results in \Cref{sec:app_semantic_localization}.

\subsection{Inverting Radiance Fields}
\begin{figure}[th]
    \centering
    \includegraphics[width=\linewidth]{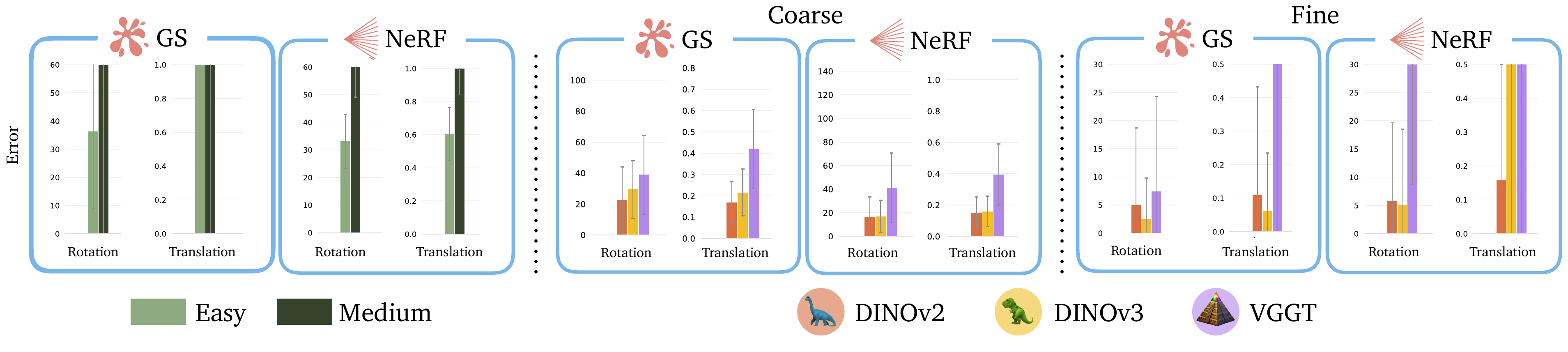}
    \caption{
    \textbf{Inverting radiance field.} (Left) Existing RF inversion methods struggle in  the absence of a good initial guess. (Center) In contrast, \algname computes better coarse pose estimates using the distilled semantics. (Right) Via photometric optimization, \algname refine the coarse estimates for higher accuracy.}
    \label{fig:rf_inversion}
\end{figure}

We evaluate the accuracy of visual-only and spatially-grounded features in radiance field inversion. Surprisingly, we find that visual-geometry features underperform visual-only features, summarized in \Cref{fig:rf_inversion}. Specifically, in the coarse pose estimation phase which significantly relies on semantics, DINOv2 achieves the lowest rotation and translation errors, while VGGT computes the least accurate pose estimates. 
These results suggest that DINOv2 features might be better suited for coarse pose estimation compared to VGGT features, despite the geometry-grounding procedure. Consequently, our findings indicate that existing methods for geometry-grounding may degrade the versatility of semantic features as general-purpose image features, constituting an interesting area for future work.

Further, we compare \algname to existing baseline methods for radiance field inversion. Particularly, we compare DINOv2-based \algname with \citep{chen2025splat} and \citep{yen2021inerf} for pose estimation in GS and NeRFs, respectively. Since the baselines require an initial guess, we assess the performance of the baselines across two initialization domains, defined by the magnitude of the initial rotation and translation error, $R_{\mathrm{err}}$ and $T_{\mathrm{err}}$, respectively:
(ii) low initial error with ${R_{\mathrm{err}} = 30\deg}$, ${T_{\mathrm{err}} = 0.5\meter}$, and (iii) medium initial error with ${R_{\mathrm{err}} = 100\deg}$, ${T_{\mathrm{err}} = 1\meter}$. We reiterate that \algname does not utilize any initial guess.

\Cref{fig:rf_inversion} summarizes the results. We observe that the baselines struggle without a good initial guess. Unlike these methods, \algname computes more accurate pose estimates using semantics in the coarse phase, without any initial guess. Moreover, via photometric optimization, \algname improves the accuracy of the coarse estimates. However, we note that the success of fine inversion depends on the relative error magnitude of the coarse pose estimates. Particularly, DINOv2 generally achieves the highest success rate in the fine pose estimation phase, primarily due to its higher-accuracy coarse pose estimates. We show the unweighted mean and standard deviation of the errors of the fine pose estimates in \Cref{fig:rf_inversion}.

\section{Conclusion}
\label{sec:conclusion}
We explore the relative performance of visual-geometry semantic features compared to visual-only features in distilled radiance fields, across three critical areas for state-of-the-art robot policies. First, our work examines the semantic content of these image features, finding that spatially-grounded features provide more prominent geometric detail compared to visual-only features, which could be useful in fine-grained robotics task, e.g., dexterous manipulation. Second, we evaluate the accuracy achieved by these features in open-vocabulary semantic localization, and observe no significant difference in their performance. Third, we derive a novel method for inverting radiance fields and compare the performance of visual-only and visual-geometry features in this task. We demonstrate the effectiveness of our method compared to existing baselines, without requiring an initial guess, unlike the baselines. Moreover, we find that visual-only features outperform spatially-grounded features in radiance field inversion.

\section{Limitations and Future Work}
\label{sec:limitations_future_work}
\p{Self-Supervised Geometry-Grounding}
Our findings suggest that existing solutions for geometry-grounding may impair the versatility of pretrained features, e.g., in radiance field inversion. We believe that this limitation may be due to the fully-supervised approach used by prior work. Future work will explore self-supervised approaches for spatial-grounding to eliminate inductive biases, improve adaptability, and enable larger-scale pre-training. 

\p{Synergy between Geometry and Vision}
In addition, our experiments revealed that visual-geometry semantics did not improve the semantic object localization accuracy, despite its more significant structural content, suggesting the lack of sufficient synergy between the geometric and visual contents. Future work will introduce more effective strategies for establishing synergy between the geometry-oriented and visual-oriented semantic features for more robust scene understanding.

\p{Efficient Inference}
Existing geometry-grounded vision backbones require notable compute overhead compared to ungrounded backbones, amplified by the absence of lightweight variants. Future work will examine more efficient architectures for spatially-grounded vision backbones to enable their use in real-time applications, e.g., in robot manipulation.

\section*{Acknowledgments}
The authors were partially supported by the NSF CAREER Award \#2044149, the Office of Naval Research (N00014-23-1-2148), and a Sloan Fellowship.

\clearpage

\bibliographystyle{plainnat}
\bibliography{references.bib}

\clearpage
\beginappendix{
    \section{Appendix}
\label{sec:appendix}

\subsection{Preliminaries}
\label{sec:app_preliminaries}
We discuss semantic distillation in radiance fields.

\p{Semantic Distillation}
Prior work \citep{zhi2021place, kobayashi2022decomposing} has explored distilling semantic information from foundation models into NeRFs and GS for open-vocabulary semantic segmentation, object localization, and scene-editing. We limit our discussion to distillation methods that learn a neural field \citep{kerr2023lerf}, 
due to their generality to both NeRFs and GS and faster training/inference speeds \citep{shorinwa2024fast}. Neural semantic distillation methods train a semantic field ${f: \mbb{R}^{3} \mapsto \mbb{R}^{d}}$ mapping a $3$D point to a $d$-dimensional semantic embedding. In general, these methods leverage hashgrid encodings \citep{muller2022instant} to learn varying resolutions of semantic detail. The semantic field is trained simultaneously with the radiance field with direct supervision by the ground-truth embeddings from the teacher network, e.g., CLIP/DINO.

\subsection{Evaluation Setup}
\label{subsec:app_eval_setup}
We provide additional details on the task and evaluation setup.

\p{Datasets}
We consider nine scenes across three benchmark datasets in open-vocabulary semantic localization in radiance fields and a robotics-centric dataset. Specifically, we evaluate the distilled semantic features in the \emph{Ramen}, \emph{Teatime}, and \emph{Waldo\_kitchen} scenes in the LERF dataset \citep{kerr2023lerf}; the \emph{Bed}, \emph{Covered Desk}, and \emph{Table} scenes in the 3D-OVS dataset \citep{liu2023weakly}; and the \emph{Office}, \emph{Kitchen}, and \emph{Drone} scenes in the robotics dataset \citep{shorinwa2025siren}. While the 3D-OVS dataset is limited to primarily small-scale scenes, the LERF and robotics datasets provide broader coverage over both relatively small-scale and large-scale scenes. Moreover, whereas the LERF and 3D-OVS datasets were primarily collected using a smartphone camera, the robotics dataset was collected using cameras onboard a drone and quadruped. We represent each dataset using a unique icon in the subsequent figures.

\p{Implementation}
We train a GS and NeRF representation per scene using the camera poses provided with the datasets for $30000$ iterations within the Nerfstudio framework \citep{tancik2023nerfstudio}. To extract visual-geometry semantics, we use the official VGGT checkpoints provided by the authors of \citep{wang2025vggt}. Likewise, we extract visual-only semantics from DINOv2 \citep{oquab2023dinov2} and DINOv3 \citep{simeoni2025dinov3} using their official checkpoints and use the CLIP ResNet model RN50x64 for language semantics. We train the radiance fields on an Nvidia L40 GPU with 48GB VRAM, although we emphasize that in many cases, the GPU memory is not the main compute bottleneck in our implementation. In fact, preliminary experiments were conducted on an Nvidia GeForce RTX 4090 GPU with 24GB VRAM.

\p{Metrics}
We evaluate the relative performance between spatially-grounded and visual-only semantics using task-specific metrics. For semantic content analysis, we use the geometric fidelity factor (GFF) metric (see \Cref{sec:method}).  For semantic object localization and radiance field inversion, we use the perceptual metrics SSIM, PSNR, and LPIPS (see \Cref{sec:method_segmentation}), and the $\SE(3)$ error with rotation in degrees and translation in meters (see \Cref{sec:inverting_radiance_fields}), respectively.
We compute these metrics across $100$ novel (unseen) camera poses in most scenes. In the 3D-OVS dataset, we use the training camera poses due to the limited scene coverage.

\subsection{Additional Semantic Content Results}
\label{sec:app_semantic_content}
Below, we include visualizations for the PCA and segmentation experiments, sampling one or two camera views from each scene.

\begin{figure}[H]
    \centering
    \includegraphics[page=1, width=\linewidth]{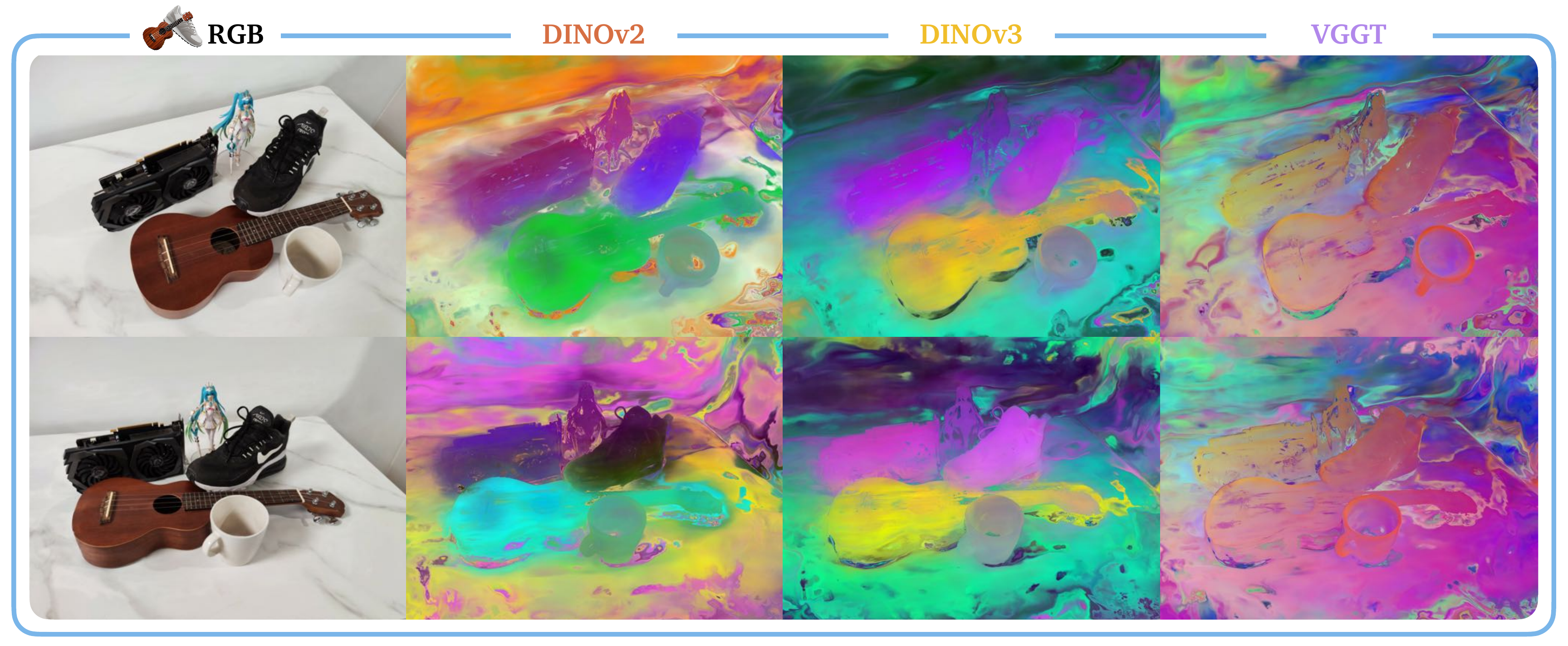}
    \caption{Semantic content of distilled features visualized for GS for the 3D-OVS table scene.}
    \label{fig:gs_pca_3dovs_table}
\end{figure}

\begin{figure}[H]
    \centering
    \includegraphics[page=2, width=\linewidth]{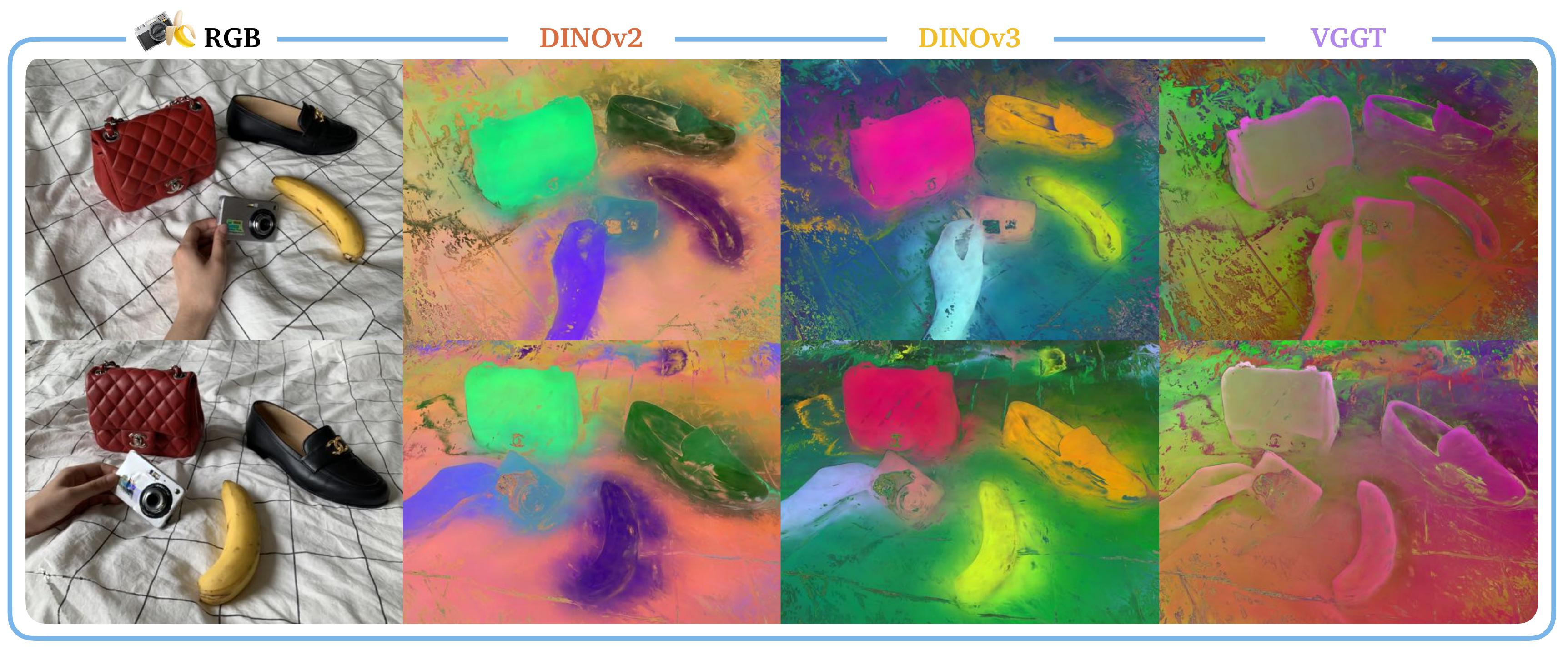}
    \caption{Semantic content of distilled features visualized for GS for the 3D-OVS bed scene.}
    \label{fig:gs_pca_3dovs_bed}
\end{figure}

\begin{figure}[H]
    \centering
    \includegraphics[page=3, width=\linewidth]{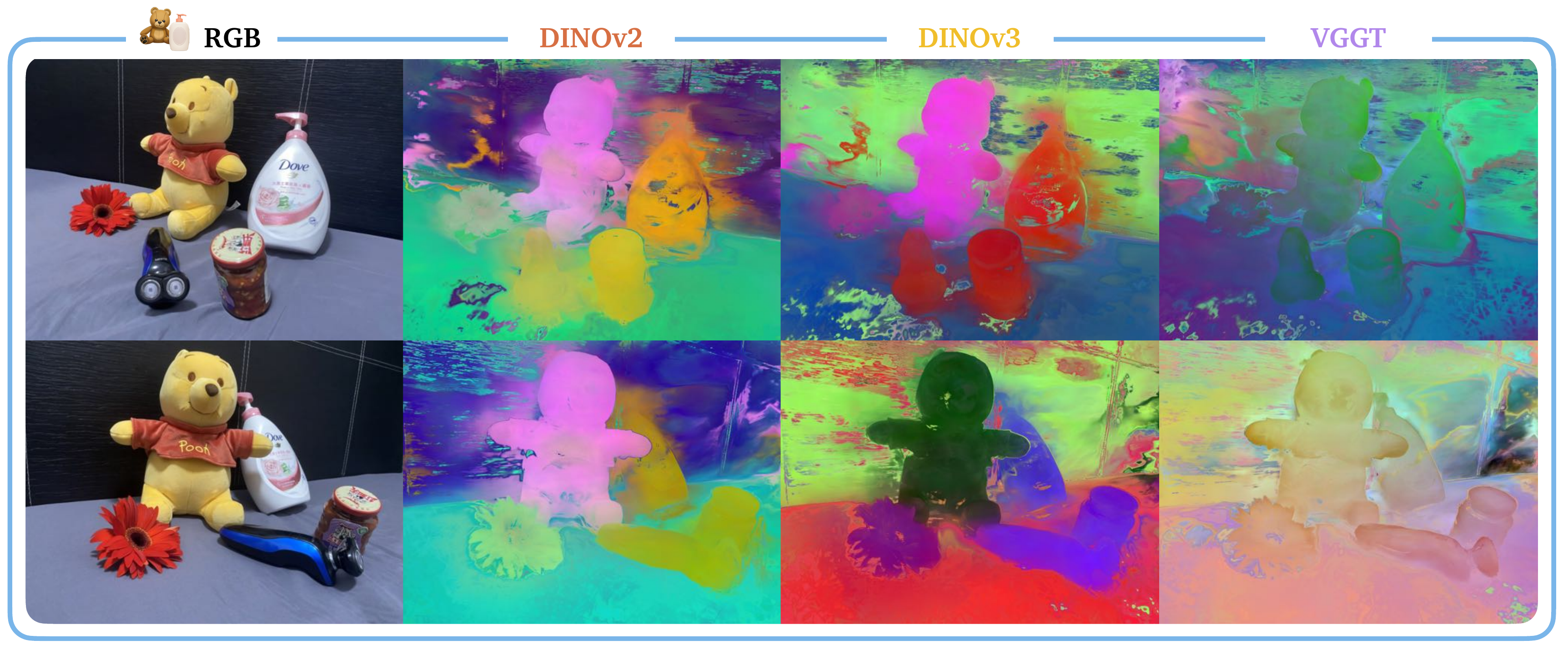}
    \caption{Semantic content of distilled features visualized for GS for the 3D-OVS covered desk scene.}
    \label{fig:gs_pca_3dovs_desk}
\end{figure}

\begin{figure}[H]
    \centering
    \includegraphics[page=1, width=\linewidth]{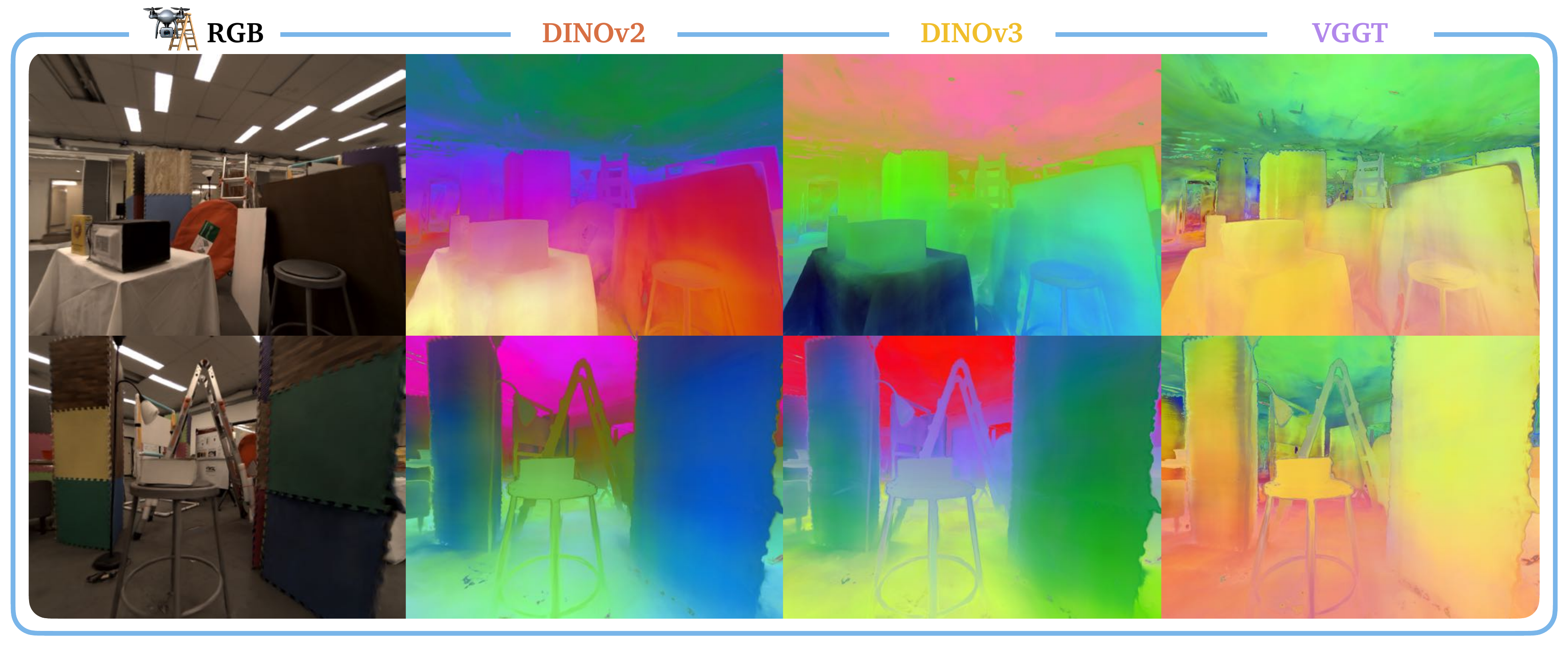}
    \caption{Semantic content of distilled features visualized for GS for the drone scene.}
    \label{fig:gs_pca_robot_drone}
\end{figure}

\begin{figure}[H]
    \centering
    \includegraphics[page=2, width=\linewidth]{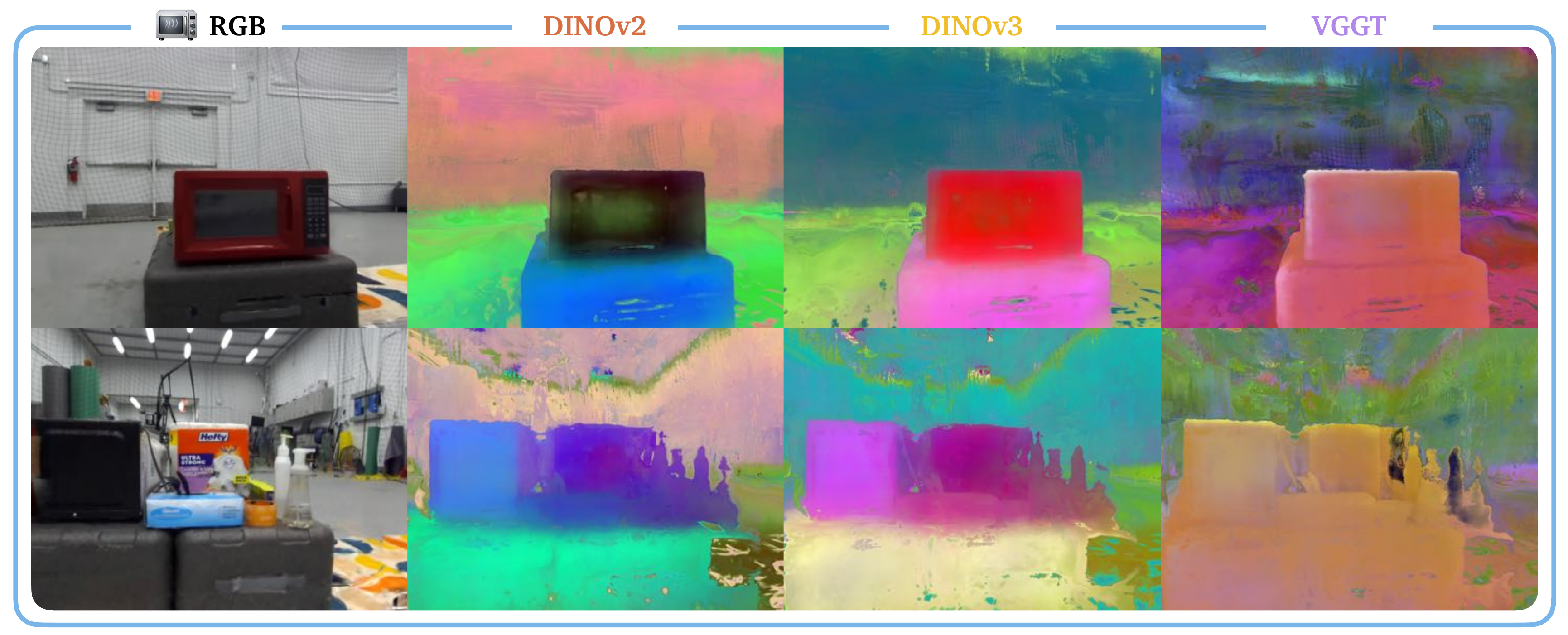}
    \caption{Semantic content of distilled features visualized for GS for the quadruped kitchen scene.}
    \label{fig:gs_pca_robot_kitchen}
\end{figure}

\begin{figure}[H]
    \centering
    \includegraphics[page=3, width=\linewidth]{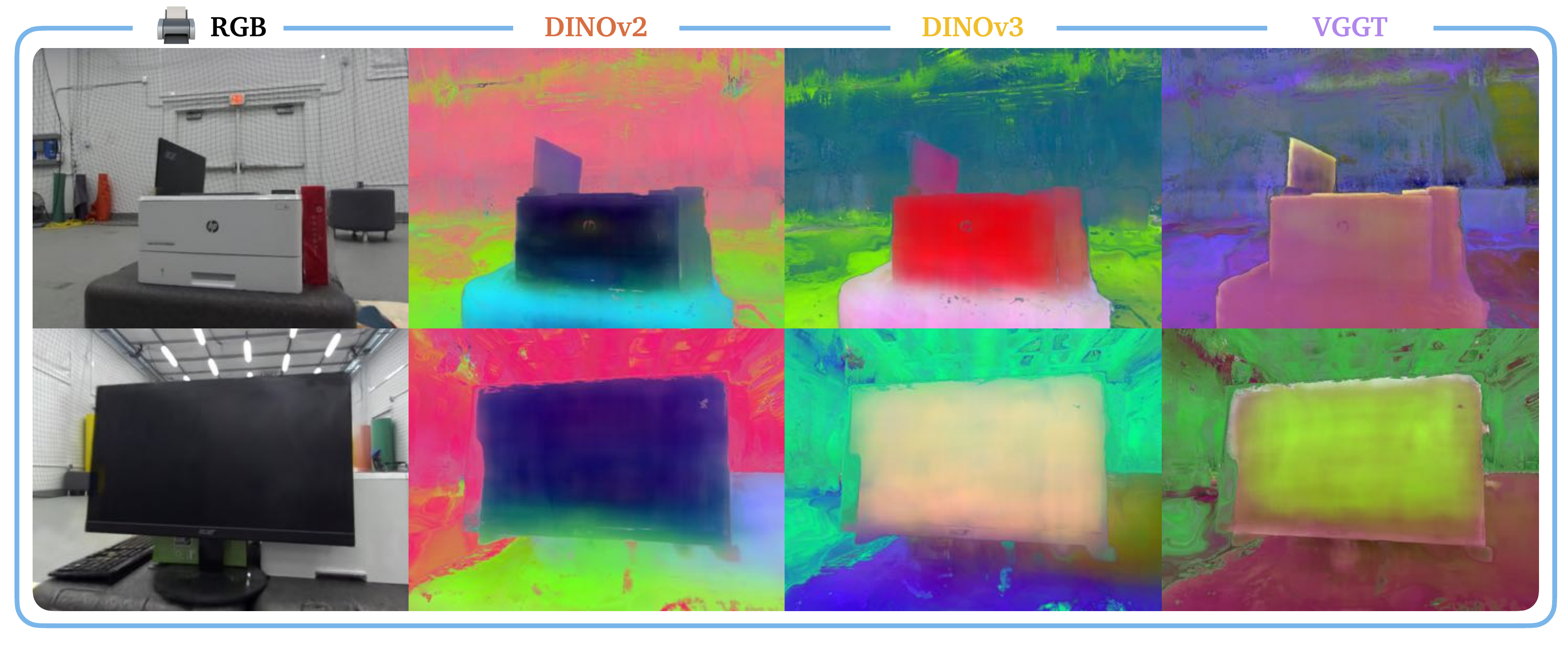}
    \caption{Semantic content of distilled features visualized for GS for the quadruped office scene.}
    \label{fig:gs_pca_robot_office}
\end{figure}

\begin{figure}[H]
    \centering
    \includegraphics[page=1, width=\linewidth]{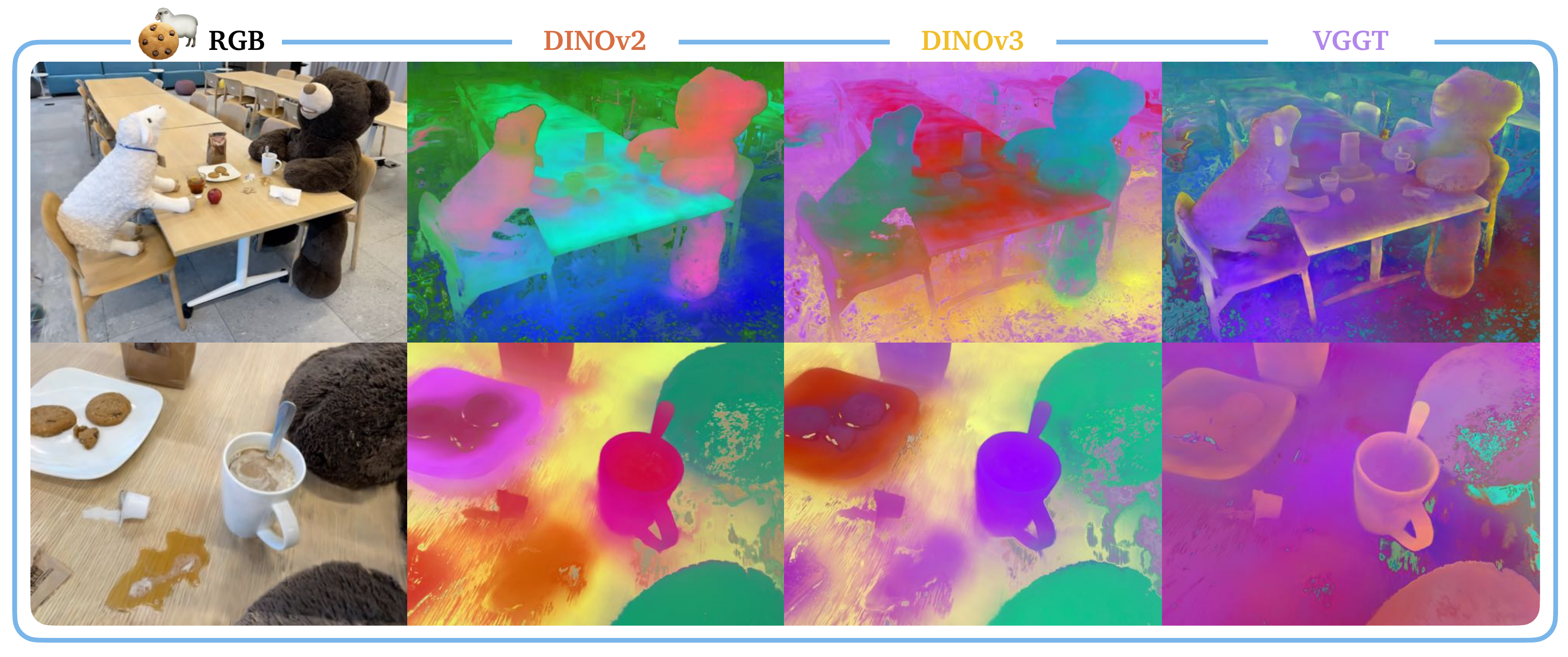}
    \caption{Semantic content of distilled features visualized for GS for the LERF teatime scene.}
    \label{fig:gs_pca_lerf_teatime}
\end{figure}

\begin{figure}[H]
    \centering
    \includegraphics[page=2, width=\linewidth]{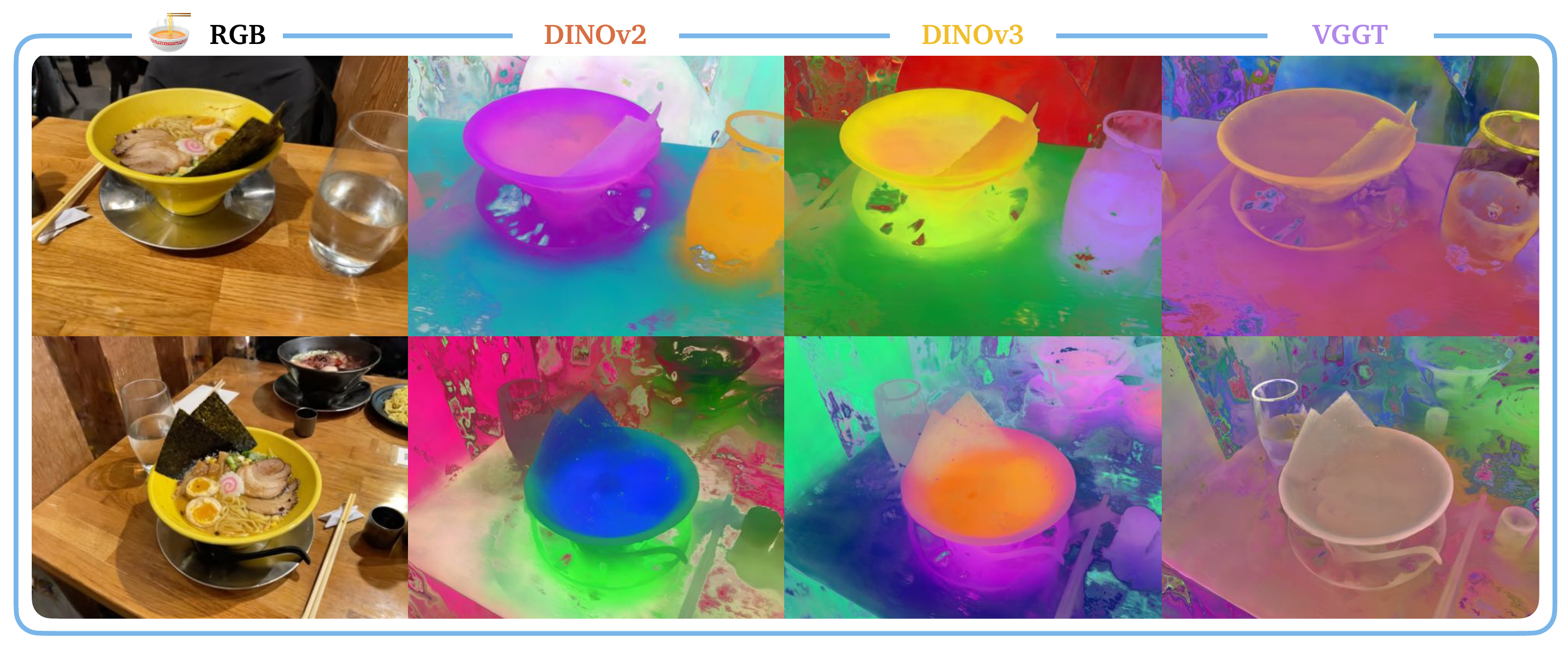}
    \caption{Semantic content of distilled features visualized for GS for the LERF ramen scene.}
    \label{fig:gs_pca_lerf_ramen}
\end{figure}

\begin{figure}[H]
    \centering
    \includegraphics[page=3, width=\linewidth]{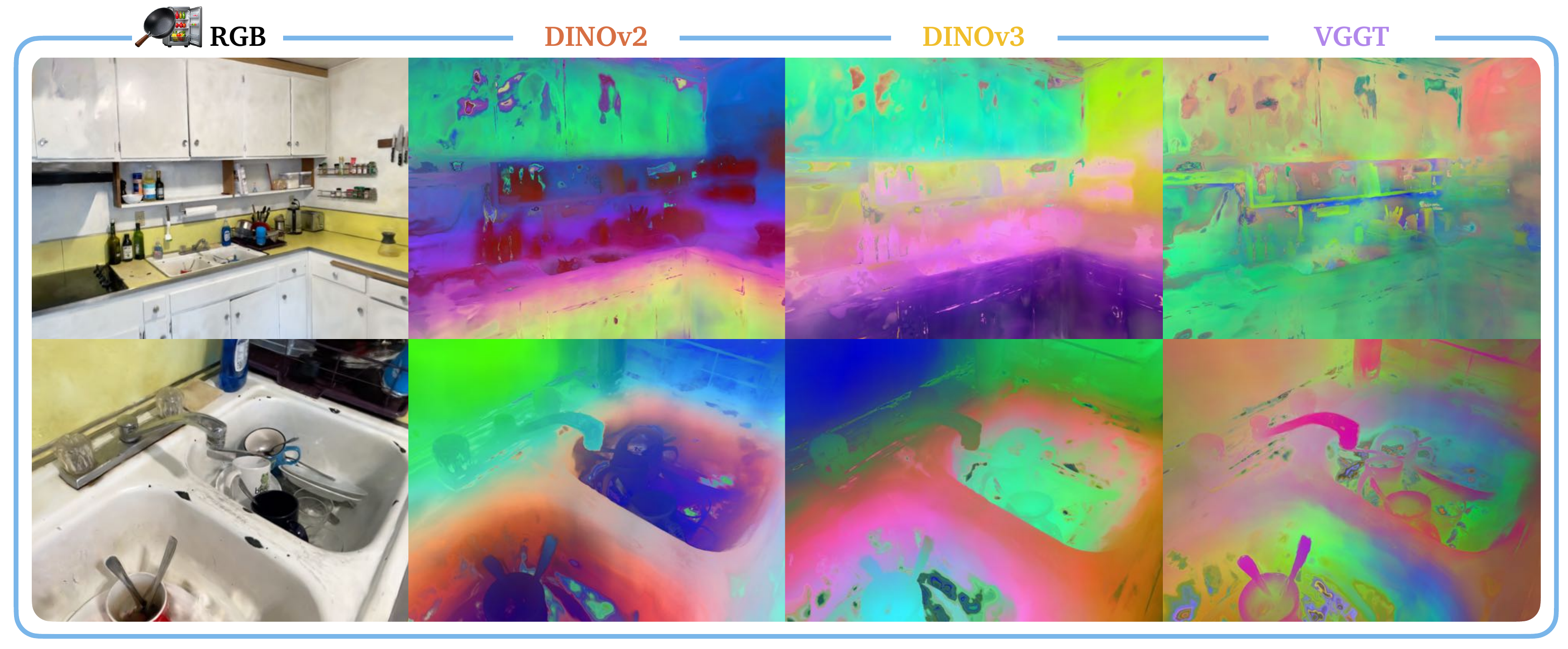}
    \caption{Semantic content of distilled features visualized for GS for the LERF kitchen scene.}
    \label{fig:gs_pca_lerf_kitchen}
\end{figure}

\begin{figure}[H]
    \centering
    \includegraphics[page=1, width=\linewidth]{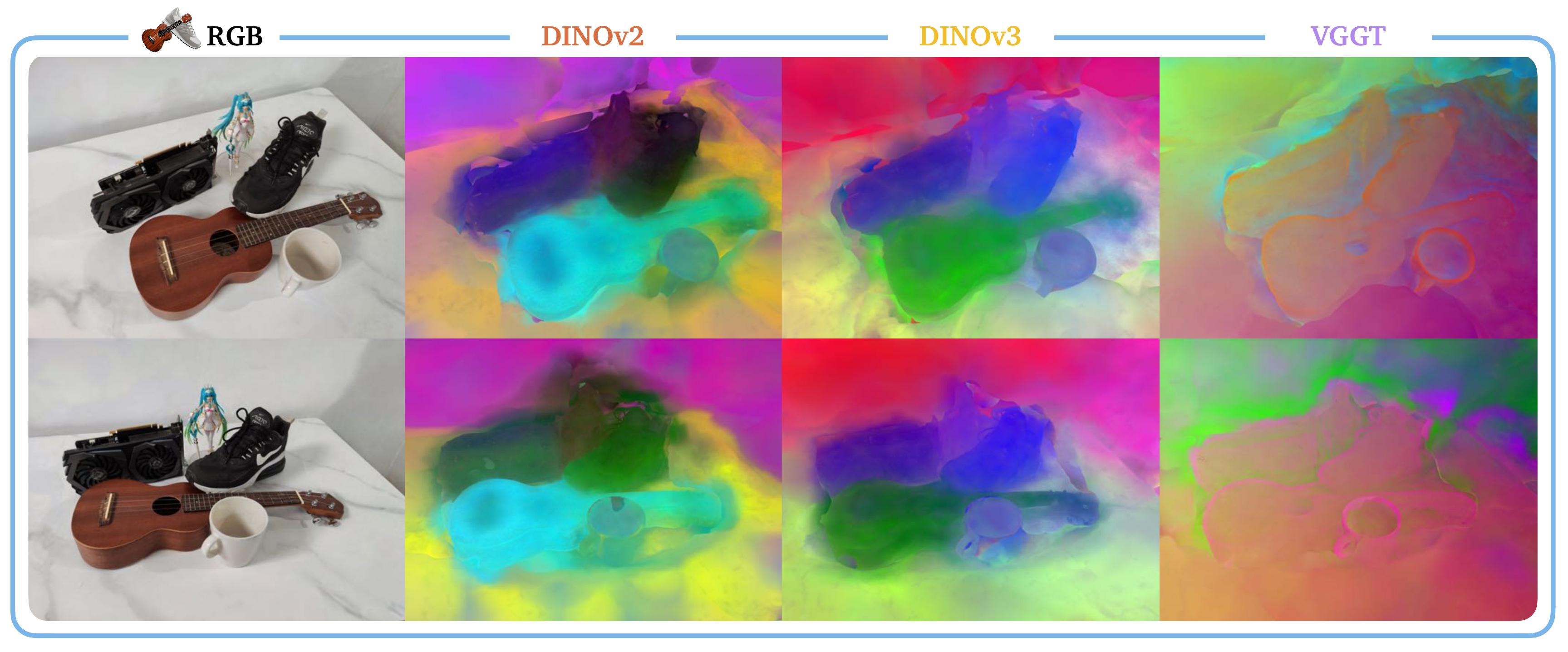}
    \caption{Semantic content of distilled features visualized for NERF for the 3D-OVS table scene.}
    \label{fig:nerf_pca_3dovs_table}
\end{figure}

\begin{figure}[H]
    \centering
    \includegraphics[page=2, width=\linewidth]{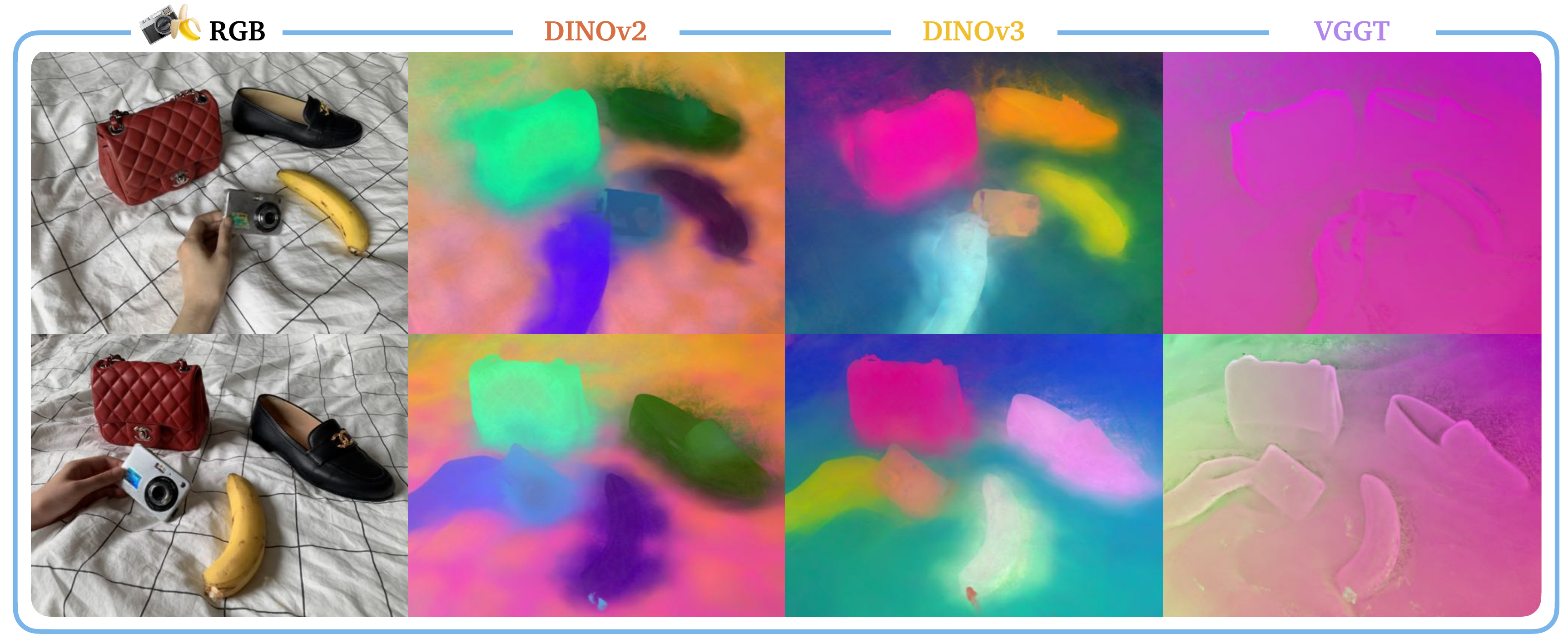}
    \caption{Semantic content of distilled features visualized for NERF for the 3D-OVS bed scene.}
    \label{fig:nerf_pca_3dovs_bed}
\end{figure}

\begin{figure}[H]
    \centering
    \includegraphics[page=3, width=\linewidth]{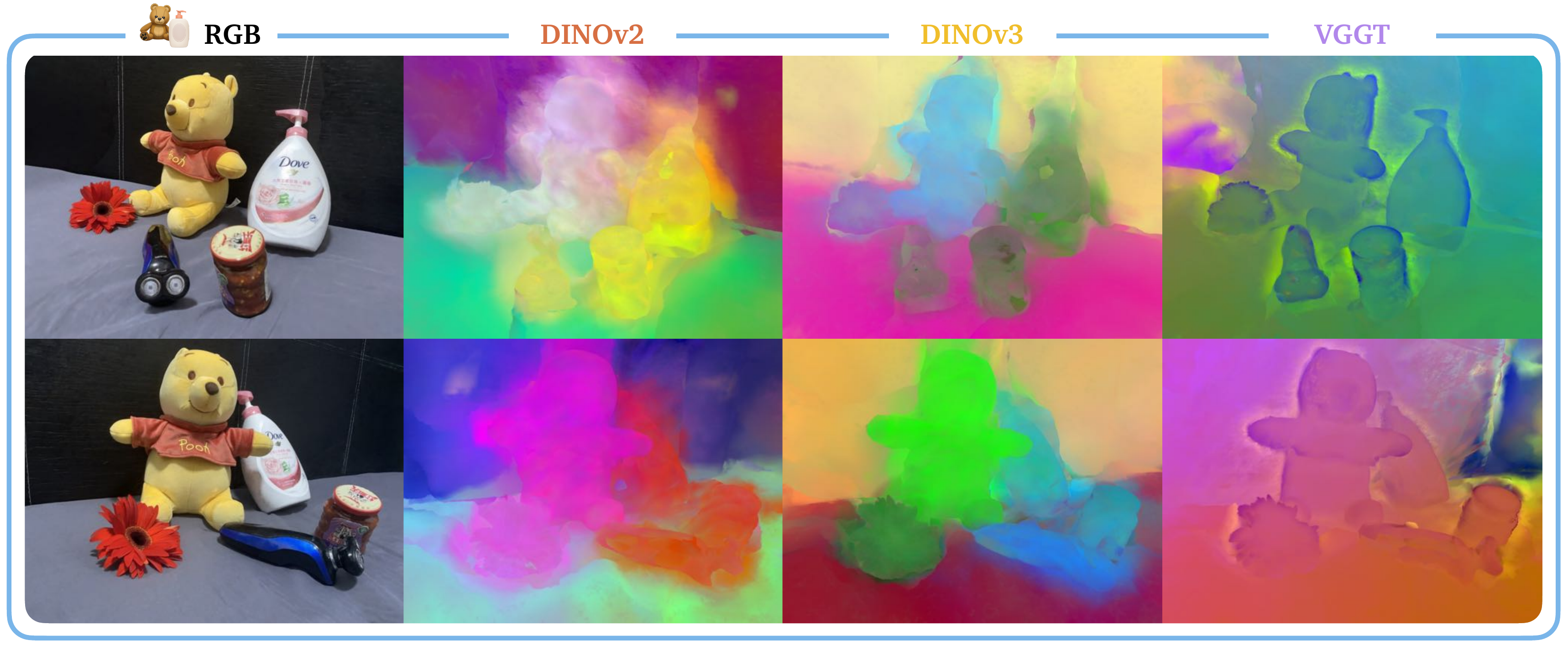}
    \caption{Semantic content of distilled features visualized for NERF for the 3D-OVS covered desk scene.}
    \label{fig:nerf_pca_3dovs_desk}
\end{figure}

\begin{figure}[H]
    \centering
    \includegraphics[page=1, width=\linewidth]{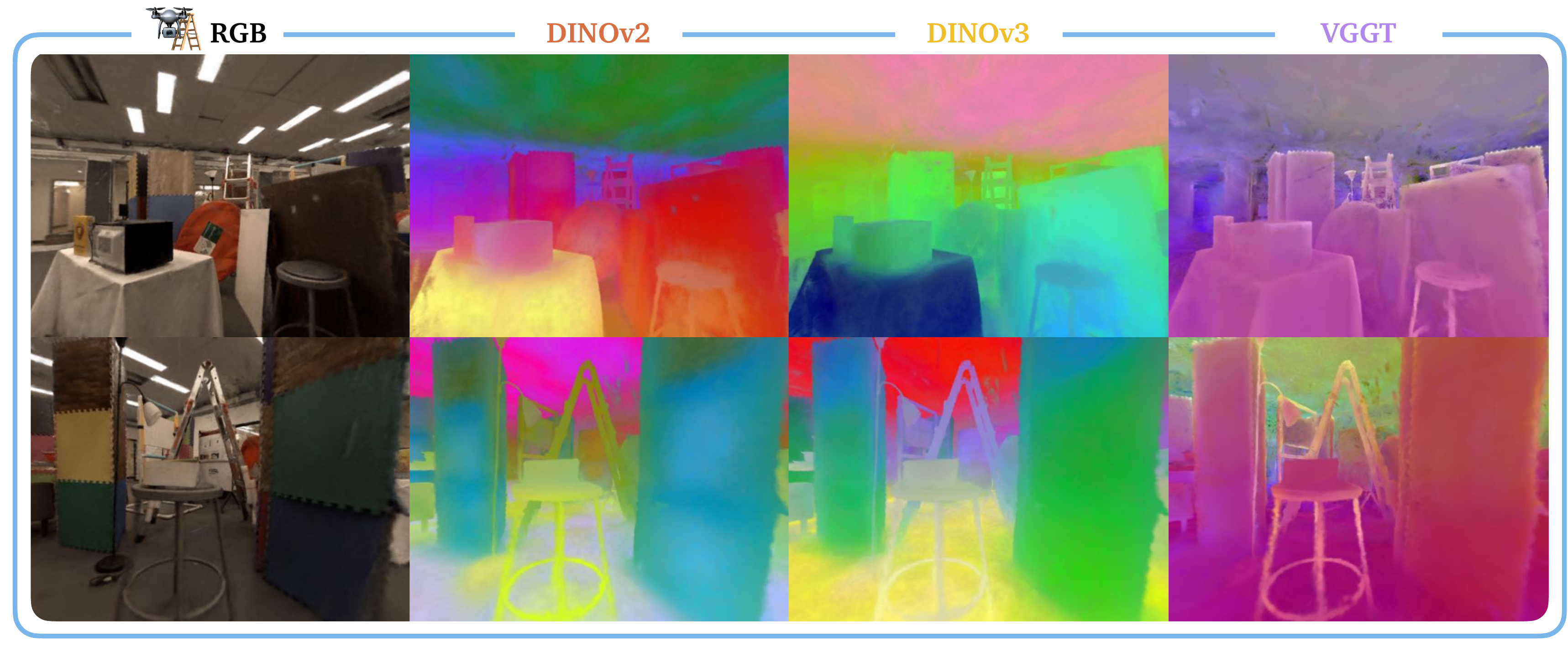}
    \caption{Semantic content of distilled features visualized for NERF for the drone scene.}
    \label{fig:nerf_pca_robot_drone}
\end{figure}

\begin{figure}[H]
    \centering
    \includegraphics[page=2, width=\linewidth]{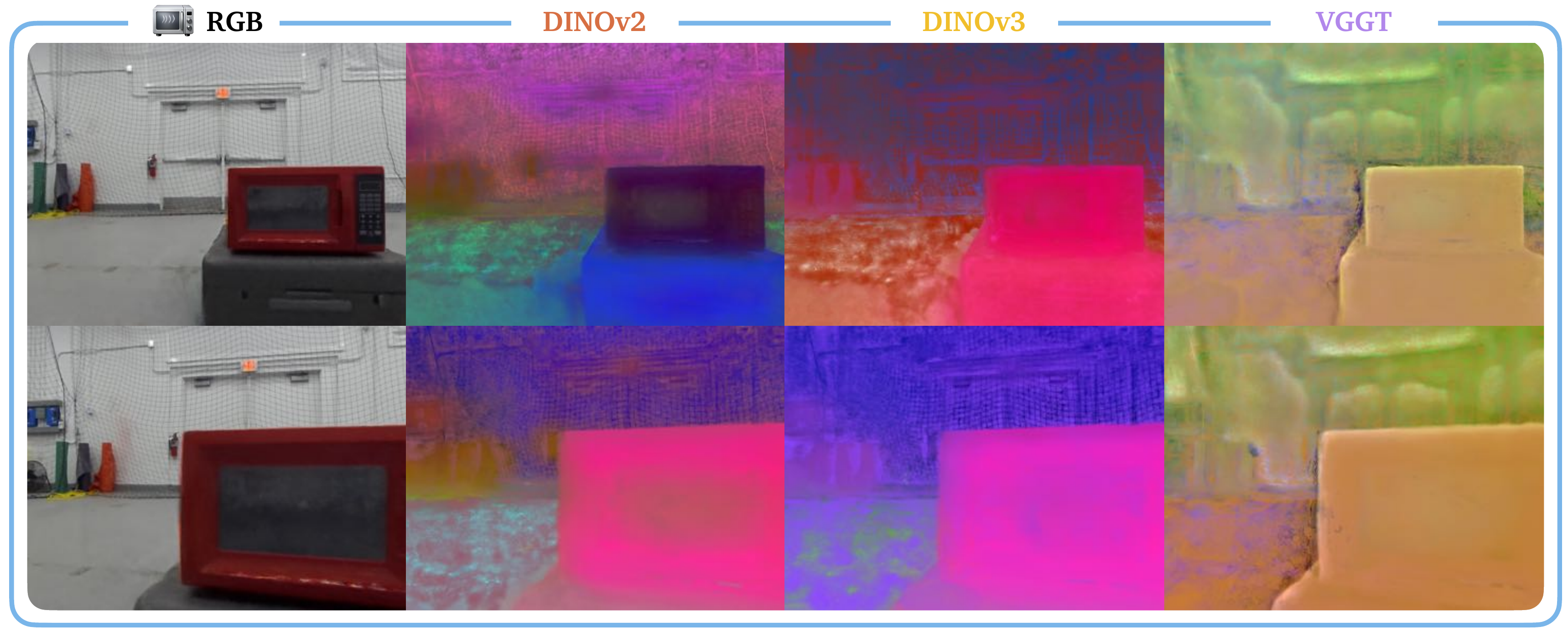}
    \caption{Semantic content of distilled features visualized for NERF for the quadruped kitchen scene.}
    \label{fig:nerf_pca_robot_kitchen}
\end{figure}

\begin{figure}[H]
    \centering
    \includegraphics[page=3, width=\linewidth]{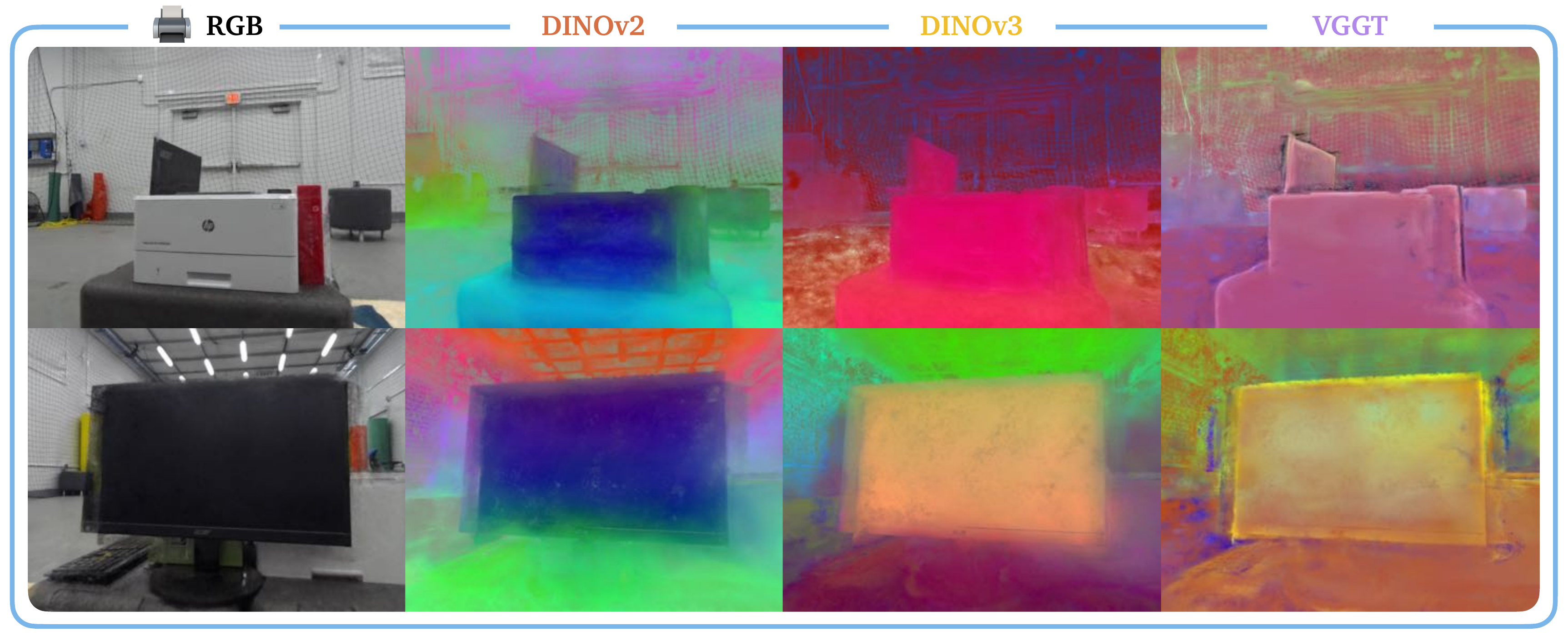}
    \caption{Semantic content of distilled features visualized for NERF for the quadruped office scene.}
    \label{fig:nerf_pca_robot_office}
\end{figure}

\begin{figure}[H]
    \centering
    \includegraphics[page=1, width=\linewidth]{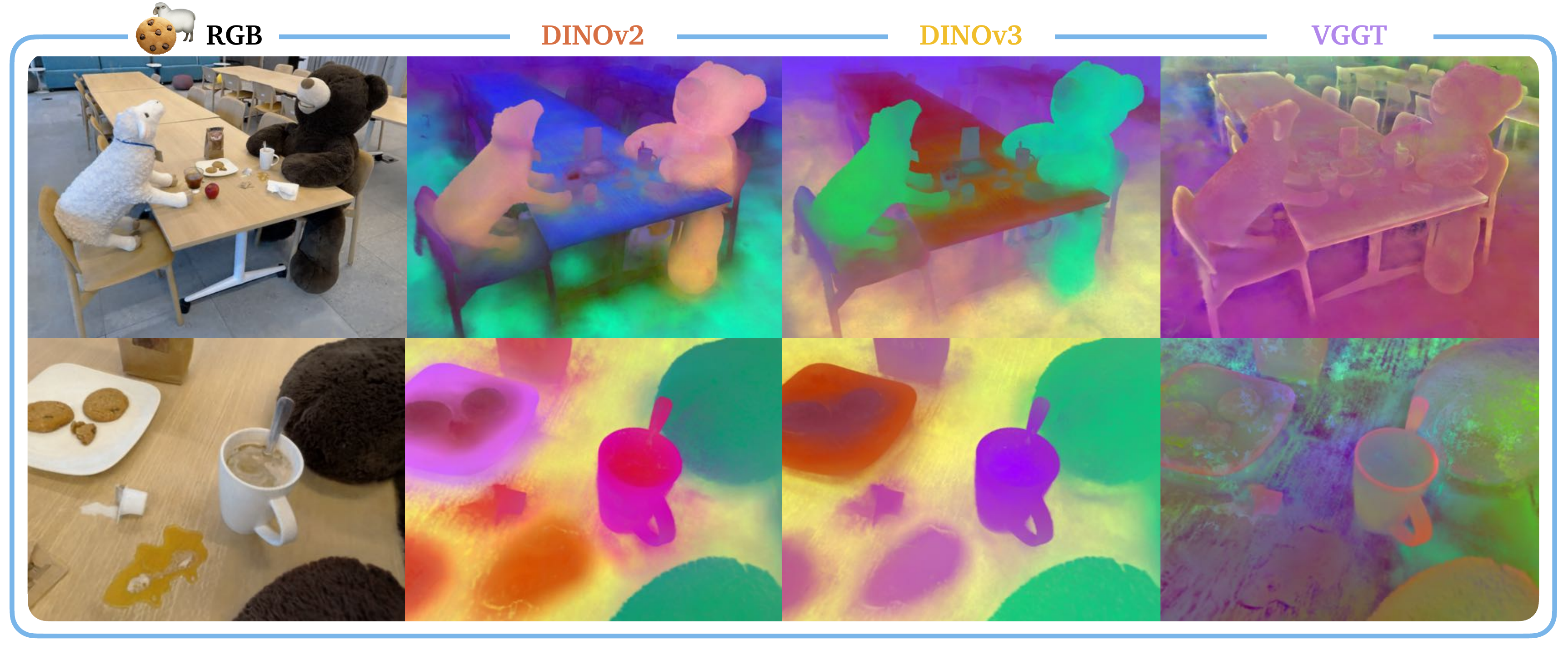}
    \caption{Semantic content of distilled features visualized for NERF for the LERF teatime scene.}
    \label{fig:nerf_pca_lerf_teatime}
\end{figure}

\begin{figure}[H]
    \centering
    \includegraphics[page=2, width=\linewidth]{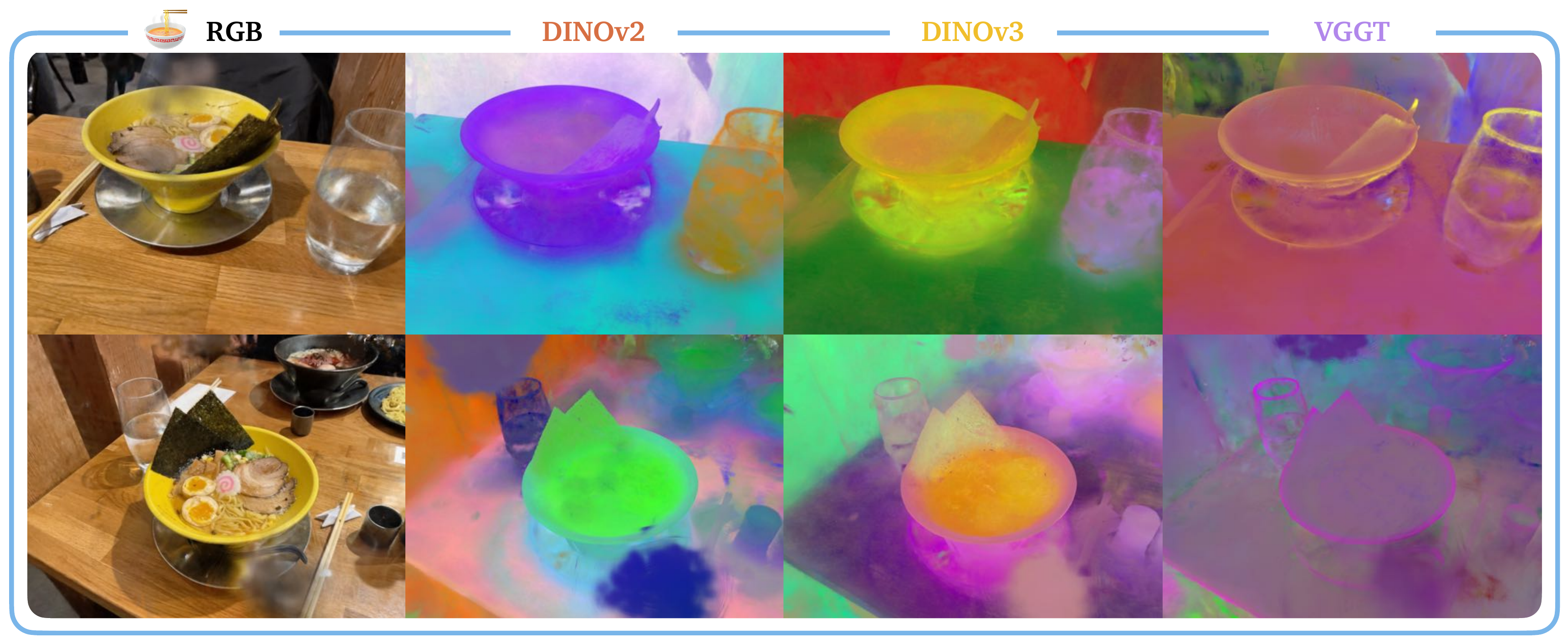}
    \caption{Semantic content of distilled features visualized for NERF for the LERF ramen scene.}
    \label{fig:nerf_pca_lerf_ramen}
\end{figure}

\begin{figure}[H]
    \centering
    \includegraphics[page=3, width=\linewidth]{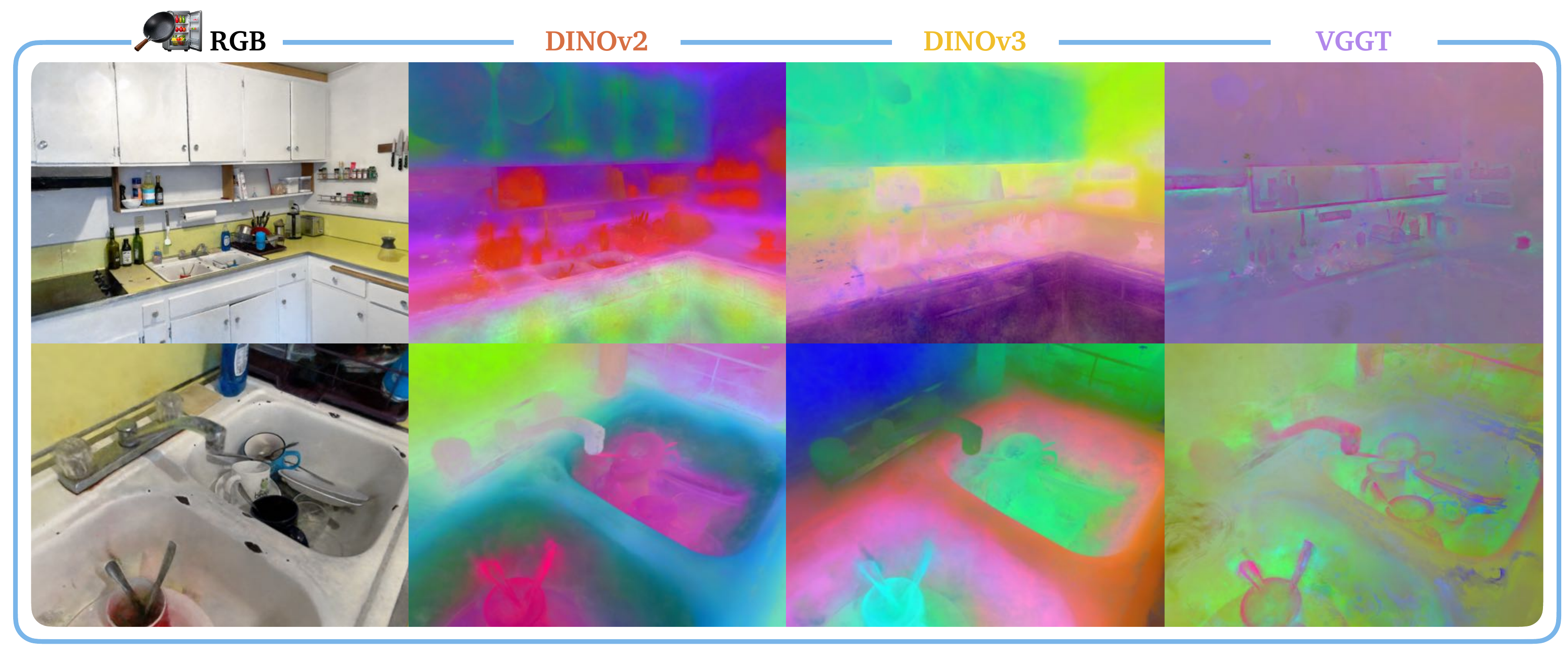}
    \caption{Semantic content of distilled features visualized for NERF for the LERF kitchen scene.}
    \label{fig:nerf_pca_lerf_kitchen}
\end{figure}

\begin{figure}[H]
    \centering
    \includegraphics[page=1, width=0.8\linewidth]{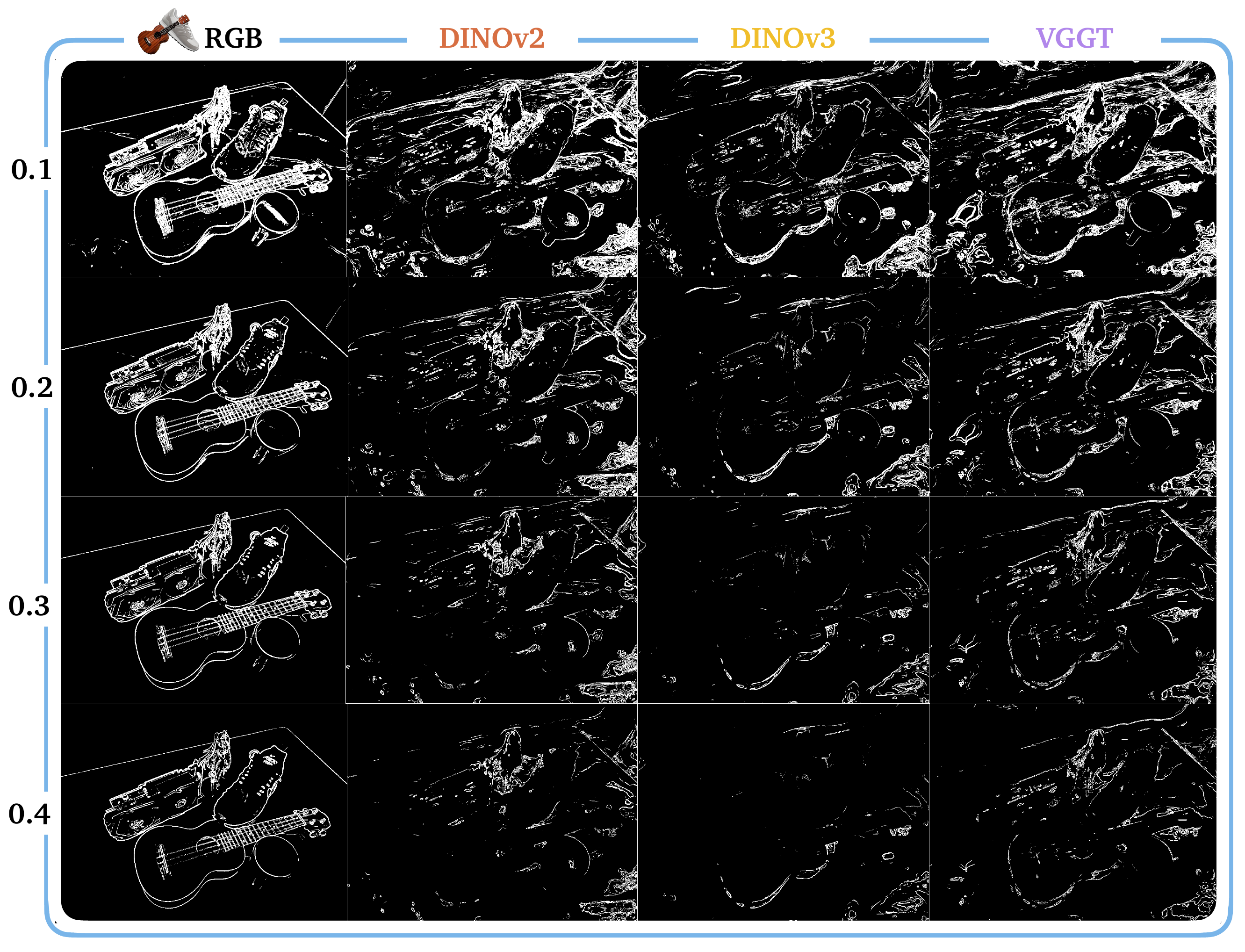}
    \caption{GFF of visual-geometry and visual-only features visualized for GS for the 3D-OVS table scene.}
    \label{fig:gs_pca_edge_3dovs_table}
\end{figure}

\begin{figure}[H]
    \centering
    \includegraphics[page=2, width=0.8\linewidth]{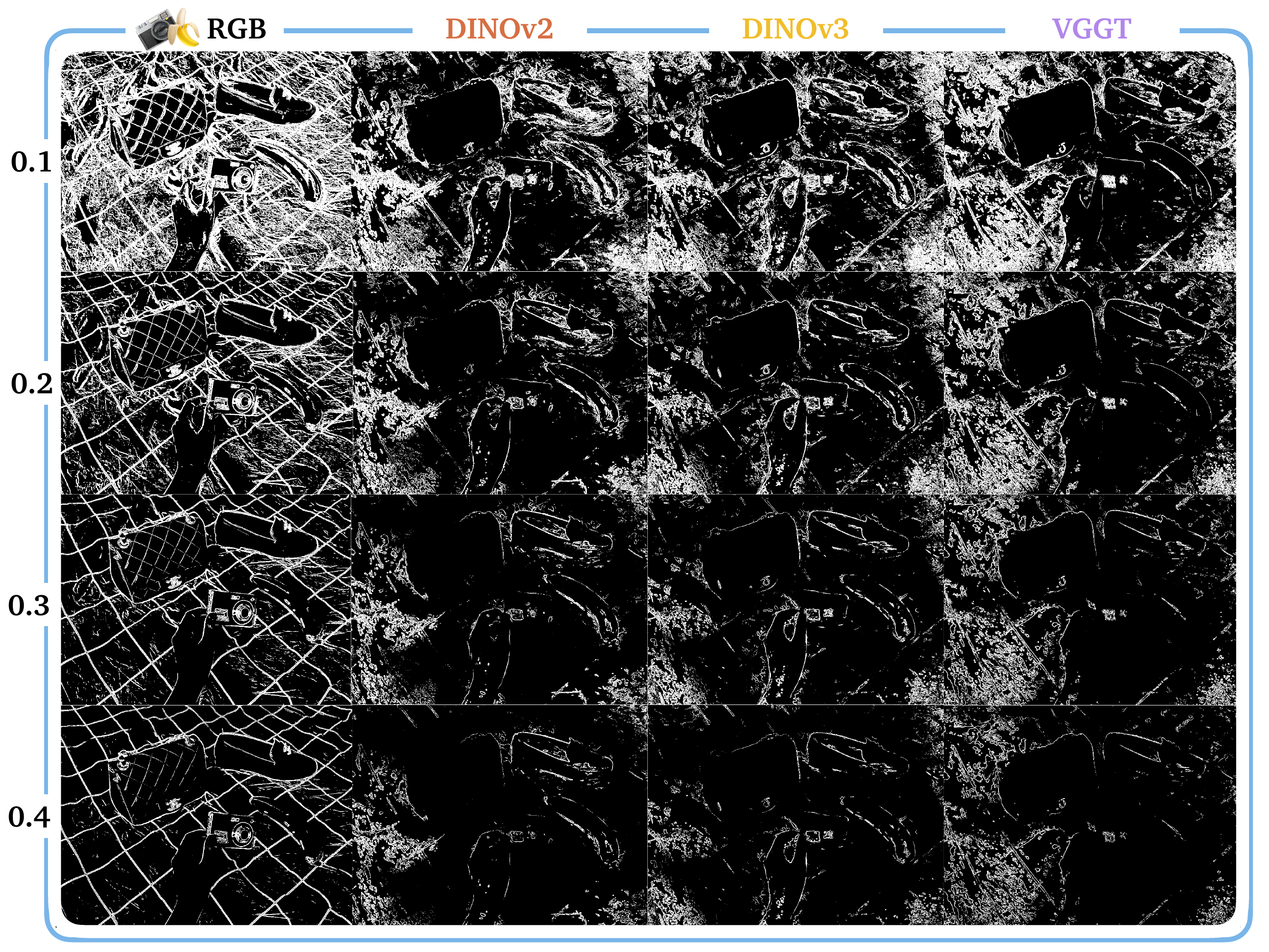}
    \caption{GFF of visual-geometry and visual-only features visualized for GS for the 3D-OVS bed scene.}
    \label{fig:gs_pca_edge_3dovs_bed}
\end{figure}

\begin{figure}[H]
    \centering
    \includegraphics[page=3, width=0.8\linewidth]{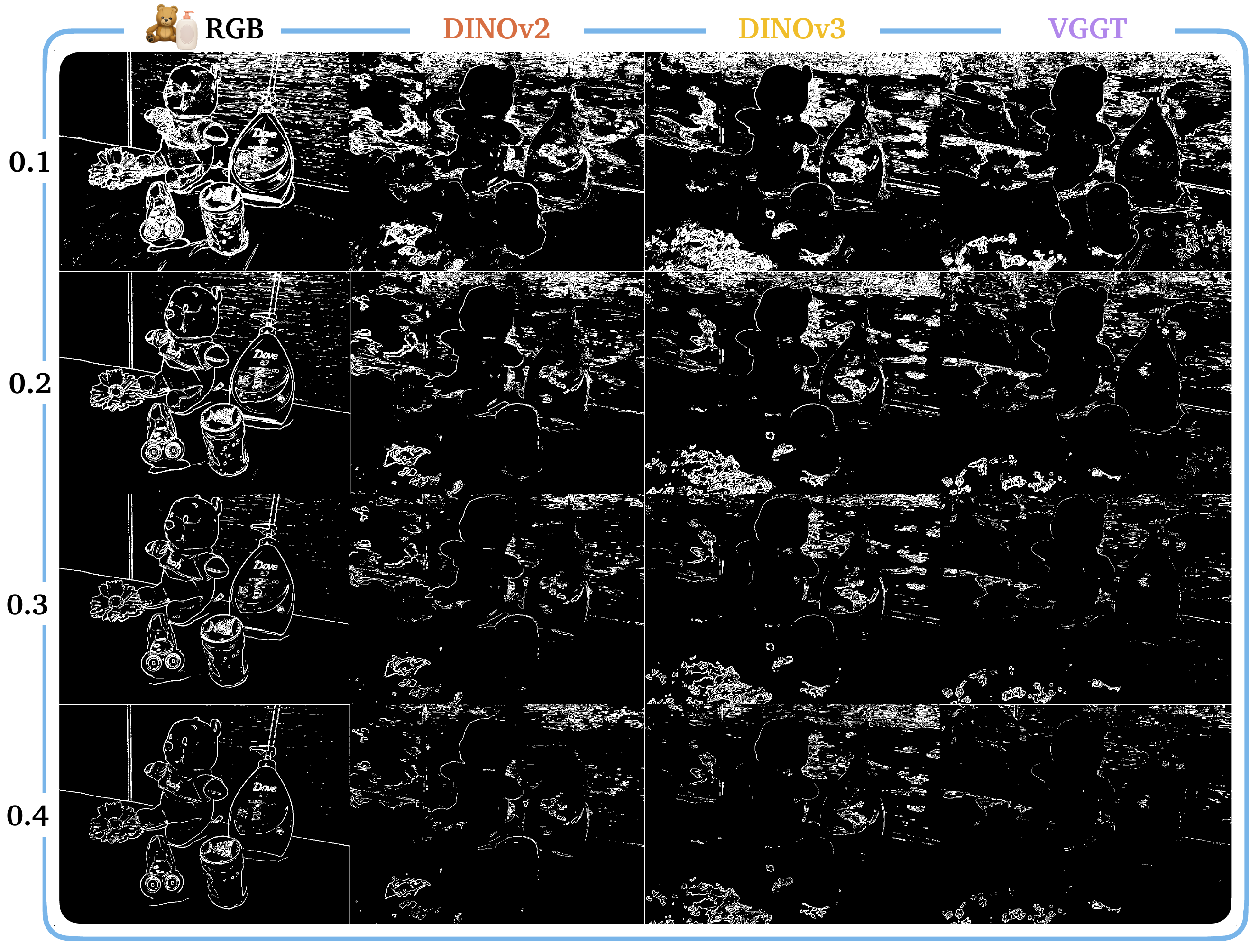}
    \caption{GFF of visual-geometry and visual-only features visualized for GS for the 3D-OVS desk scene.}
    \label{fig:gs_pca_edge_3dovs_desk}
\end{figure}

\begin{figure}[H]
    \centering
    \includegraphics[page=4, width=0.8\linewidth]{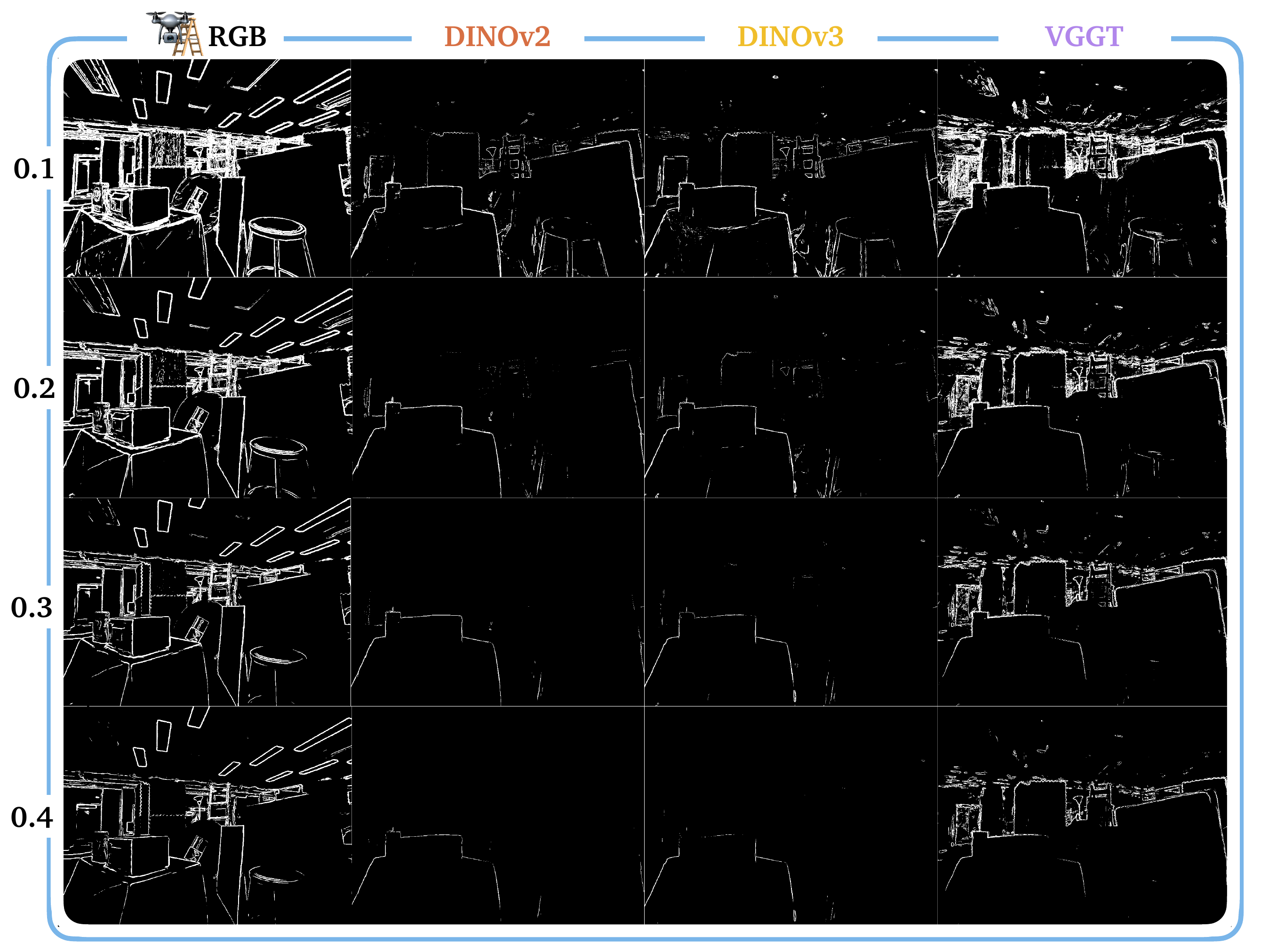}
    \caption{GFF of visual-geometry and visual-only features visualized for GS for the drone scene.}
    \label{fig:gs_pca_edge_robot_drone}
\end{figure}

\begin{figure}[H]
    \centering
    \includegraphics[page=5, width=0.8\linewidth]{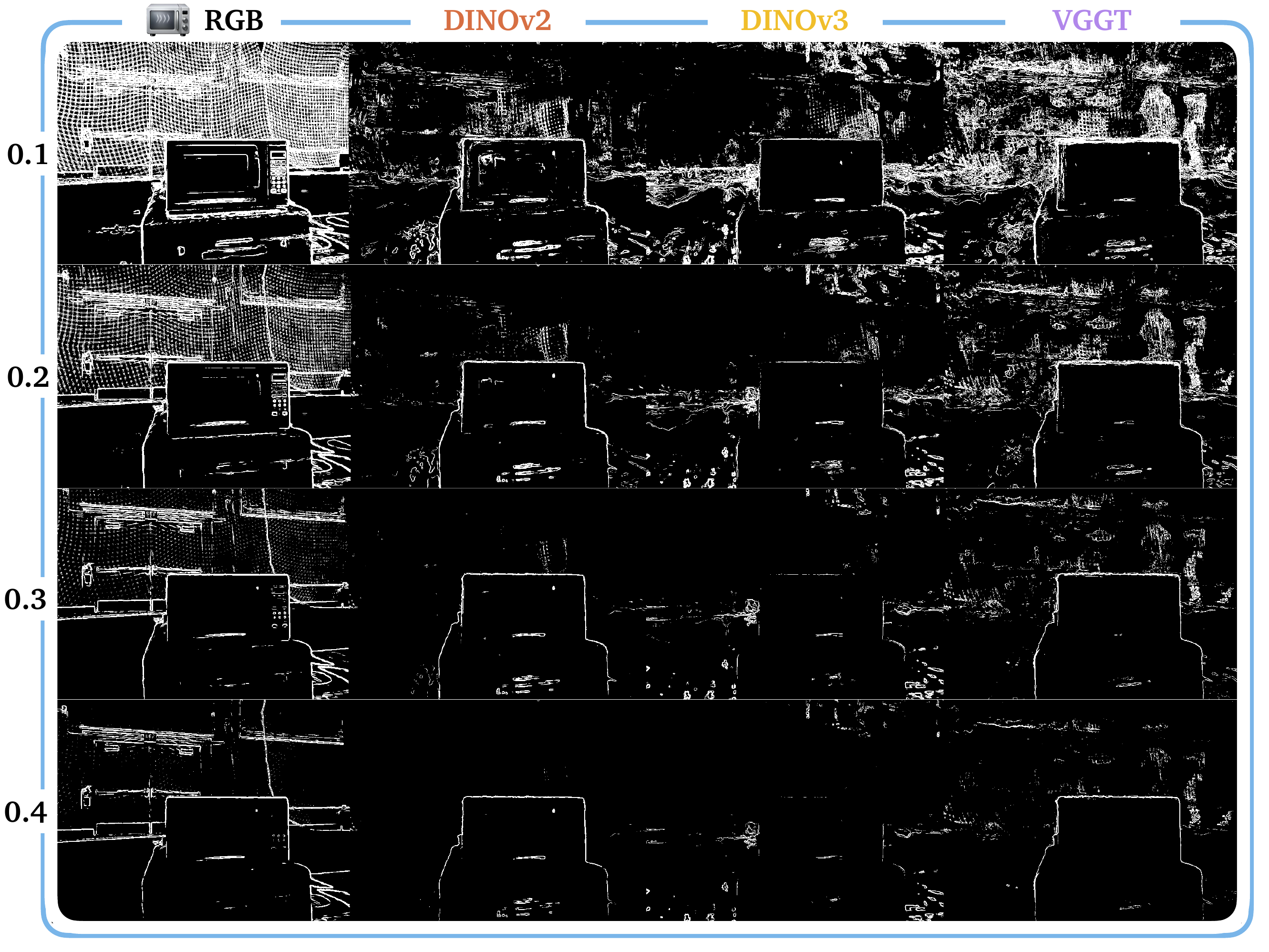}
    \caption{GFF of visual-geometry and visual-only features visualized for GS for the quadruped kitchen scene.}
    \label{fig:gs_pca_edge_robot_kitchen}
\end{figure}

\begin{figure}[H]
    \centering
    \includegraphics[page=6, width=0.8\linewidth]{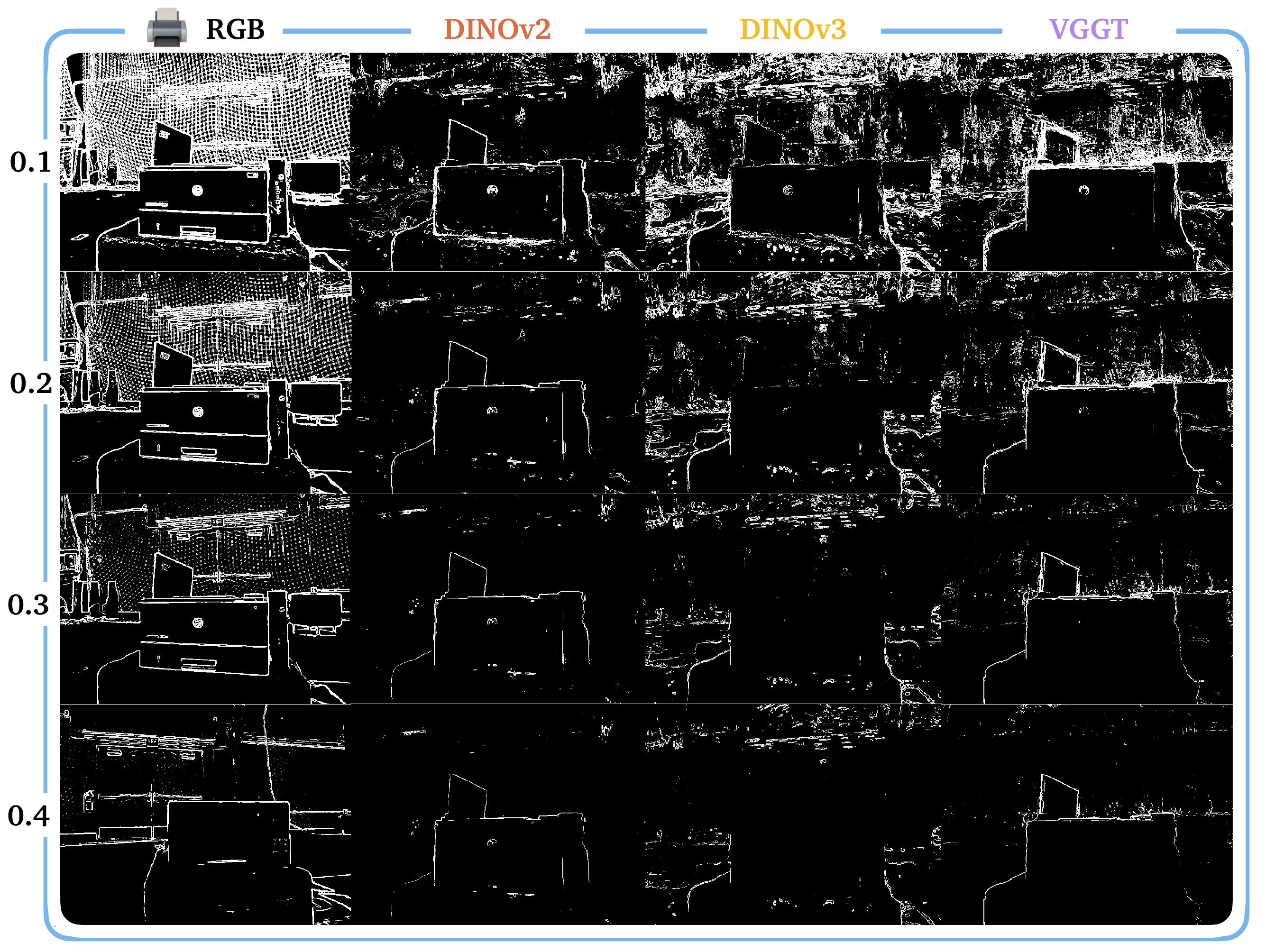}
    \caption{GFF of visual-geometry and visual-only features visualized for GS for the quadruped office scene.}
    \label{fig:gs_pca_edge_robot_office}
\end{figure}

\begin{figure}[H]
    \centering
    \includegraphics[page=7, width=0.8\linewidth]{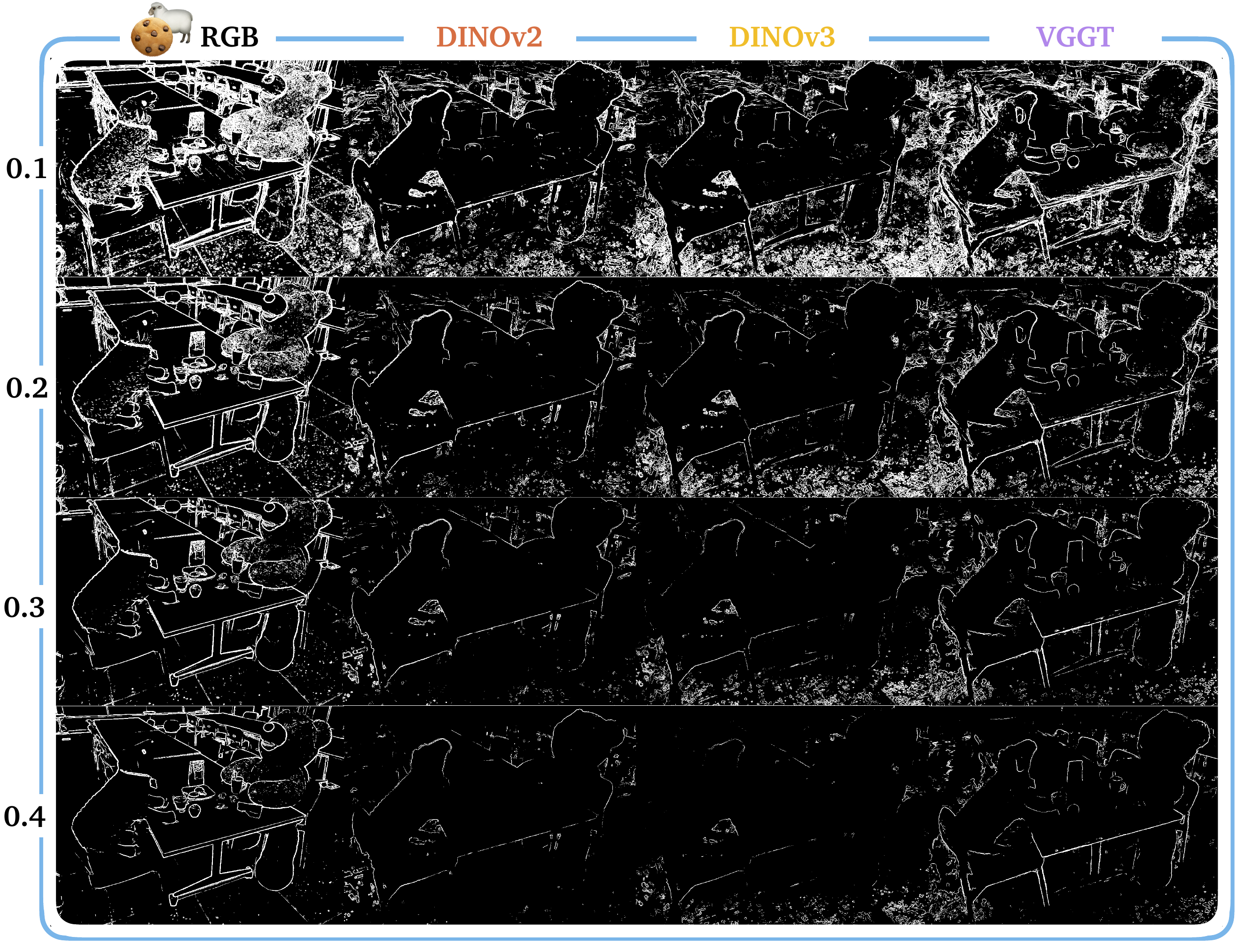}
    \caption{GFF of visual-geometry and visual-only features visualized for GS for the LERF teatime scene.}
    \label{fig:gs_pca_edge_lerf_teatime}
\end{figure}

\begin{figure}[H]
    \centering
    \includegraphics[page=8, width=0.8\linewidth]{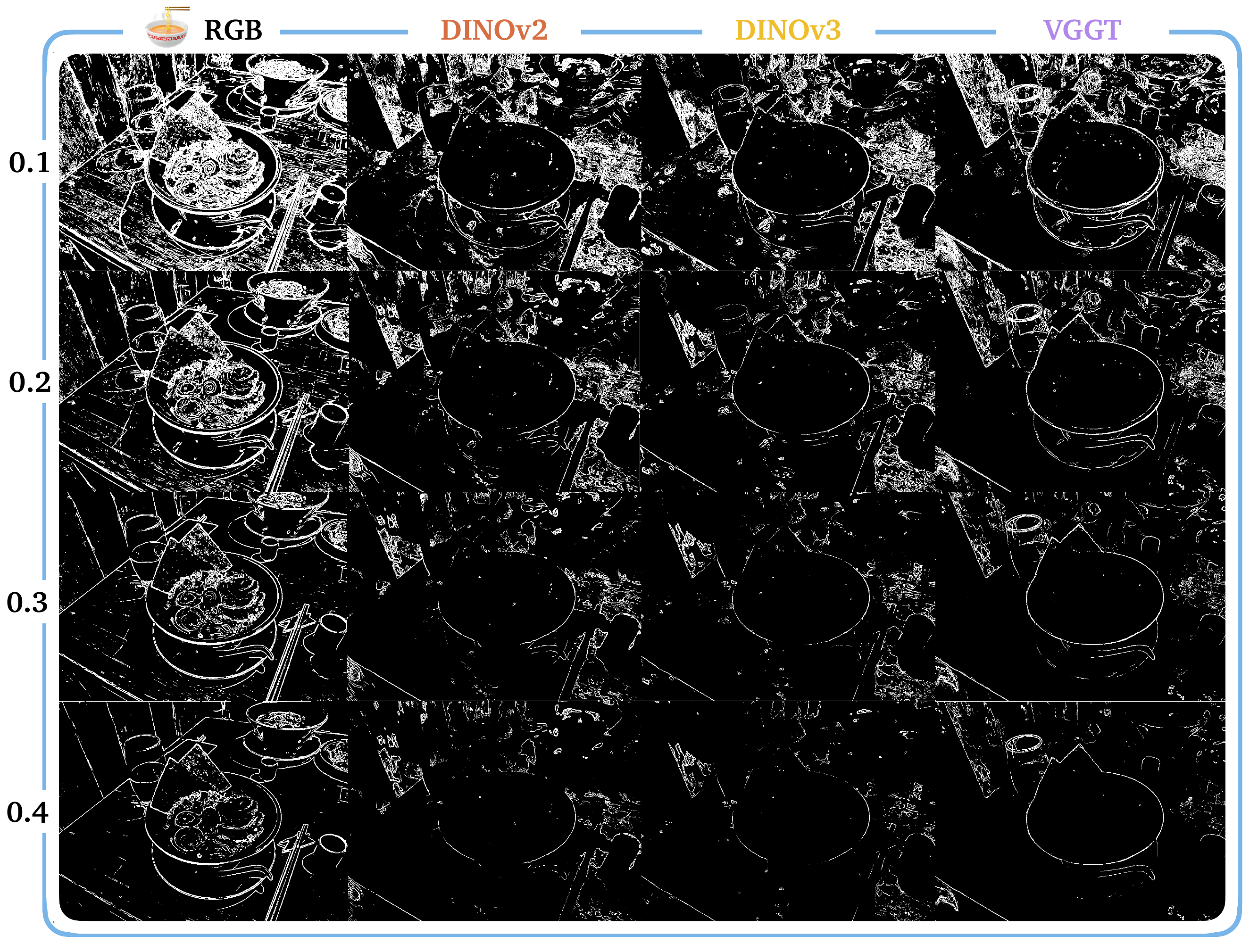}
    \caption{GFF of visual-geometry and visual-only features visualized for GS for the LERF ramen scene.}
    \label{fig:gs_pca_edge_lerf_ramen}
\end{figure}

\begin{figure}[H]
    \centering
    \includegraphics[page=9, width=0.8\linewidth]{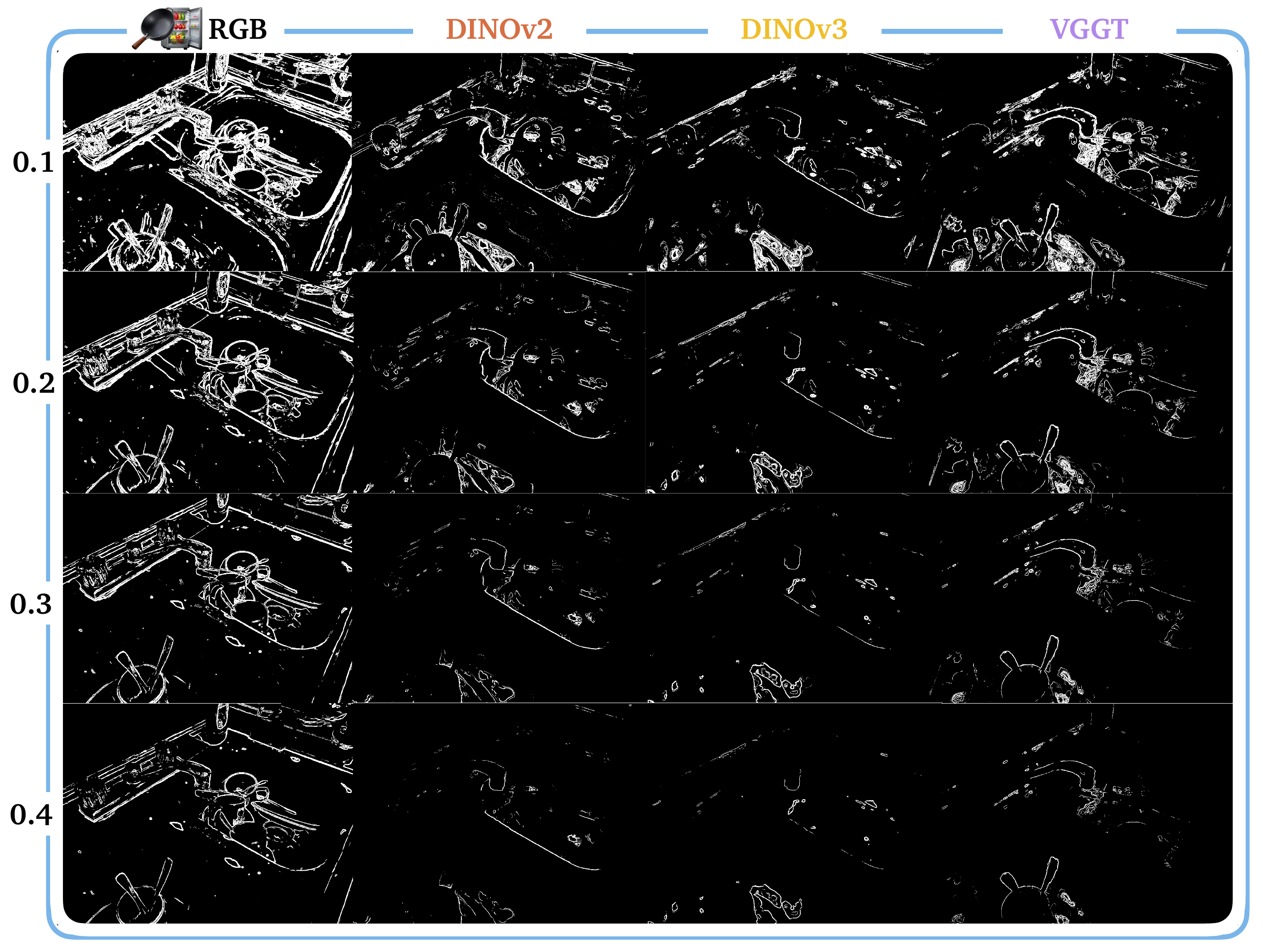}
    \caption{GFF of visual-geometry and visual-only features visualized for GS for the LERF kitchen scene.}
    \label{fig:gs_pca_edge_lerf_kitchen}
\end{figure}

\begin{figure}[H]
    \centering
    \includegraphics[page=1, width=0.8\linewidth]{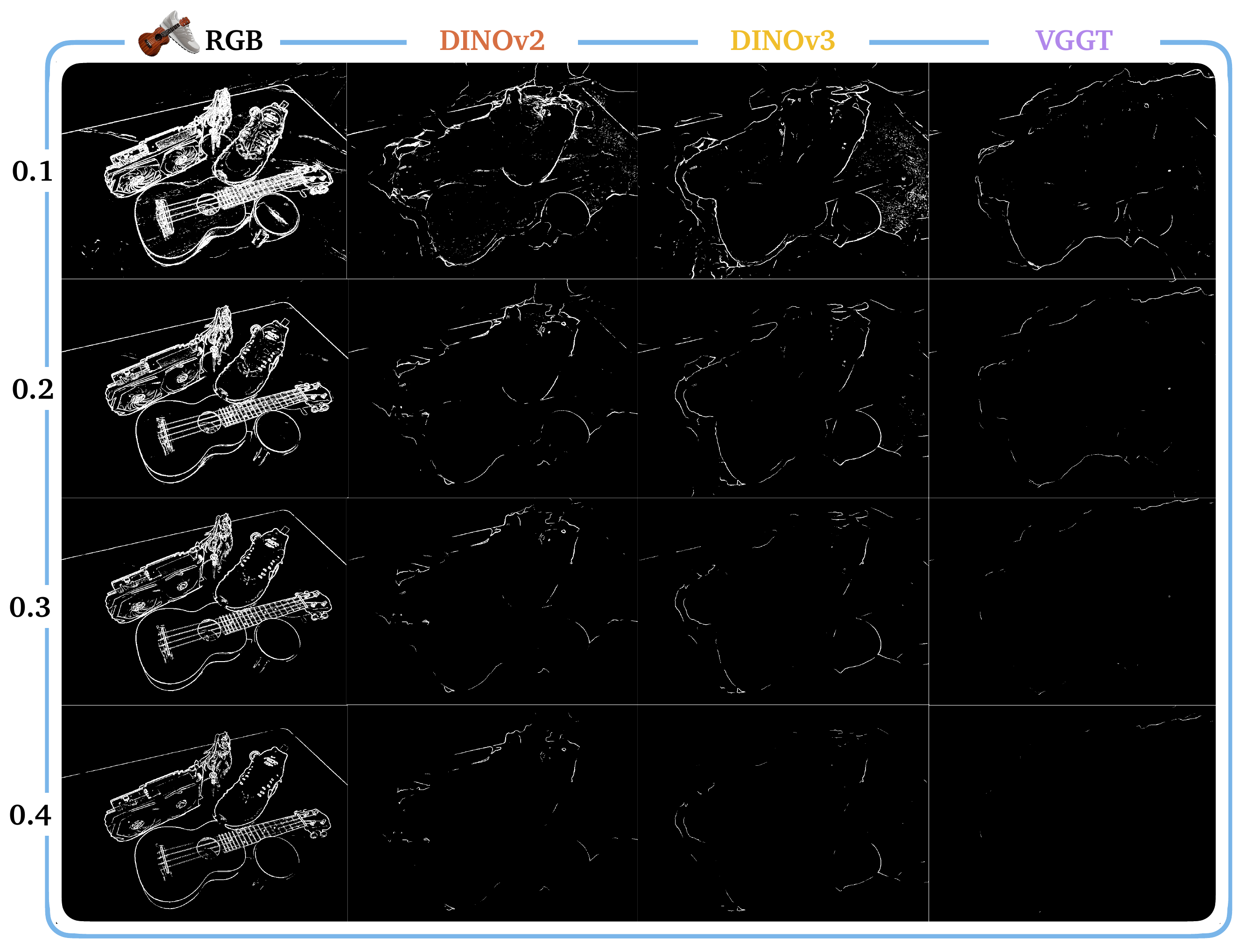}
    \caption{GFF of visual-geometry and visual-only features visualized for NERF for the 3D-OVS table scene.}
    \label{fig:nerf_pca_edge_3dovs_table}
\end{figure}

\begin{figure}[H]
    \centering
    \includegraphics[page=2, width=0.8\linewidth]{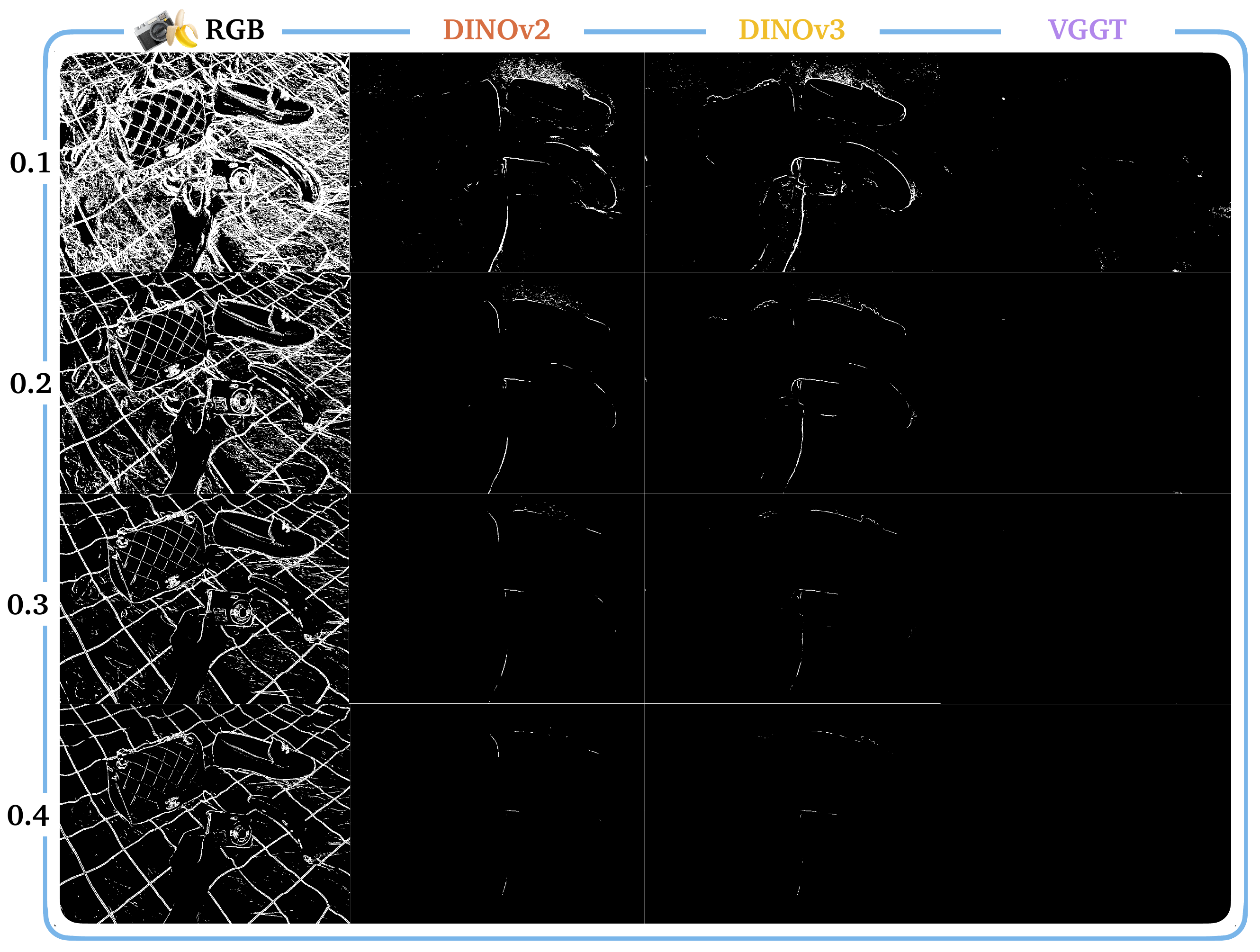}
    \caption{GFF of visual-geometry and visual-only features visualized for NERF for the 3D-OVS bed scene.}
    \label{fig:nerf_pca_edge_3dovs_bed}
\end{figure}

\begin{figure}[H]
    \centering
    \includegraphics[page=3, width=0.8\linewidth]{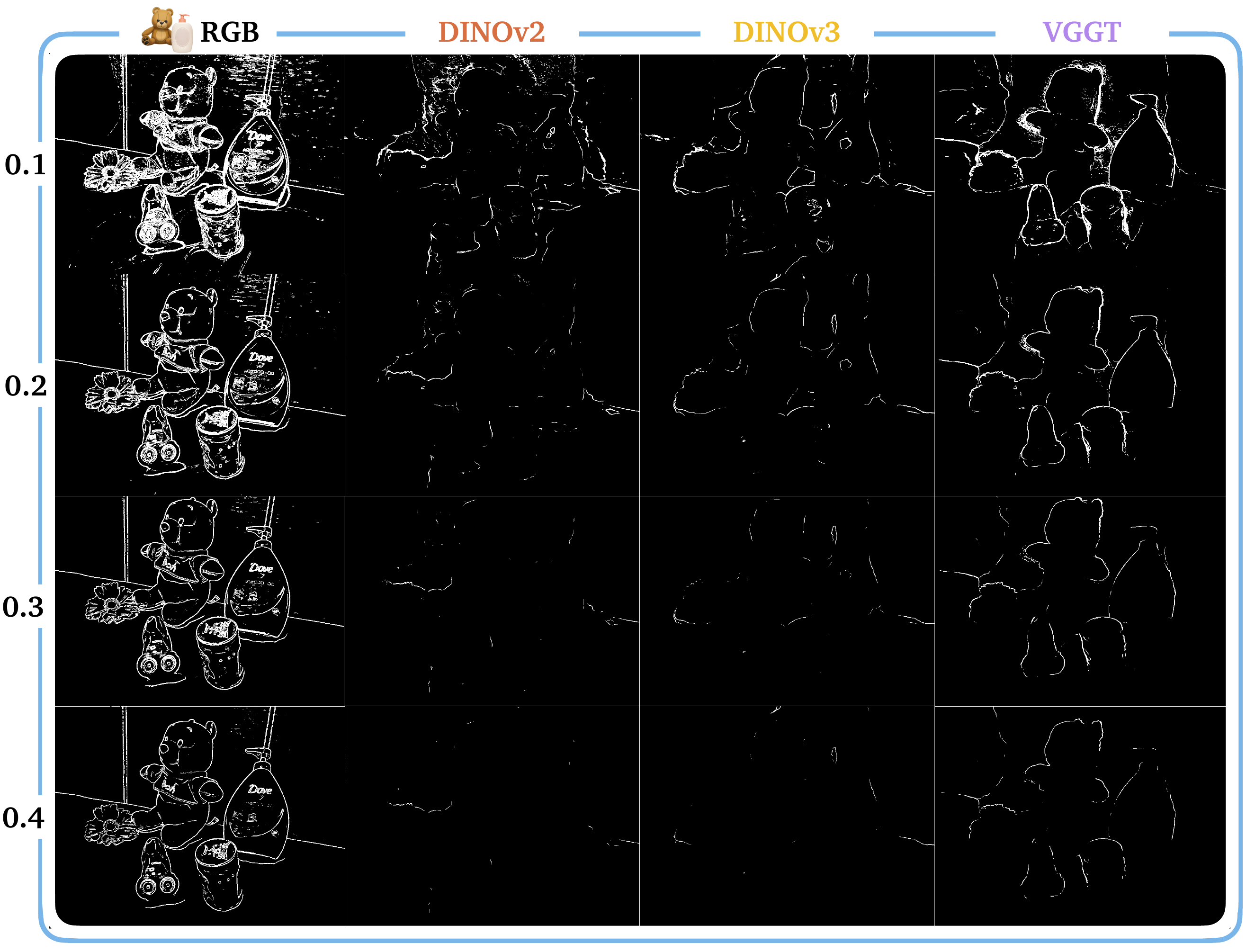}
    \caption{GFF of visual-geometry and visual-only features visualized for NERF for the 3D-OVS desk scene.}
    \label{fig:nerf_pca_edge_3dovs_desk}
\end{figure}

\begin{figure}[H]
    \centering
    \includegraphics[page=4, width=0.8\linewidth]{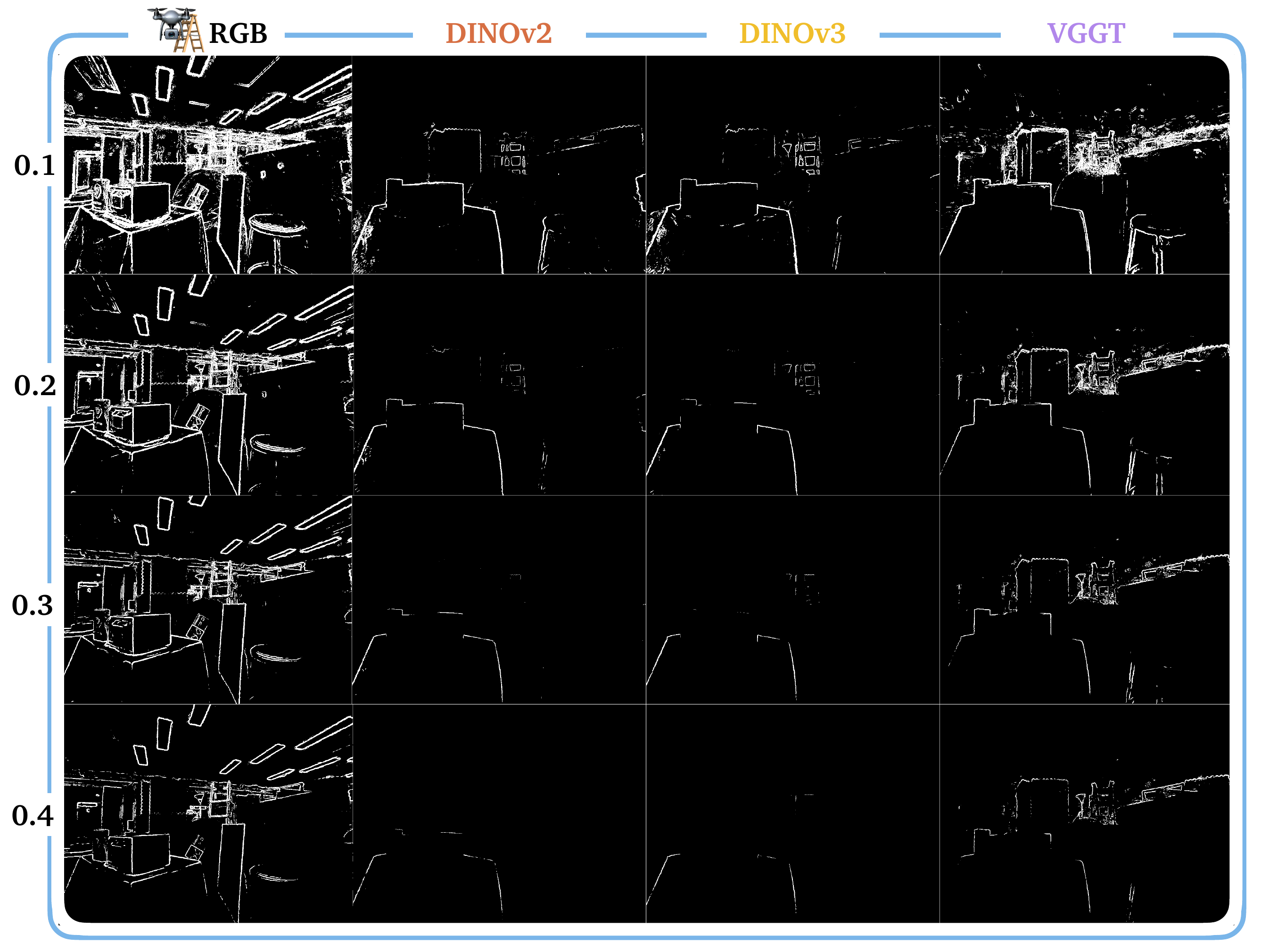}
    \caption{GFF of visual-geometry and visual-only features visualized for NERF for the drone scene.}
    \label{fig:nerf_pca_edge_robot_drone}
\end{figure}

\begin{figure}[H]
    \centering
    \includegraphics[page=5, width=0.8\linewidth]{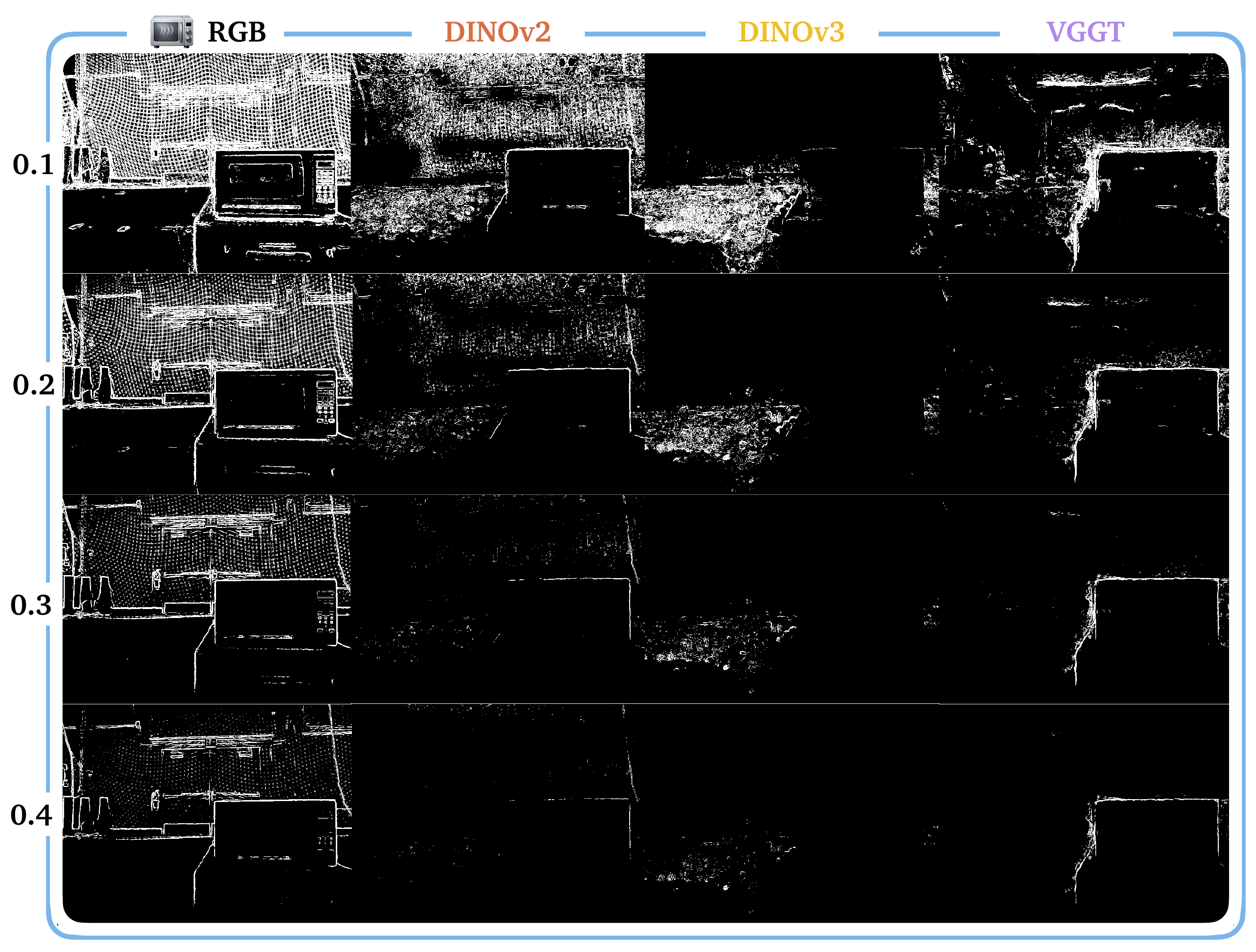}
    \caption{GFF of visual-geometry and visual-only features visualized for NERF for the quadruped kitchen scene.}
    \label{fig:nerf_pca_edge_robot_kitchen}
\end{figure}

\begin{figure}[H]
    \centering
    \includegraphics[page=6, width=0.8\linewidth]{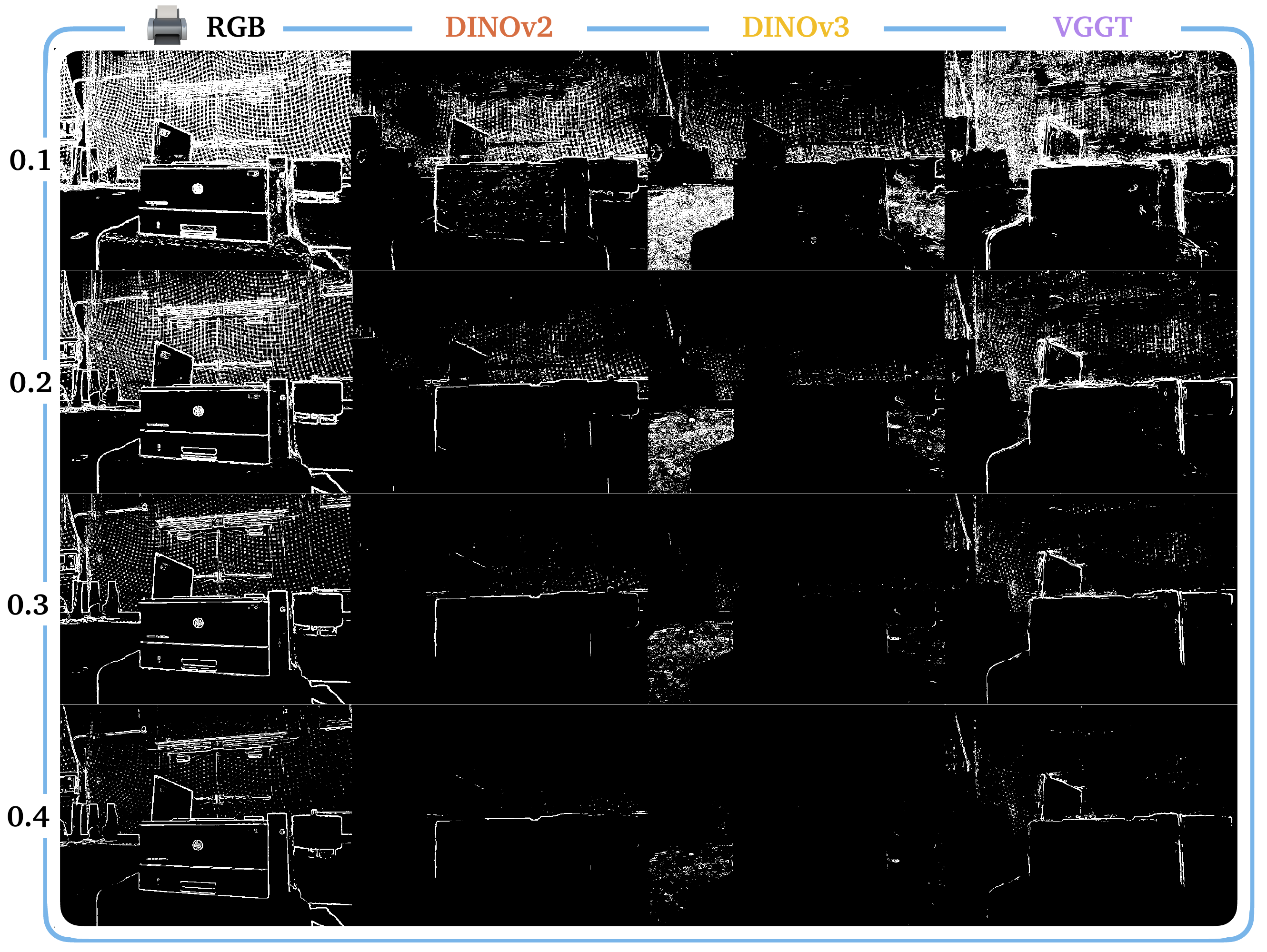}
    \caption{GFF of visual-geometry and visual-only features visualized for NERF for the quadruped office scene.}
    \label{fig:nerf_pca_edge_robot_office}
\end{figure}

\begin{figure}[H]
    \centering
    \includegraphics[page=7, width=0.8\linewidth]{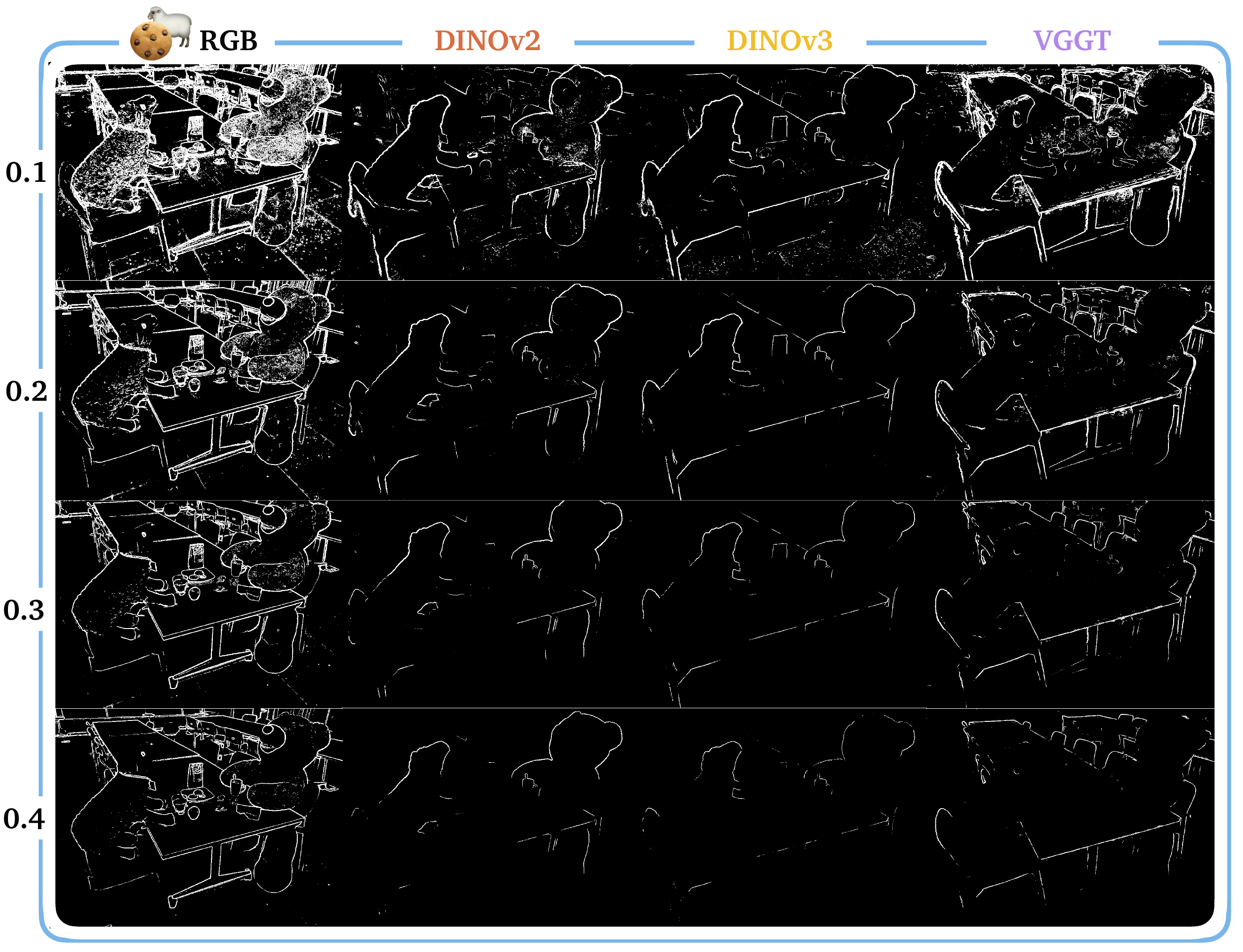}
    \caption{GFF of visual-geometry and visual-only features visualized for NERF for the LERF teatime scene.}
    \label{fig:nerf_pca_edge_lerf_teatime}
\end{figure}

\begin{figure}[H]
    \centering
    \includegraphics[page=8, width=0.8\linewidth]{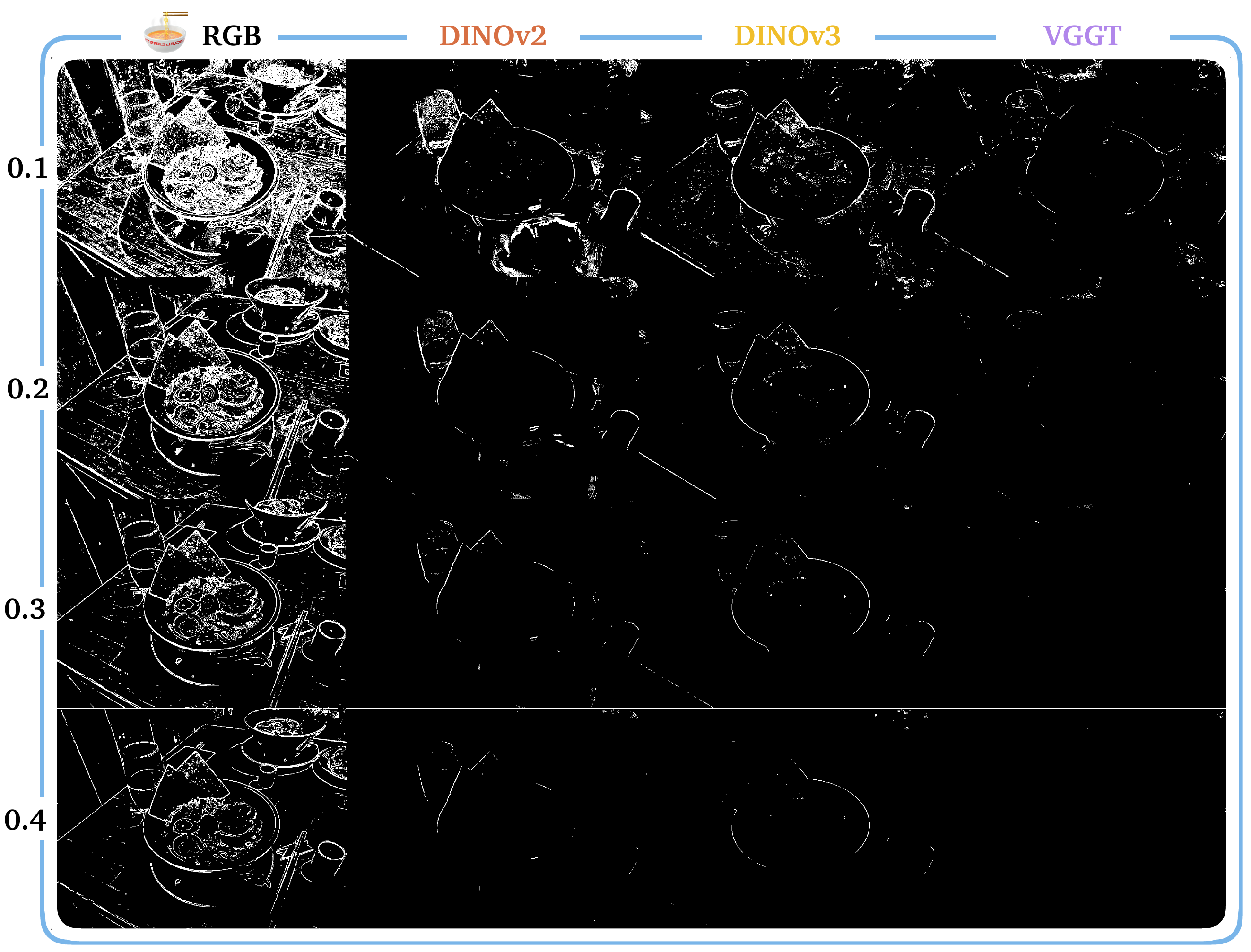}
    \caption{GFF of visual-geometry and visual-only features visualized for NERF for the LERF ramen scene.}
    \label{fig:nerf_pca_edge_lerf_ramen}
\end{figure}

\begin{figure}[H]
    \centering
    \includegraphics[page=9, width=0.8\linewidth]{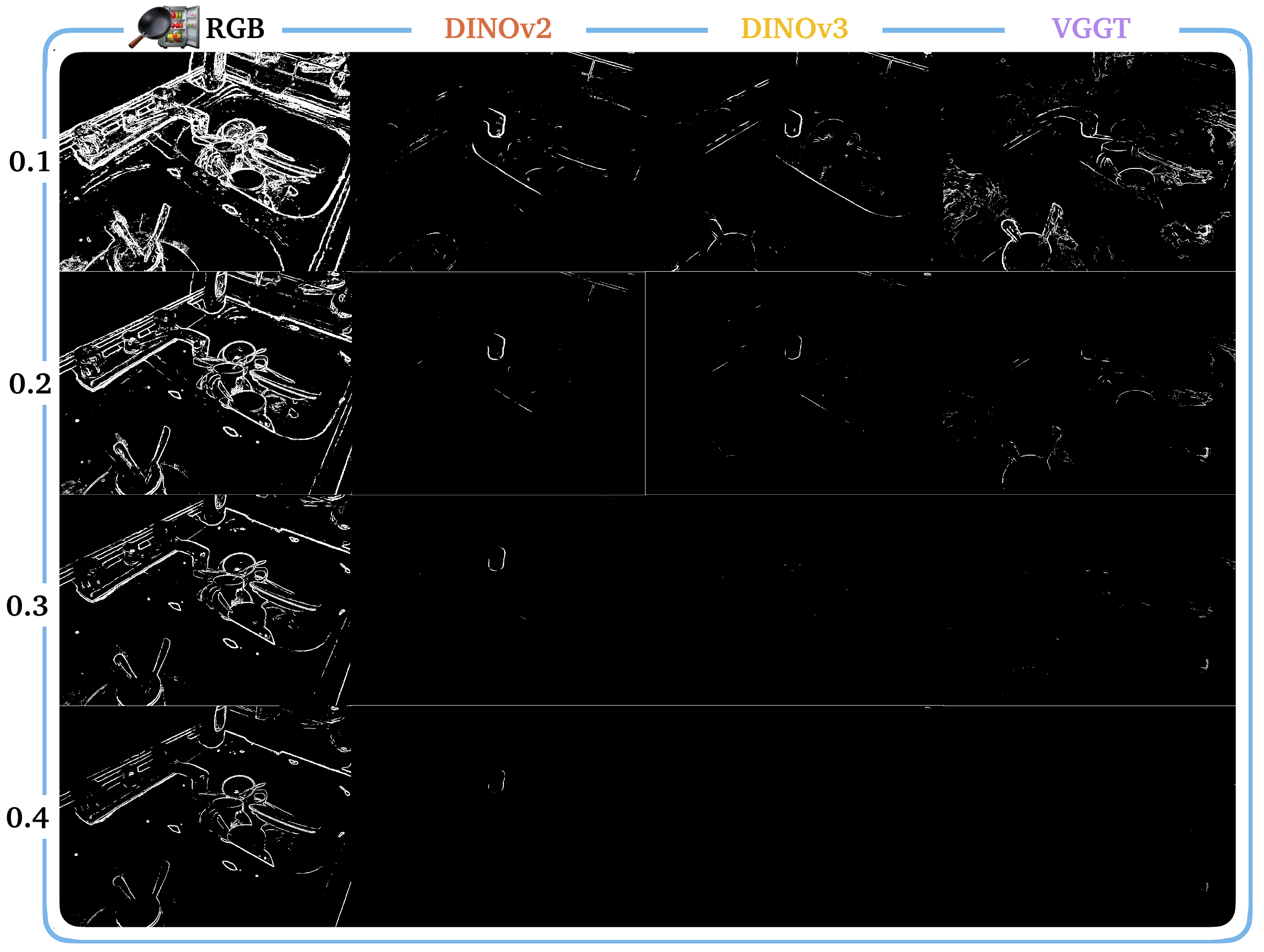}
    \caption{GFF of visual-geometry and visual-only features visualized for NERF for the LERF kitchen scene.}
    \label{fig:nerf_pca_edge_lerf_kitchen}
\end{figure}

\subsection{Additional Semantic Localization Results}
\label{sec:app_semantic_localization}
Below, we include semantic localization visualizations, sampling one or two camera views from each scene.

\begin{figure}[H]
    \centering
    \includegraphics[width=\linewidth]{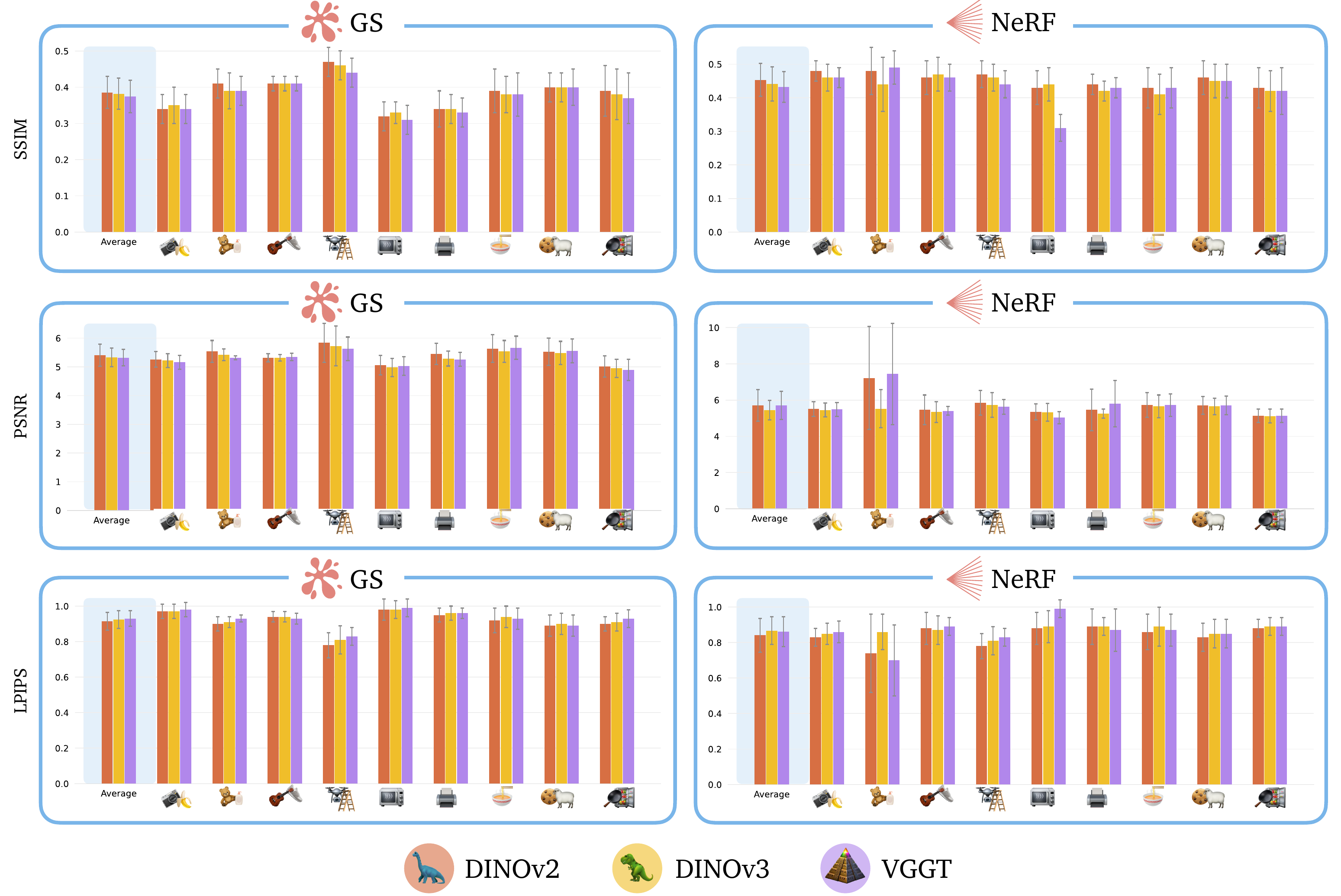}
    \caption{
    Quantitative results for the three metrics (SSIM, PSNR, LPIPS) of semantic object localization across each scene. 
    }
    \label{fig:segmentation_full_results}
\end{figure}

\begin{figure}[H]
    \centering
    \includegraphics[page=1, width=\linewidth]{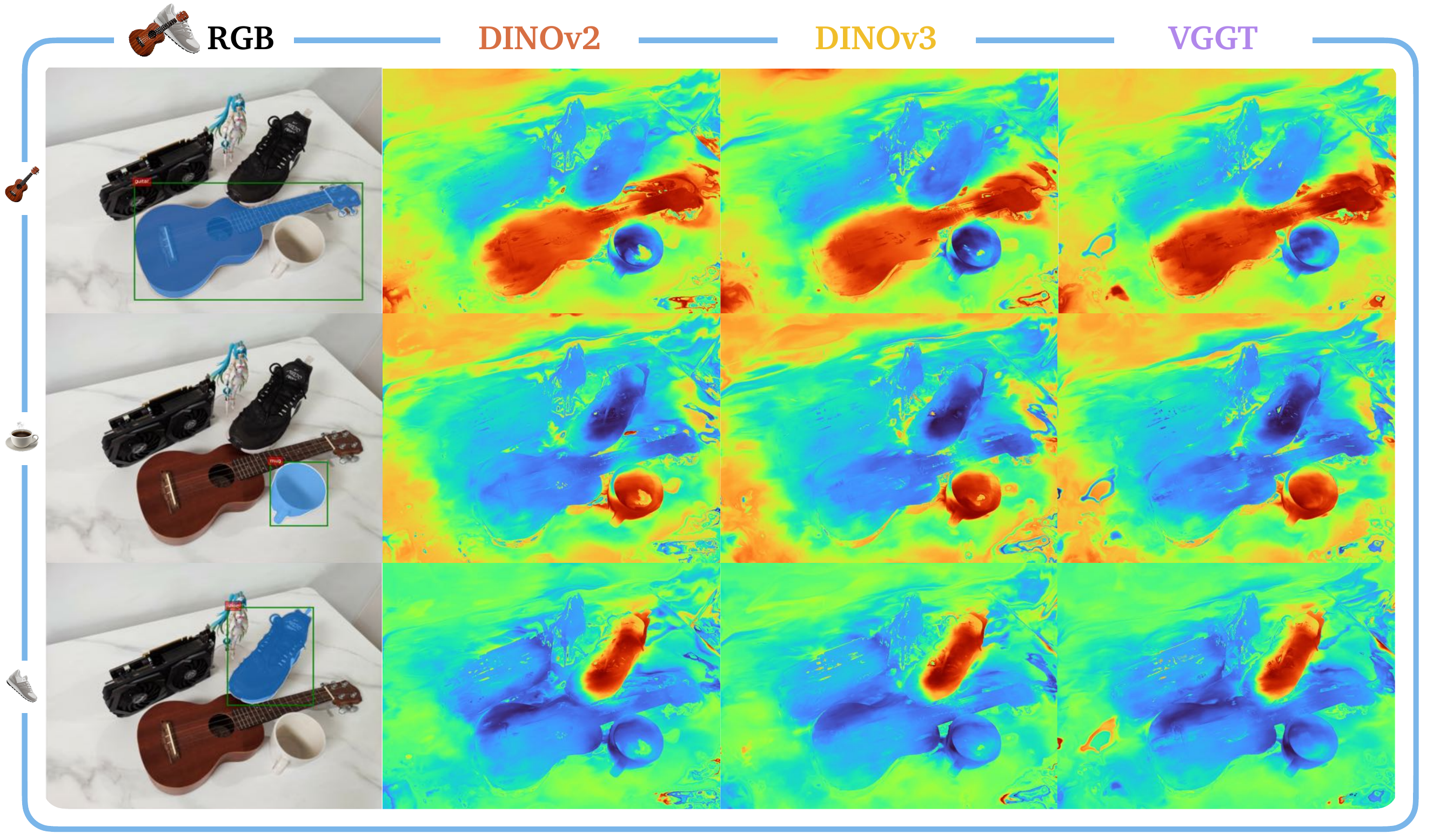}
    \caption{Semantic object localization visualized for GS for the 3D-OVS table scene.}
    \label{fig:gs_segmentation_3dovs_table}
\end{figure}

\begin{figure}[H]
    \centering
    \includegraphics[page=2, width=\linewidth]{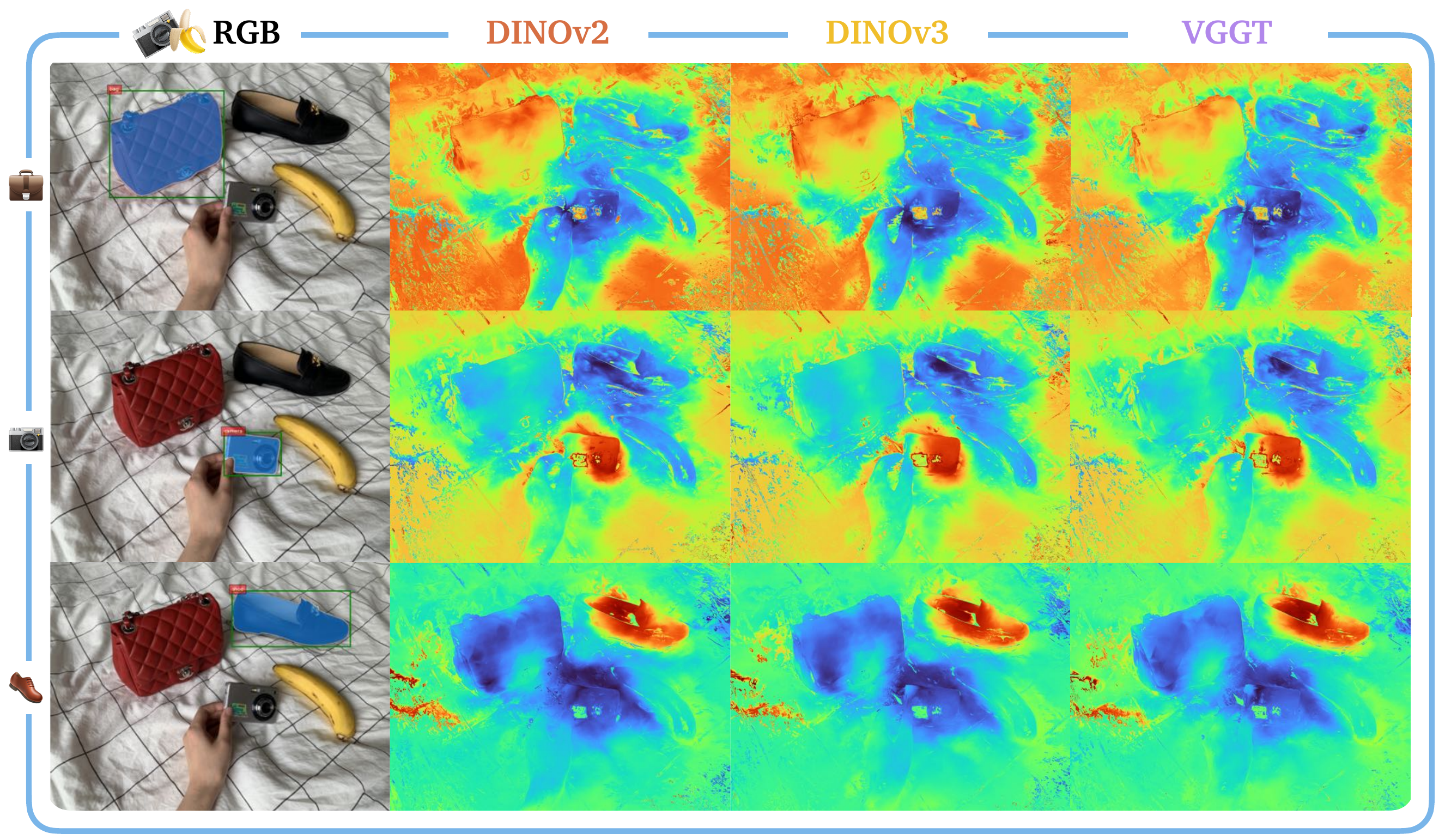}
    \caption{Semantic object localization visualized for GS for the 3D-OVS bed scene.}
    \label{fig:gs_segmentation_3dovs_bed}
\end{figure}

\begin{figure}[H]
    \centering
    \includegraphics[page=3, width=\linewidth]{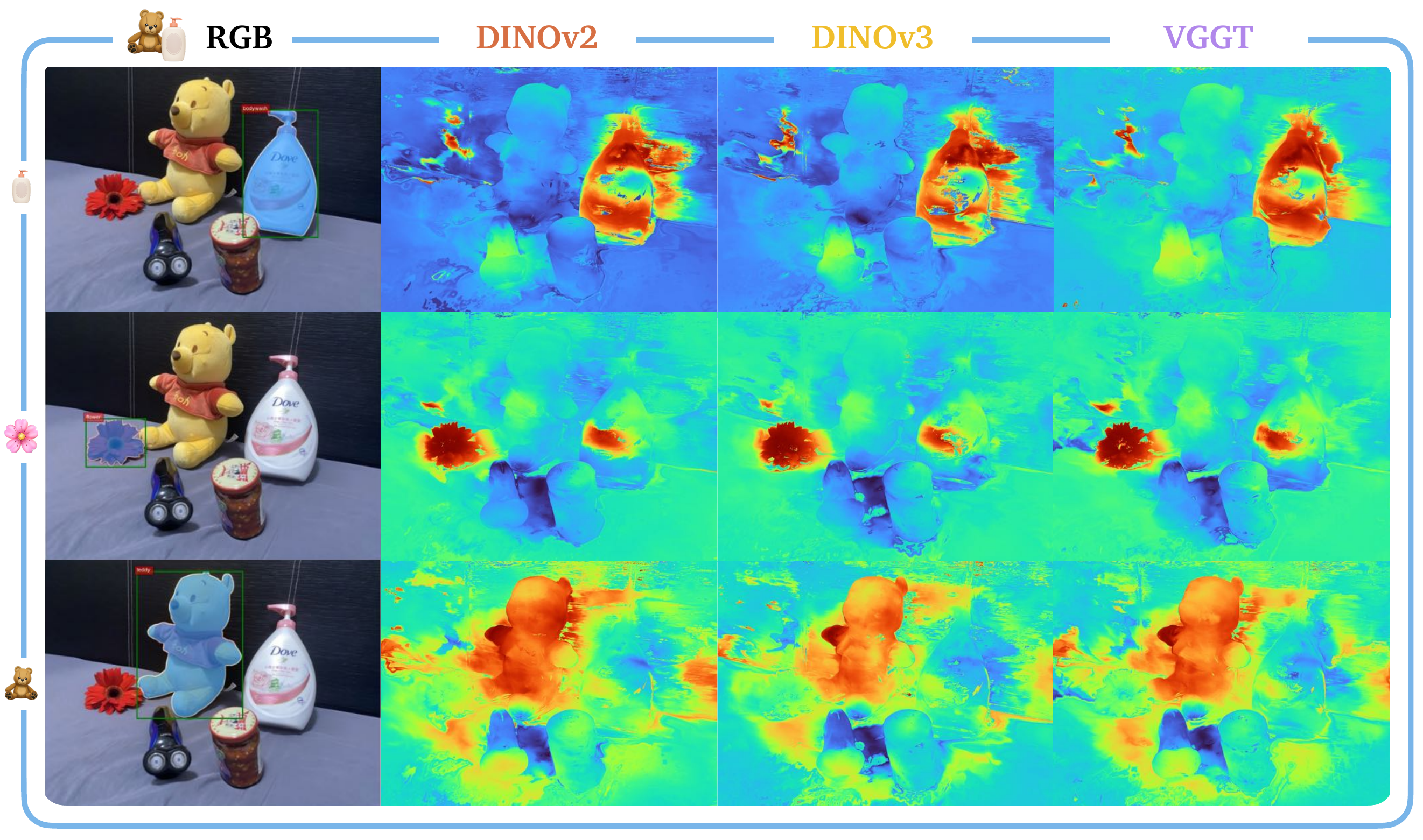}
    \caption{Semantic object localization visualized for GS for the 3D-OVS covered desk scene.}
    \label{fig:gs_segmentation_3dovs_desk}
\end{figure}

\begin{figure}[H]
    \centering
    \includegraphics[page=1, width=\linewidth]{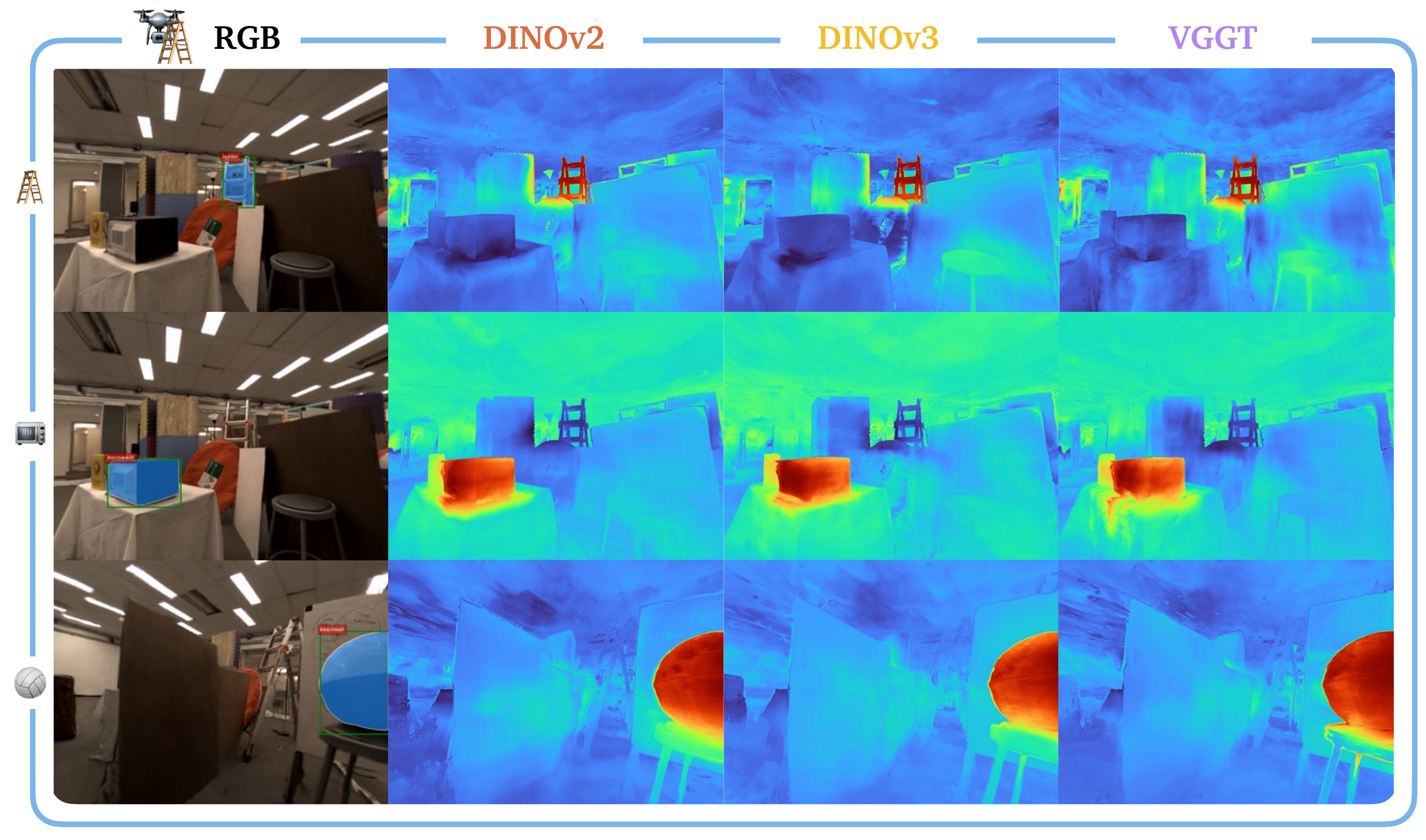}
    \caption{Semantic object localization visualized for GS for the drone scene.}
    \label{fig:gs_segmentation_robot_drone}
\end{figure}

\begin{figure}[H]
    \centering
    \includegraphics[page=2, width=\linewidth]{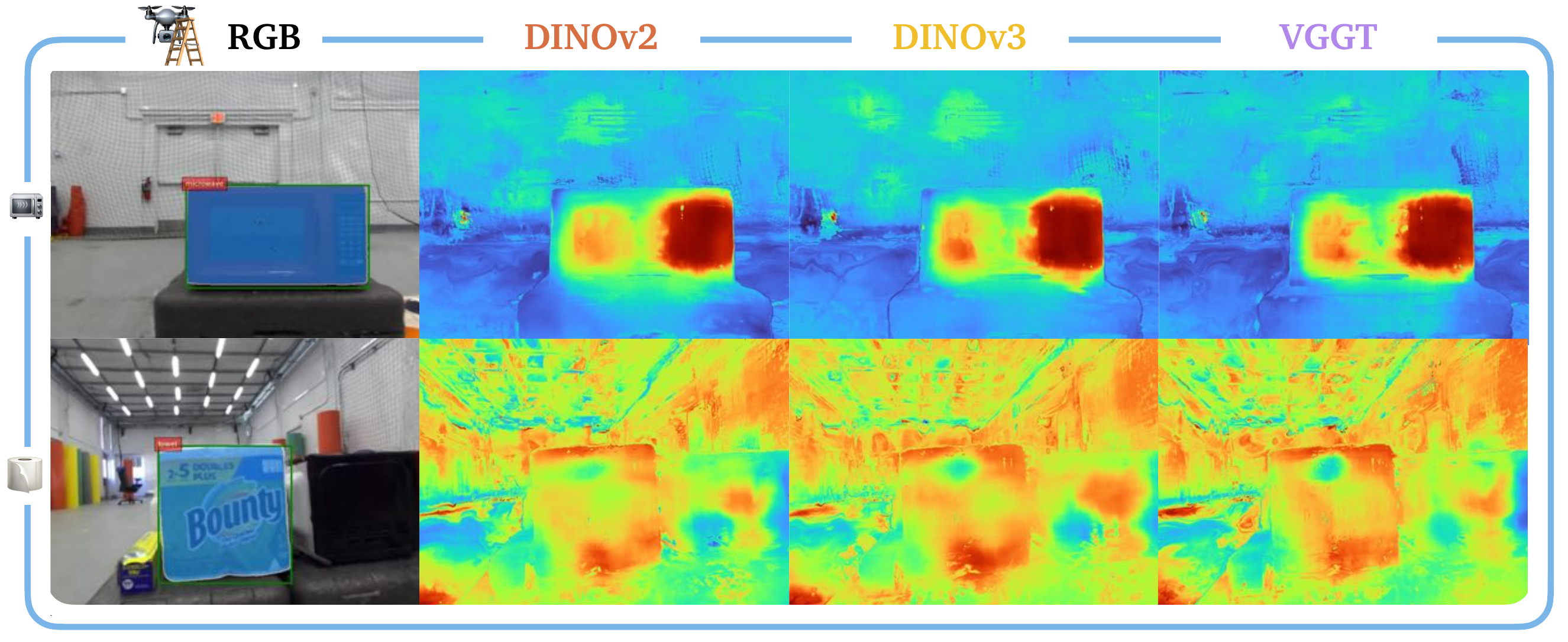}
    \caption{Semantic object localization visualized for GS for the quadruped kitchen scene.}
    \label{fig:gs_segmentation_robot_kitchen}
\end{figure}

\begin{figure}[H]
    \centering
    \includegraphics[page=3, width=\linewidth]{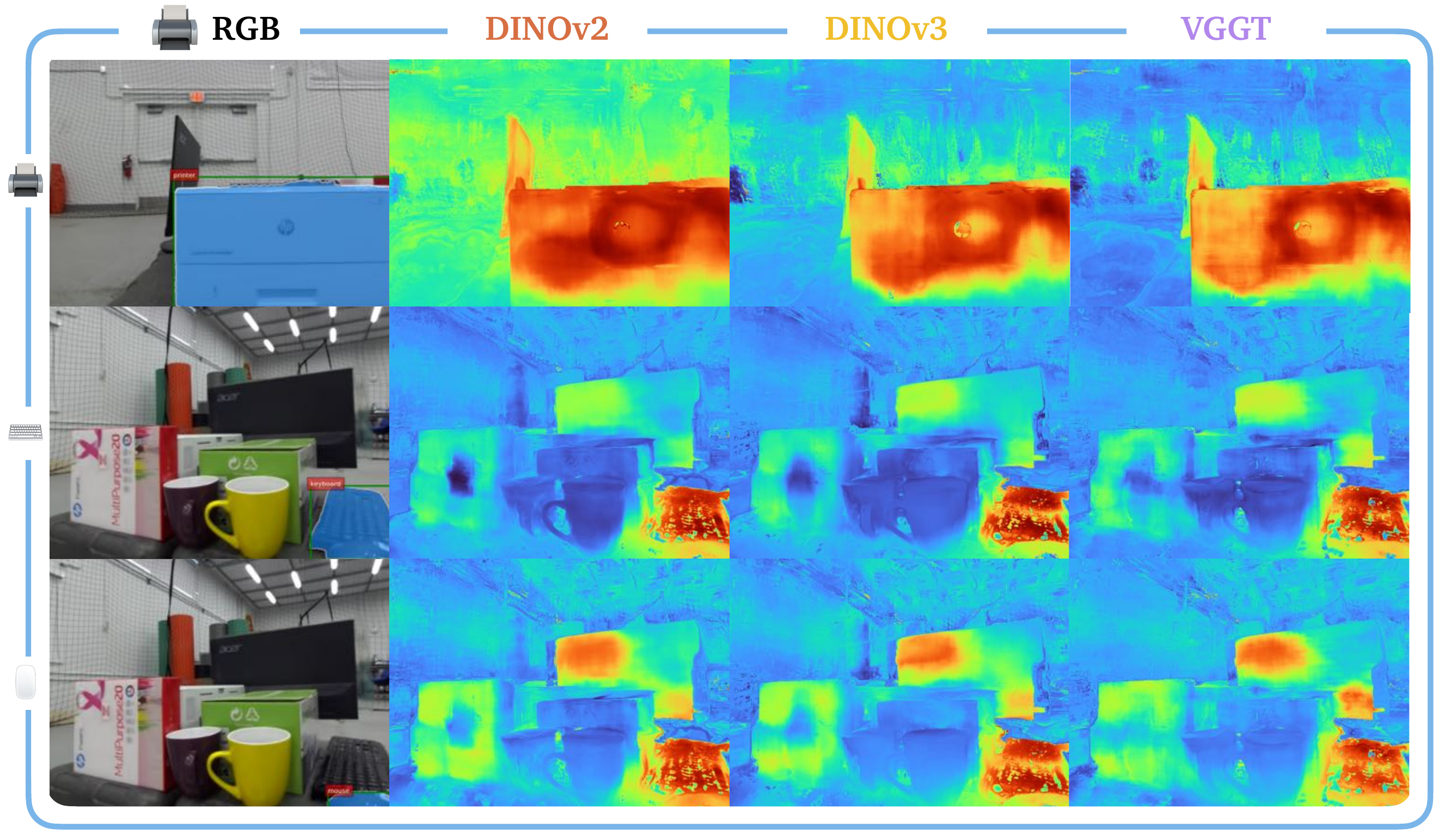}
    \caption{Semantic object localization visualized for GS for the 3D-OVS bed scene.}
    \label{fig:gs_segmentation_robot_office}
\end{figure}

\begin{figure}[H]
    \centering
    \includegraphics[page=1, width=\linewidth]{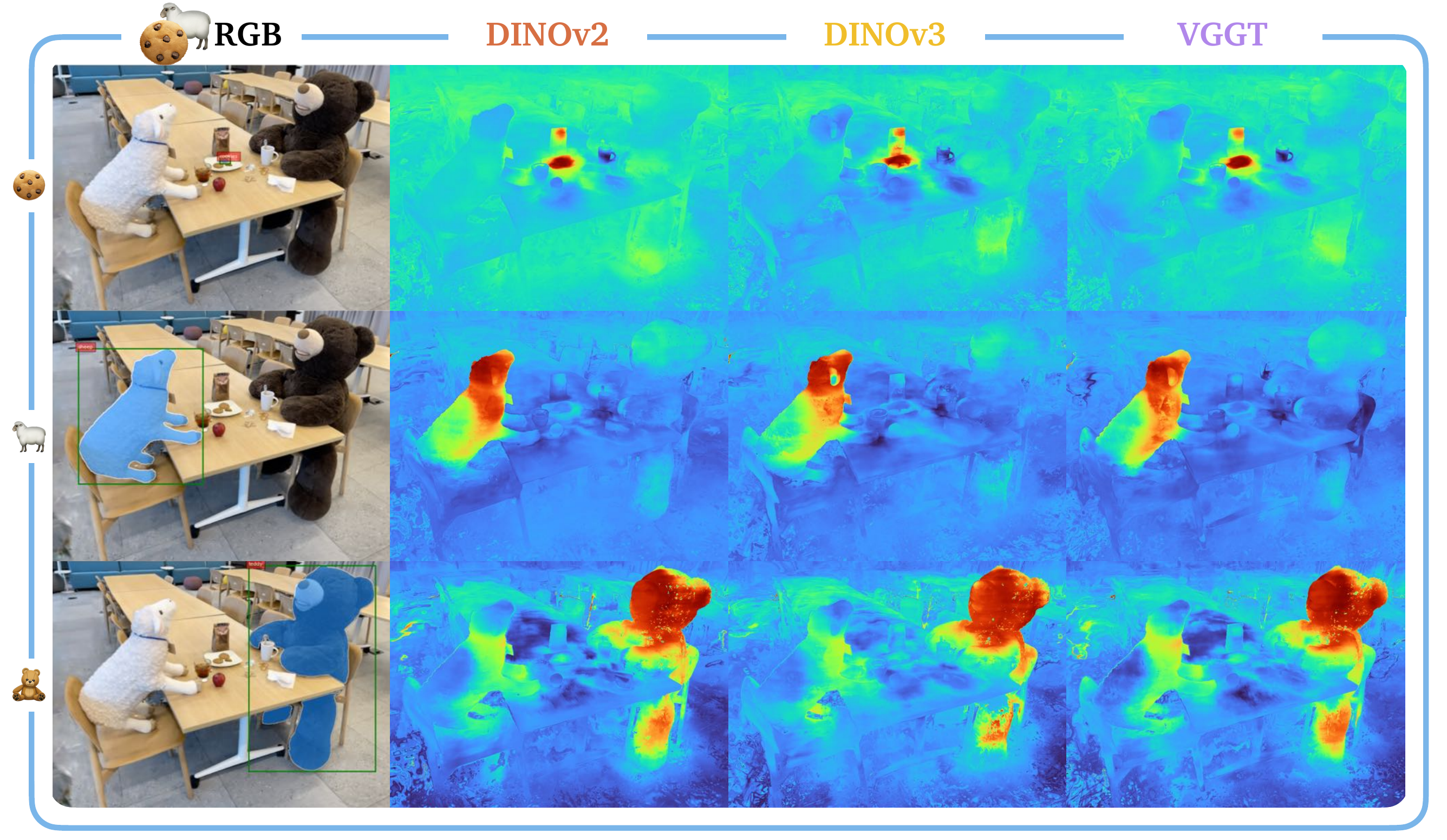}
    \caption{Semantic object localization visualized for GS for the LERF teatime scene.}
    \label{fig:gs_segmentation_lerf_teatime}
\end{figure}

\begin{figure}[H]
    \centering
    \includegraphics[page=2, width=\linewidth]{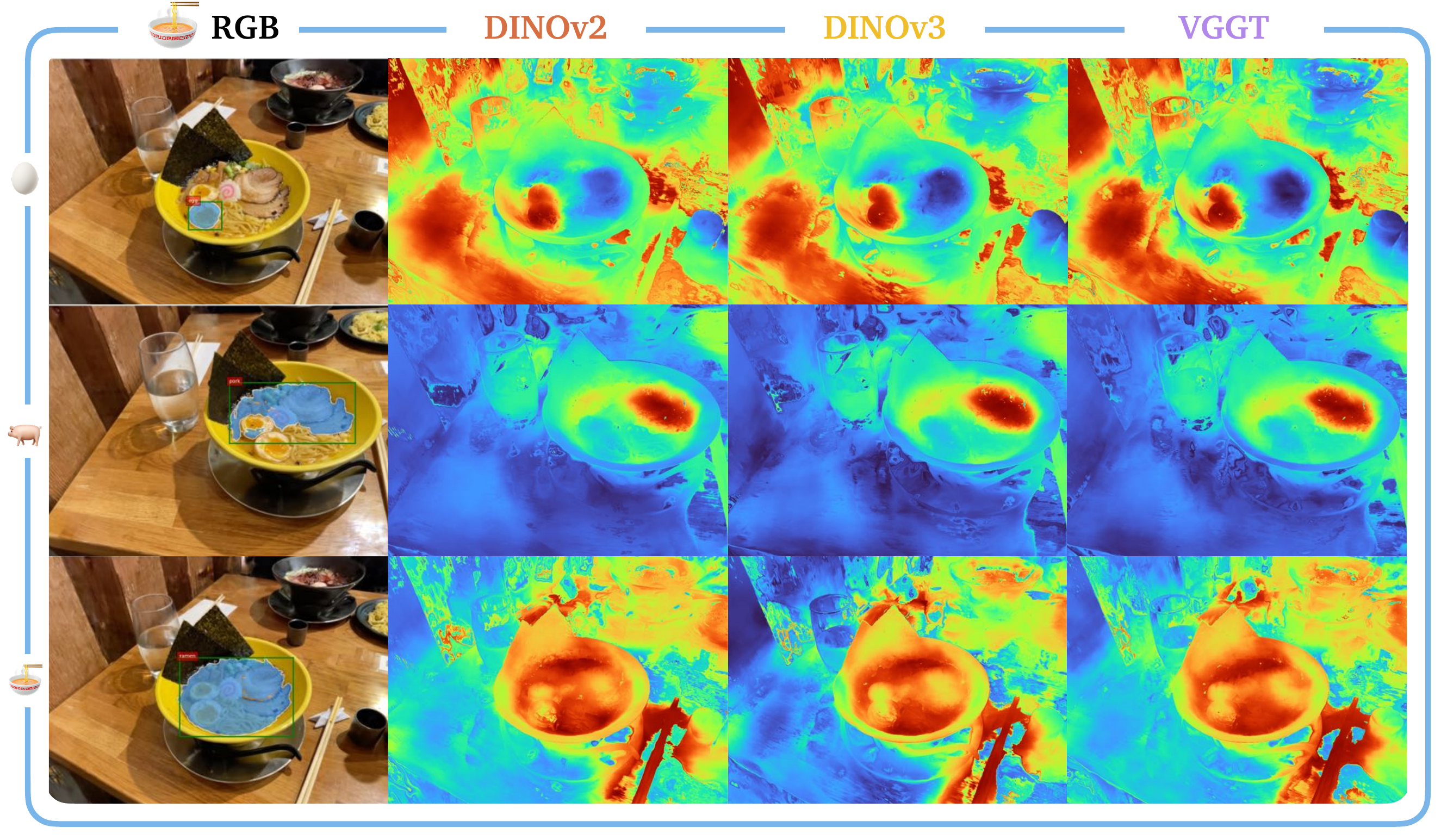}
    \caption{Semantic object localization visualized for GS for the LERF ramen scene.}
    \label{fig:gs_segmentation_lerf_ramen}
\end{figure}

\begin{figure}[H]
    \centering
    \includegraphics[page=3, width=\linewidth]{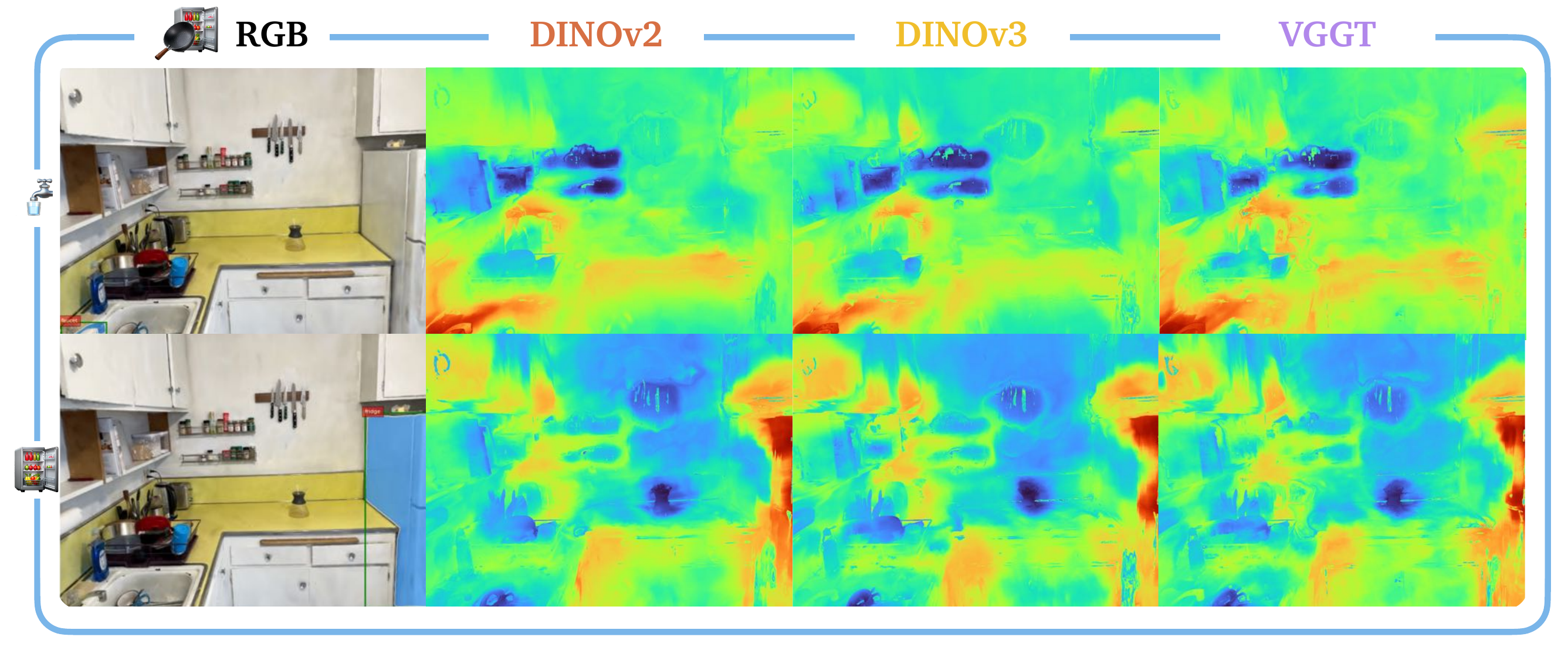}
    \caption{Semantic object localization visualized for GS for the LERF kitchen scene.}
    \label{fig:gs_segmentation_lerf_kitchen}
\end{figure}

\begin{figure}[H]
    \centering
    \includegraphics[page=1, width=\linewidth]{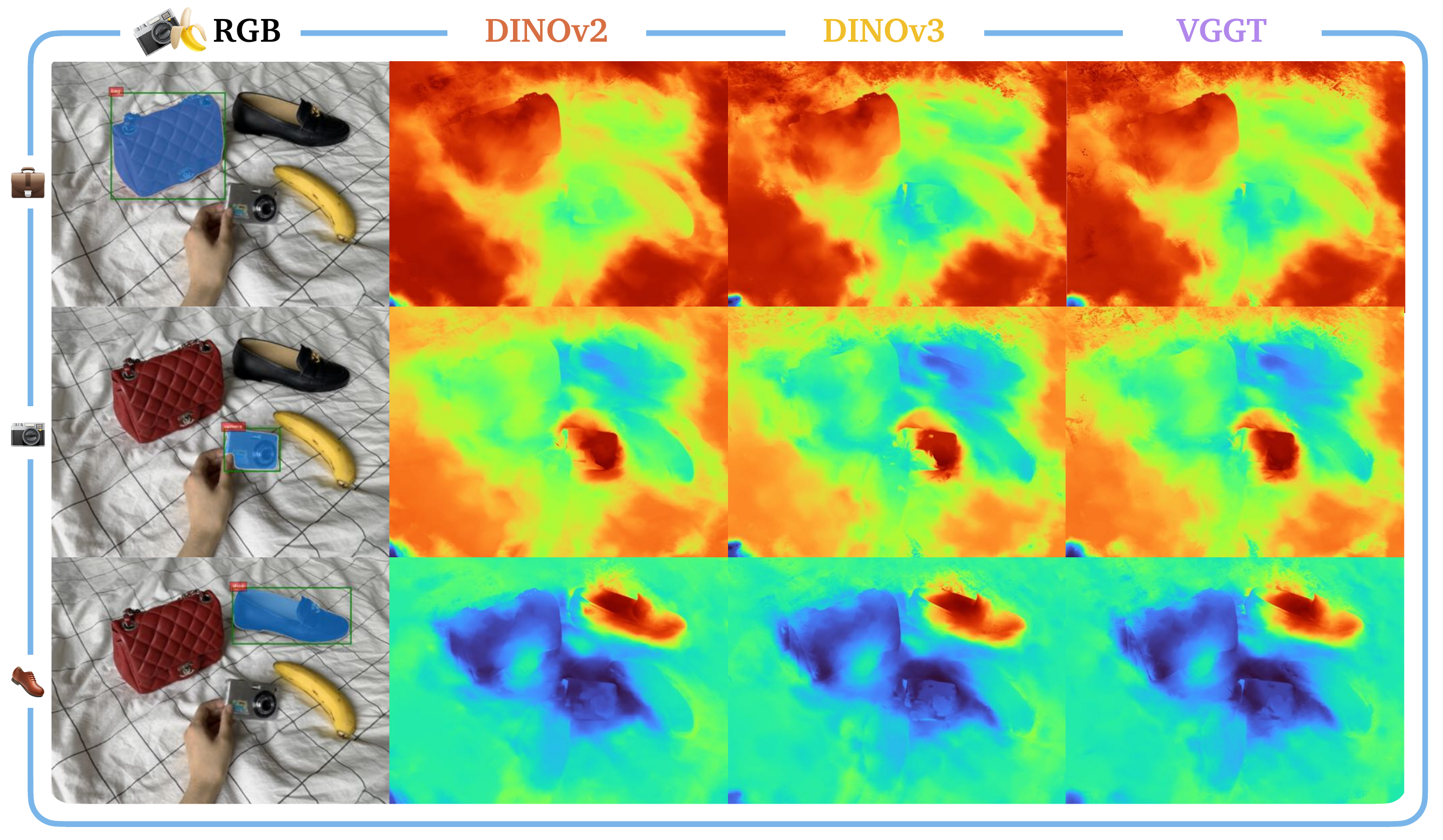}
    \caption{Semantic object localization visualized for NERF for the 3D-OVS table scene.}
    \label{fig:nerf_segmentation_3dovs_table}
\end{figure}

\begin{figure}[H]
    \centering
    \includegraphics[page=2, width=\linewidth]{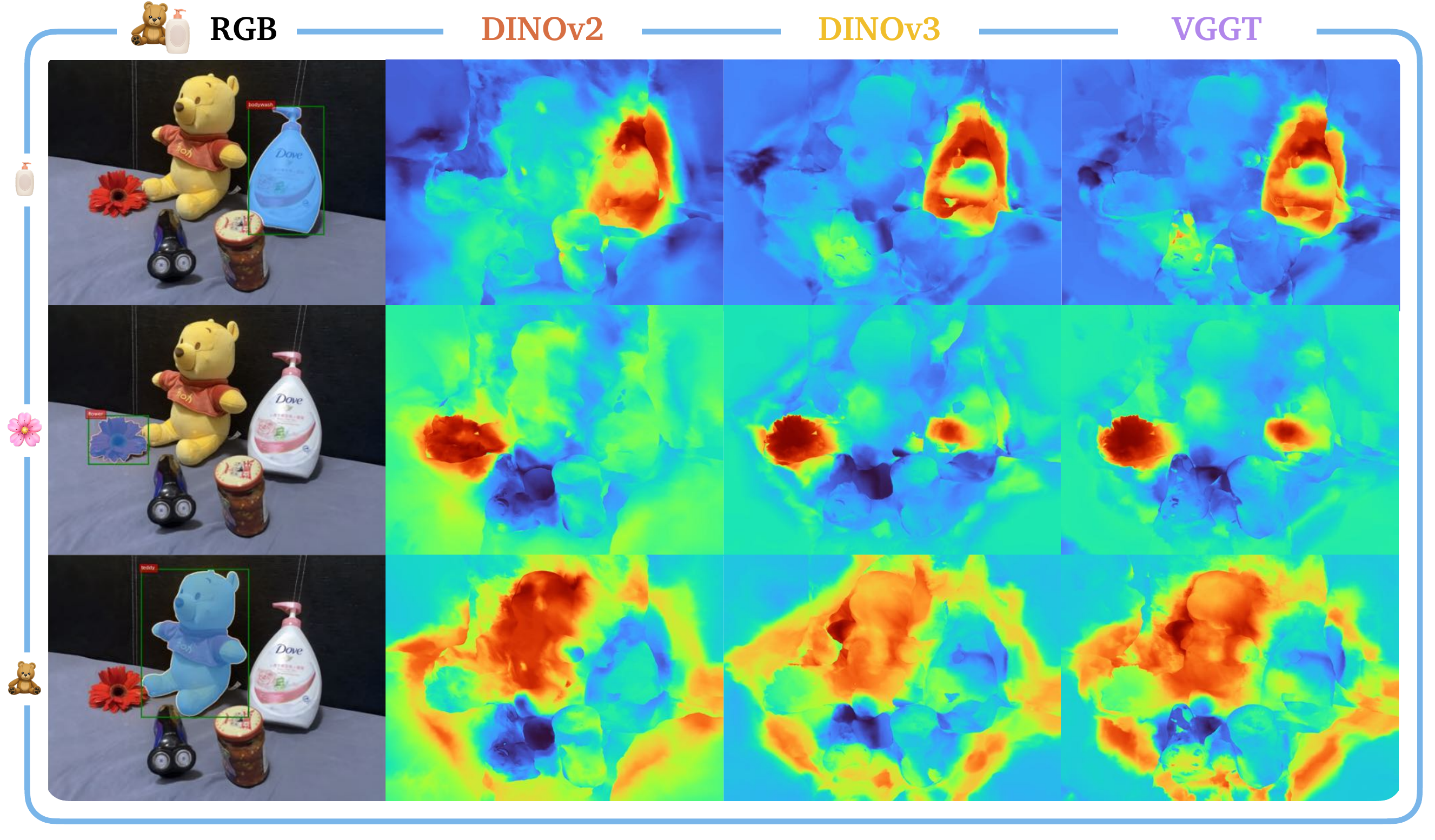}
    \caption{Semantic object localization visualized for NERF for the 3D-OVS bed scene.}
    \label{fig:nerf_segmentation_3dovs_bed}
\end{figure}

\begin{figure}[H]
    \centering
    \includegraphics[page=3, width=\linewidth]{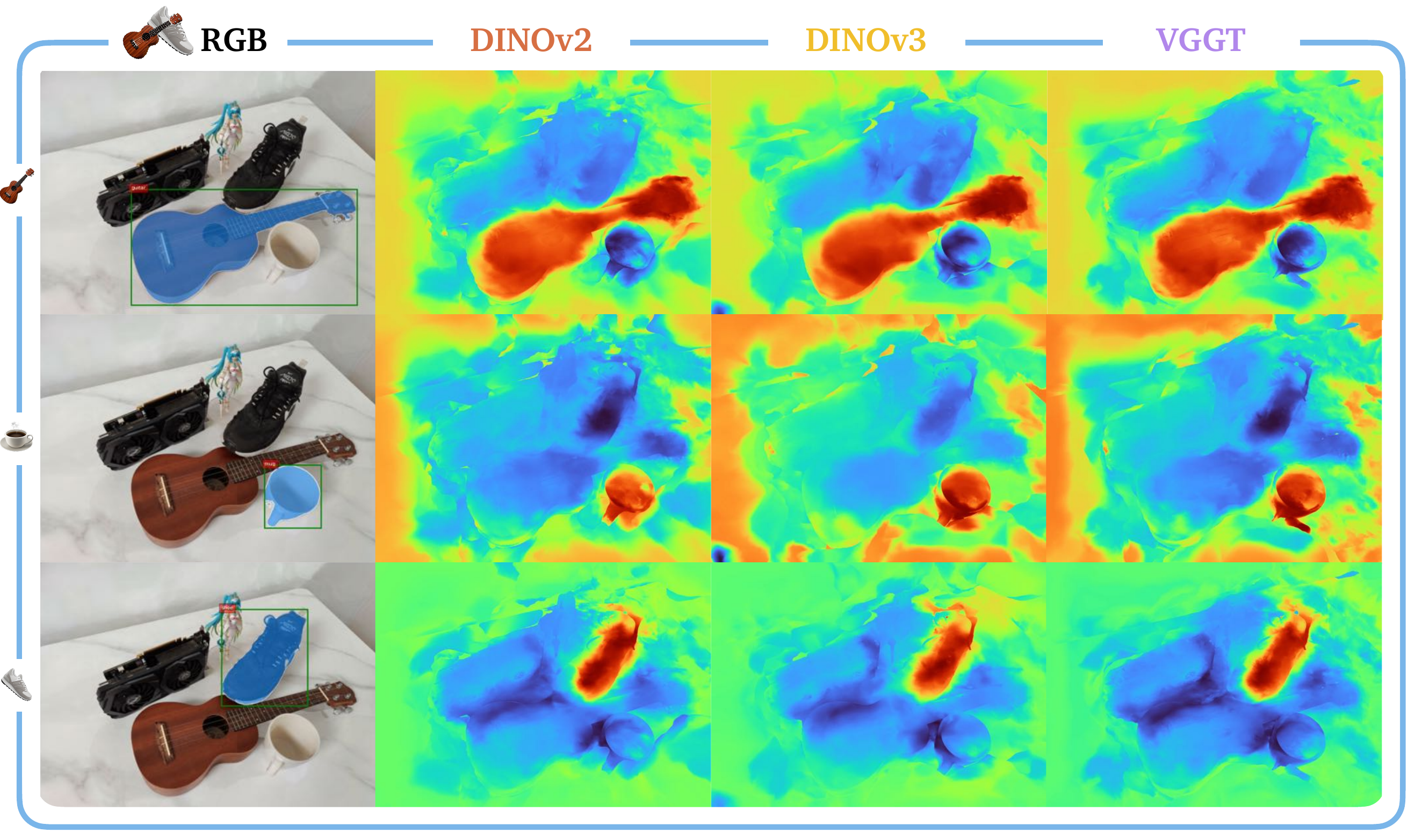}
    \caption{Semantic object localization visualized for NERF for the 3D-OVS covered desk scene.}
    \label{fig:nerf_segmentation_3dovs_desk}
\end{figure}

\begin{figure}[H]
    \centering
    \includegraphics[page=1, width=\linewidth]{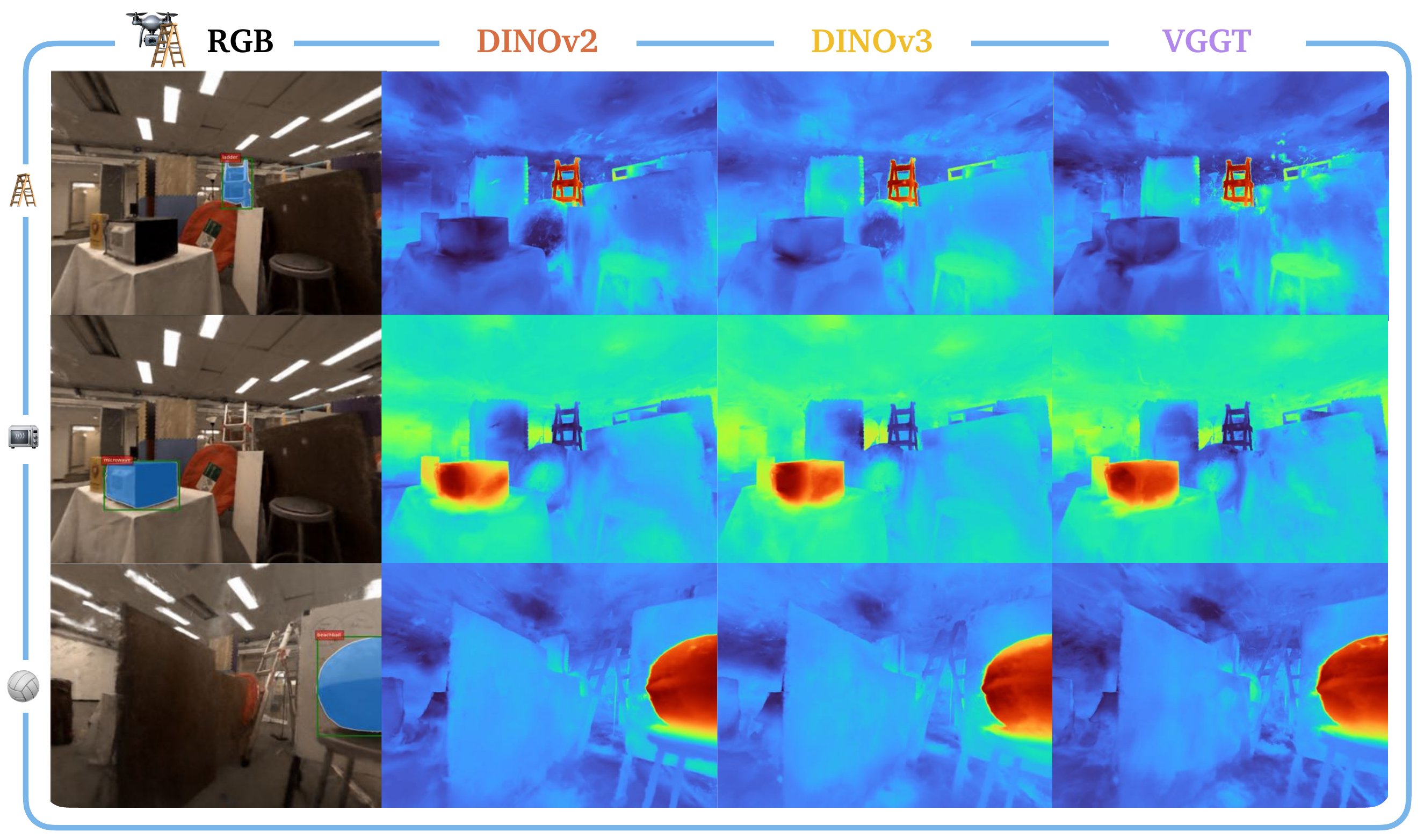}
    \caption{Semantic object localization visualized for NERF for the drone scene.}
    \label{fig:nerf_segmentation_robot_drone}
\end{figure}

\begin{figure}[H]
    \centering
    \includegraphics[page=2, width=\linewidth]{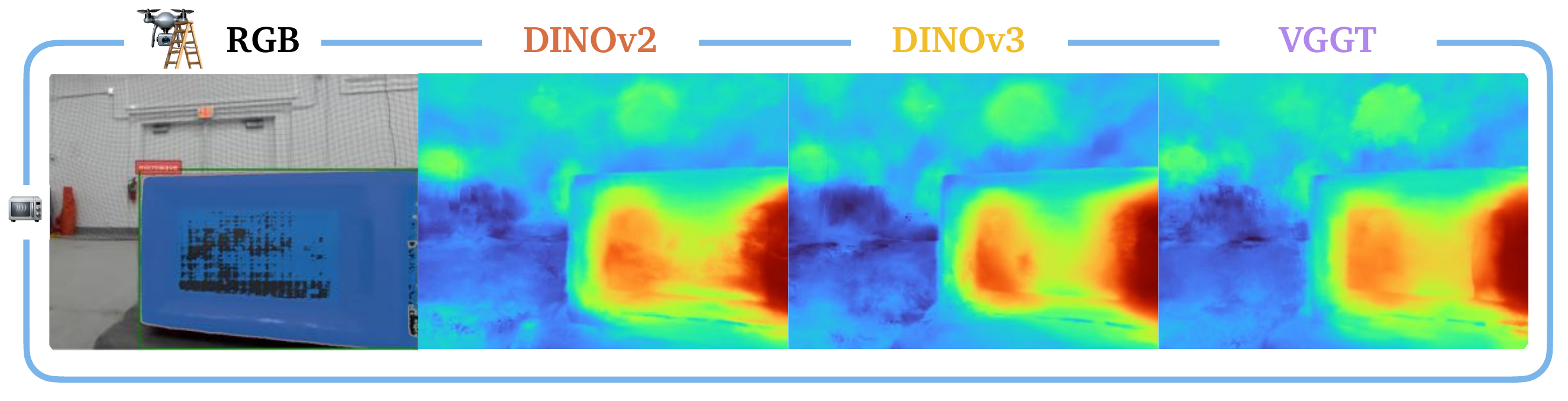}
    \caption{Semantic object localization visualized for NERF for the quadruped kitchen scene.}
    \label{fig:nerf_segmentation_robot_kitchen}
\end{figure}

\begin{figure}[H]
    \centering
    \includegraphics[page=3, width=\linewidth]{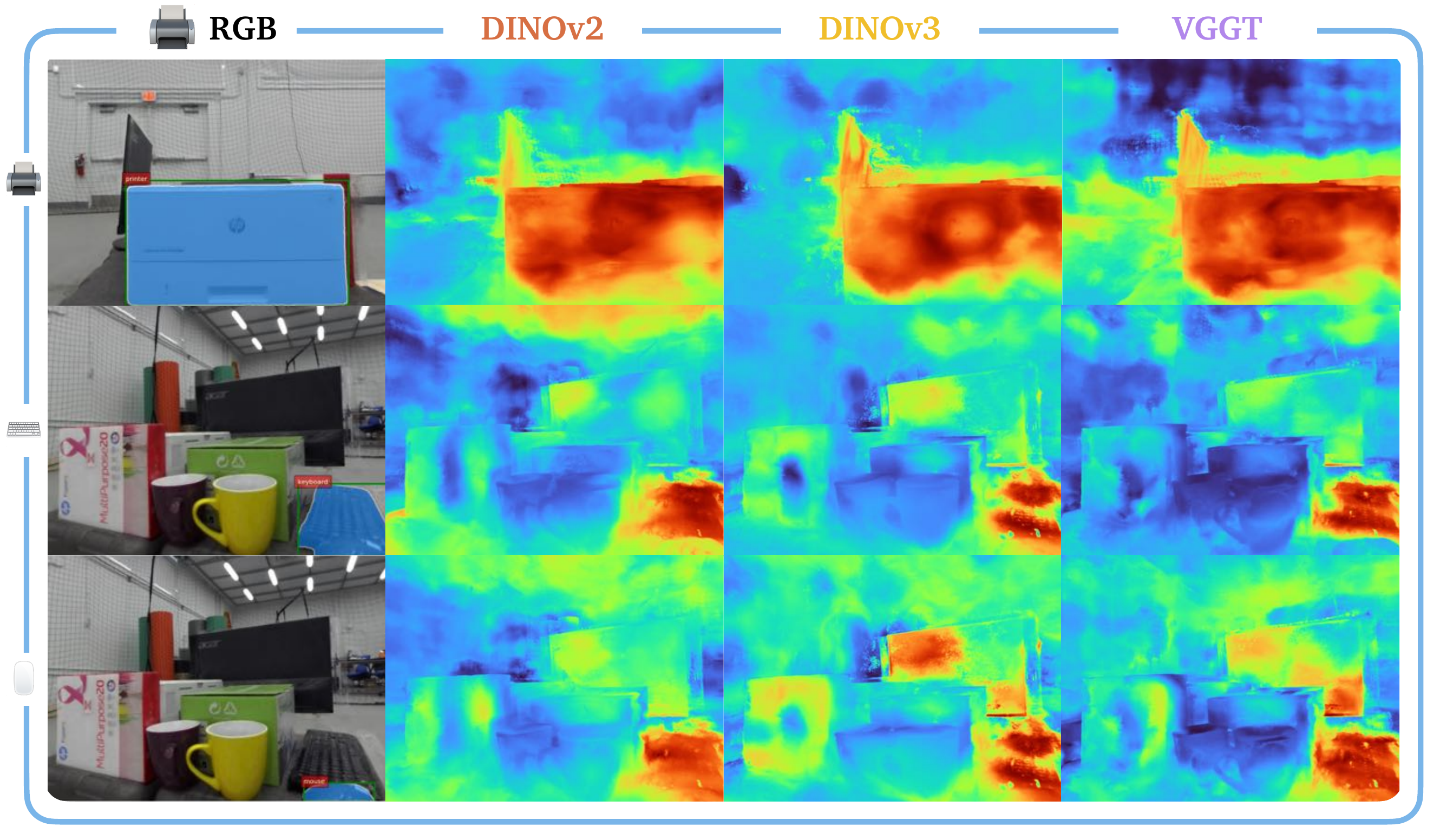}
    \caption{Semantic object localization visualized for NERF for the 3D-OVS bed scene.}
    \label{fig:nerf_segmentation_robot_office}
\end{figure}

\begin{figure}[H]
    \centering
    \includegraphics[page=1, width=\linewidth]{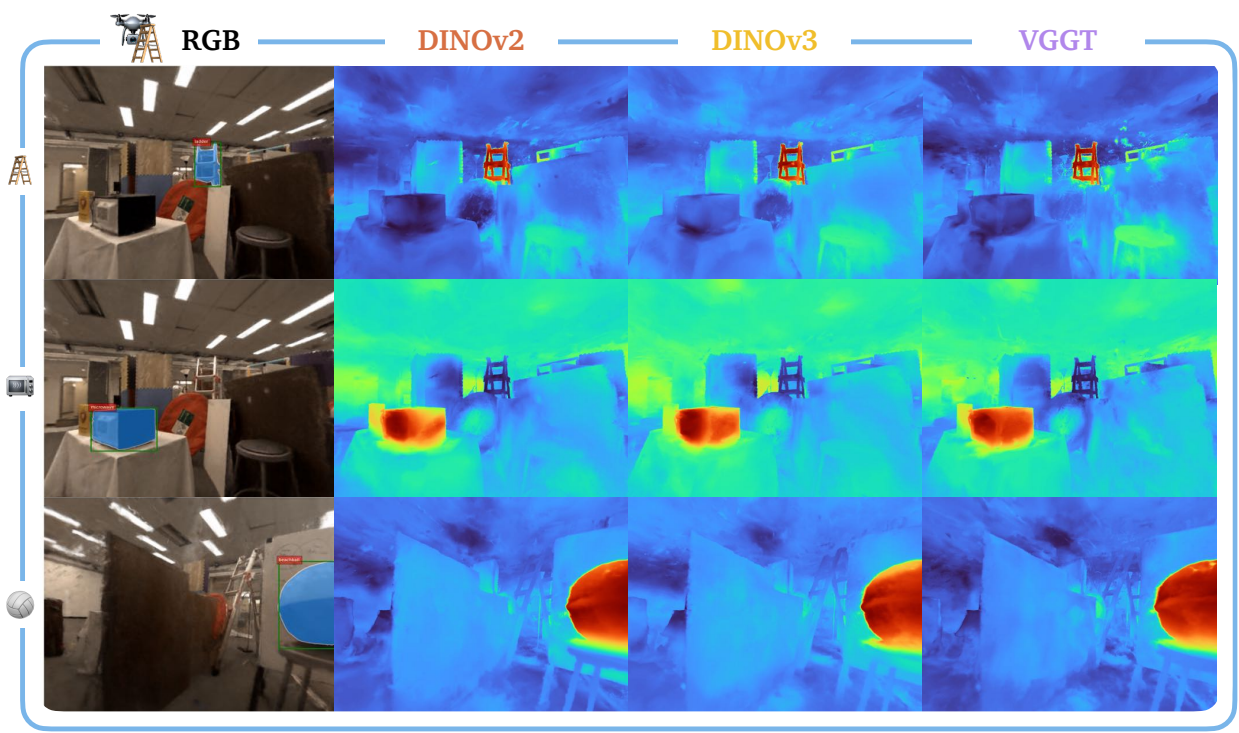}
    \caption{Semantic object localization visualized for NERF for the LERF teatime scene.}
    \label{fig:nerf_segmentation_lerf_teatime}
\end{figure}

}

\end{document}